\documentclass[acmtog,nonacm]{acmart}

\usepackage{epsfig}
\usepackage{graphicx}
\usepackage{amsmath}
\usepackage{color}
\usepackage{multirow}
\usepackage{bbm}
\usepackage{booktabs}
\usepackage{graphicx}
\usepackage{placeins}
\usepackage{adjustbox}

\usepackage{dsfont}
\usepackage{gensymb}
\usepackage{xparse}
\usepackage{csquotes}
\usepackage[ruled]{algorithm2e}
\usepackage[nameinlink]{cleveref}
\usepackage{pifont}
\newcommand{\cmark}{\ding{51}}%
\newcommand{\xmark}{\ding{55}}%

\newtoggle{nographics}
\newtoggle{lrgraphics}
\newtoggle{nocomments}

\togglefalse{nographics}
\togglefalse{lrgraphics}
\togglefalse{nocomments}
\crefname{section}{Section}{Sections}
\crefname{table}{Table}{Tables}
\crefname{figure}{Figure}{Figures}

\toggletrue{nocomments} % uncomment this to hide comments and todos

\newcommand{\Kill}[1]{}
\iftoggle{nocomments}{
    \newcommand{\SK}[1]{\ignorespaces}
    \newcommand{\DZ}[1]{\ignorespaces}
    \newcommand{\DP}[1]{\ignorespaces}
    \newcommand{\LA}[1]{\ignorespaces}
    \newcommand{\AM}[1]{\ignorespaces}
    \newcommand{\RR}[1]{\ignorespaces}
    \newcommand{\todo}[1]{\ignorespaces}
}{
    \newcommand{\SK}[1]{\textbf{\textcolor{violet}{SK: #1}}}
    \newcommand{\DZ}[1]{\textbf{\textcolor{purple}{DZ: #1}}}
    \newcommand{\DP}[1]{\textbf{\textcolor{cyan}{DP: #1}}}
    \newcommand{\LA}[1]{\textbf{\textcolor{orange}{AA: #1}}}
    \newcommand{\AM}[1]{\textbf{\textcolor{blue}{AM: #1}}}
    \newcommand{\RR}[1]{\textbf{\textcolor{green}{RR: #1}}}
    \newcommand{\todo}[1]{\textbf{\textcolor{red}{#1}}}
}

\newcommand{\tb}[1]{\textbf{#1}}

\NewDocumentCommand\rmse{}{%
    \ifmmode \text{RMSE} 
    \else RMSE\xspace 
    \fi %
}
\NewDocumentCommand\qrmse{}{%
    \ifmmode \text{RMSE-}q_{95} 
    \else RMSE-$q_{95}$\xspace 
    \fi %
}
\NewDocumentCommand\fpr{}{%
    \ifmmode \text{FPR} 
    \else FPR\xspace 
    \fi %
}
\NewDocumentCommand\recall{}{%
    \ifmmode \text{Recall} 
    \else Recall\xspace 
    \fi %
}
\NewDocumentCommand\chamfer{}{%
    \ifmmode \text{CD} 
    \else CD\xspace 
    \fi %
}
\NewDocumentCommand\hausdorff{}{%
    \ifmmode \text{HD} 
    \else HD\xspace 
    \fi %
}

\NewDocumentCommand\sinkhorn{}{%
    \ifmmode \text{SD} 
    \else SD\xspace 
    \fi %
}

\NewDocumentCommand\ndefsimodels{}{68}
\NewDocumentCommand\ndefscanmodels{}{84}

\DeclareMathOperator{\sign}{sign}
\DeclareMathOperator*{\argmax}{arg\,max}

\DeclareGraphicsExtensions{.eps,.pdf,.jpg,.png}
\iftoggle{lrgraphics}{
    \graphicspath{{src/media/lr/}{src/media/vector/}}
}{
    \graphicspath{{src/media/hr/}{src/media/vector/}}
}
\makeatletter
\iftoggle{nographics}{
    \LetLtxMacro{\includegraphics@orig}{\includegraphics}
    \RenewDocumentCommand{\includegraphics}{ s O{} m }{%
        {\setlength{\fboxsep}{0pt}%
         \colorbox{lightgray}{\phantom{\IfBooleanTF{#1}{\includegraphics@orig*}{\includegraphics@orig}[#2]{#3}}}%
        }%
    }
}{}
\makeatother

\def\vs.{vs.\spacefactor=\the\sfcode`\v}
\def\etc.{etc.\spacefactor=\the\sfcode`\c}

\makeatletter
\DeclareRobustCommand\onedot{\futurelet\@let@token\@onedot}
\def\@onedot{\ifx\@let@token.\else.\null\fi\xspace}

\def\eg{\emph{e.g}\onedot} 
\def\ie{\emph{i.e}\onedot} 
\def\cf{\emph{cf}\onedot} 
\def\etc{\emph{etc}\onedot} \def\vs{\emph{vs}\onedot}

\makeatother

\SetAlFnt{\small}
\SetAlCapFnt{\small}
\SetAlCapNameFnt{\small}
\SetAlCapHSkip{0pt}

\setcopyright{acmcopyright}
\acmJournal{TOG}
\acmYear{2022}
\acmVolume{41}
\acmNumber{4}
\acmArticle{108}
\acmMonth{7}
\acmDOI{10.1145/3528223.3530140}

\begin{document}

  \title{DEF: Deep Estimation of Sharp Geometric Features in 3D Shapes}  

  \author{Albert Matveev}
\authornote{Both authors contributed equally to the paper}
\affiliation{%
    \institution{Skoltech}
    \city{Moscow}
    \country{Russia}}
\email{albert.matveev@skoltech.ru}
\author{Ruslan Rakhimov}
\authornotemark[1]
\affiliation{%
    \institution{Skoltech}
    \city{Moscow}
    \country{Russia}}
\email{ruslan.rakhimov@skoltech.ru}
\author{Alexey Artemov}
\authornote{The author served as a technical lead for the project}
\affiliation{%
    \institution{Skoltech}
    \city{Moscow}
    \country{Russia}}
\email{a.artemov@skoltech.ru}
\author{Gleb Bobrovskikh}
\affiliation{%
    \institution{Skoltech}
    \city{Moscow}
    \country{Russia}}
\email{g.bobrovskikh@skoltech.ru}
\author{Vage Egiazarian}
\affiliation{%
    \institution{Skoltech}
    \city{Moscow}
    \country{Russia}}
\email{vage.egiazarian@skoltech.ru}
\author{Emil Bogomolov}
\affiliation{%
    \institution{Skoltech}
    \city{Moscow}
    \country{Russia}}
\email{e.bogomolov@skoltech.ru}
\author{Daniele Panozzo}
\affiliation{%
 \institution{New York University}
 \department{Courant Institute of Mathematical Sciences}
 \city{New York}
 \country{USA}
}
\email{panozzo@nyu.edu}
\author{Denis Zorin}
\affiliation{%
 \institution{New York University}
 \department{Courant Institute of Mathematical Sciences}
 \city{New York}
 \country{USA}
}
\email{dzorin@cs.nyu.edu}
\author{Evgeny Burnaev}
\affiliation{%
    \institution{Skoltech, AIRI}
    \city{Moscow}
    \country{Russia}}
\email{e.burnaev@skoltech.ru}

\renewcommand\shortauthors{Matveev, et al}

  \begin{abstract}

We propose Deep Estimators of Features (DEFs), a learning-based framework for predicting sharp geometric features in sampled 3D shapes.
Differently from existing data-driven methods, which reduce this problem to feature classification, we propose to \emph{regress a scalar field} representing the distance from point samples to the closest feature line on \emph{local patches}.
Our approach is the first that scales to massive point clouds by fusing distance-to-feature estimates obtained on individual patches.

We extensively evaluate our approach against related state-of-the-art methods on newly proposed synthetic and real-world 3D CAD model benchmarks.
Our approach not only outperforms these (with improvements in Recall and False Positives Rates), but generalizes to real-world scans after training our model on synthetic data and fine-tuning it on a small dataset of scanned data. 

We demonstrate a downstream application, where we reconstruct an explicit representation of straight and curved sharp feature lines from range scan data.

We make code, pre-trained models, and our training and evaluation datasets available at~\url{https://github.com/artonson/def}.

\end{abstract}
  



\begin{teaserfigure}
\centerline{\includegraphics[width=0.95\textwidth]{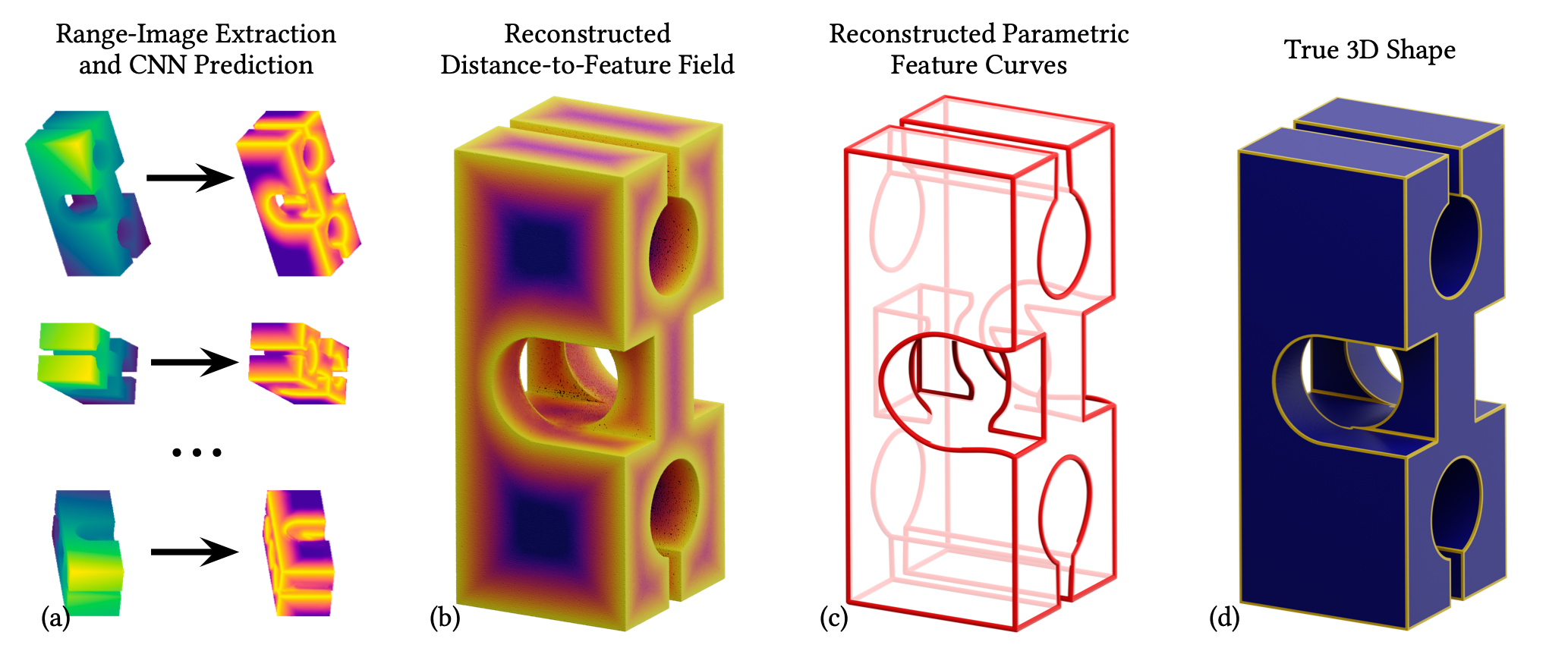}}
\caption{(a) We leverage large collections of annotated geometric data to learn highly efficient patch-based deep models of distance-to-feature fields for range scan data. 
(b) We develop a view synthesis-based approach to combining the inference of such distance-to-feature predictions into a complete estimate for a full 3D shape.
(c) Building upon our fields, we demonstrate the usage of our distance field in a downstream application, where we extract explicit representations of parametric feature curves from raw range scan data.
(d) As a result, we deliver an accurate reconstruction of geometry and topology for both straight and curved feature lines, as displayed by a reference CAD model.}
\label{fig:teaser}
\end{teaserfigure}

  \maketitle

\section{Introduction}
\label{sec:intro}

Most human-made shapes have sharp geometric features, narrow curve-like regions with normals changing rapidly across the region.
Sharp features are manually defined and explicitly stored in CAD models, and they are fundamental to faithfully represent the shape and function of CAD models.
Detecting and reconstructing sharp features from scanned data is a vital geometry processing task: sharp feature curves can be used to improve the quality of many algorithms, such as surface reconstruction, including approximation with smooth patches, shape classification, and sketch-style rendering of surfaces.  

Algorithms based on a priori analytic models of geometric features (\eg, using curvature and its derivatives) often require per-object manual parameter tuning to detect features on a specific object (Section \ref{sec:related}), making them difficult to apply to large collections of data or use as building blocks in a larger processing pipeline.
Data-driven, learning-based methods, including ours, are a natural alternative for this task as they can leverage global information extracted from a training dataset and automatically adapt to a particular input shape without user interaction.

Our goal is to develop a reliable feature detection algorithm for sampled geometric data. 
While such data comes in a variety of forms, we focus on point-sampled data, specifically of the type produced by range scanners. 
Many other geometry representations (\eg, level set meshes obtained from grid-sampled densities) can be easily converted to this form. 
Some of the most important characteristics of sampled geometric data include: (1) samples are almost never directly on (sharp) features; (2) the number of samples can be high (\eg, for a complex model, a large number of depth images are typically  combined into a single dataset with millions of points); (3) the data may be noisy. 

We propose Deep Estimators of Features (DEF), a new approach to extracting sharp geometric features from sampled shapes, designed to work with this type of data. 
We designed our algorithm with the goals of capturing features without the need to sample them exactly, scaling to complex 3D models and large, possibly noisy, point clouds naturally, while at the same time enabling compatibility with real-world 3D acquisition setups (see Figure~\ref{fig:teaser}).

Our approach is based on defining features implicitly, by a \emph{distance-to-feature function}; the problem we solve is a regression problem for this scalar function sampled in input points. 
The advantage of using a continuous distance-to-feature function, compared, \eg, to a binary classification of points as feature and non-feature points, is that it is much more natural for samples not aligned with feature and noisy samples. 

To address the need of handling large and complex models, we use local patch-based distance-to-feature prediction instead of a single-pass global prediction on the entire shape. 

As for any supervised learning method, the quality of the results depends on the quality and size of the training dataset. 
Obtaining real 3D scanned data with ground truth is difficult, as it requires either manual annotation of scanned models, or precise fabrication and scanning of CAD data with annotated features; 
we follow the latter approach for our real dataset. 
For this reason, our method uses a two-stage training process (\cf.~\cite{gaidon2016virtual} and~\cite{handa2016understanding}): we train an initial model on a large synthetic dataset and fine-tune it on a smaller dataset of 3D scanned data. 
The former is obtained by using a simplified simulated scanning process for a large number of models from ABC dataset~\cite{koch2019abc}. 
For the latter, we fabricate and scan a smaller subset of ABC models, transferring annotations from the original CAD models.

We demonstrate that our method performs favourably on a number of metrics (\rmse, \recall, \fpr) to four classical and learning-based state-of-the-art methods: VCM~\cite{merigot2010voronoi}, Sharpness Fields~\cite{raina2019sharpness}, EC-Net~\cite{yu2018ec}, and PIE-NET~\cite{wang2020pie}.

As a sample application using our algorithm, we show that an explicit parametric representation of feature curves can be extracted from the estimated distance-to-feature fields produced by our algorithm (Figure~\ref{fig:teaser}\,(c)), producing higher quality results, both qualitatively and quantitatively, than recent learning-based methods~\cite{wang2020pie,Liu:2021:PC2WF}.

\begin{figure*}[t]
\centerline{\includegraphics[width=\textwidth,trim=0 10em 0 0,clip=True]{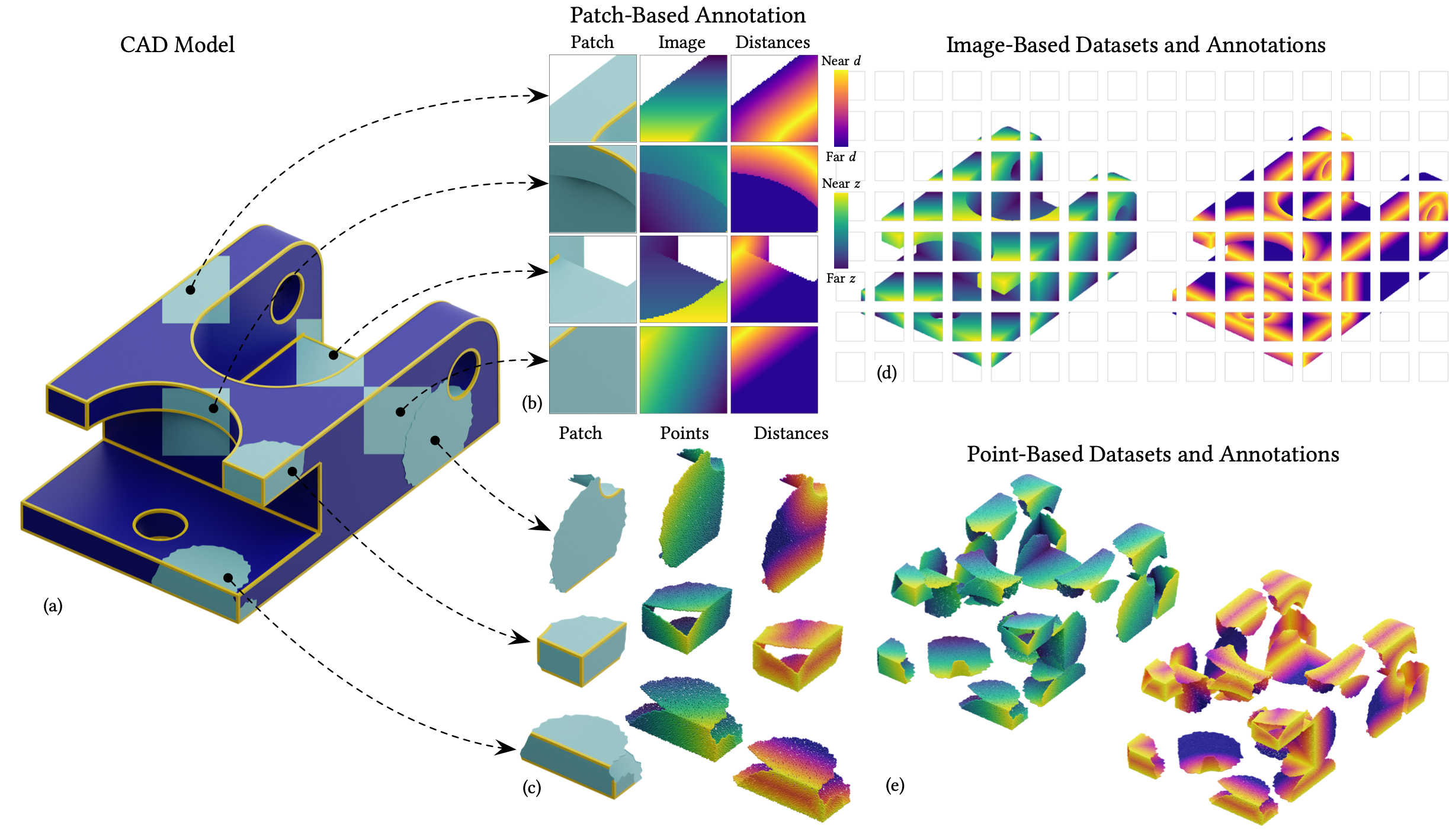}}
\caption{Our patch-based pipeline for generating image-based (b, d) and point-based (c, e) training datasets proceeds as follows:
(a) starts with a 3D CAD model,
(b)--(c) extracts local triangulated patches and associated interior sharp feature curves,
acquires ray-casted depth images and sampled point clouds,
and computes local distance-to-feature annotations.
The diversity of image and point patches in our large-scale training datasets (d)--(e) enables us to train highly effective sharp feature estimation models.}
\label{fig:datasets_synthetic_pipeline}
\end{figure*}

In summary, our contributions are:
\begin{enumerate}
\item A method for estimating coherent distance-to-feature fields for high-resolution, high complexity sampled 3D shapes, including localized, CNN-based initial estimation of the field and global fusion of local estimates.

\item A pipeline for constructing large simulated training datasets with controllable noise and different sampling patterns.
This pipeline is used to produce a collection of benchmarks suitable for comparison of geometric feature detection algorithms.

\item A process for constructing a real 3D scan dataset with ground truth distance-to-feature annotations and  a new publicly available labelled set of range scans that can be used as a realistic benchmark.

\end{enumerate}

\section{Related work}
\label{sec:related}

Estimation of sharp features has been studied extensively in computer vision and computer graphics.
We review both algorithmic methods relying on local estimation of differential surface properties and data-driven methods.

\emph{Normal Estimation, Clustering and Feature Detection on Local Sets.}
A popular family of methods~\cite{weber2010sharp,Bazazian2015,demarsin2007detection}, which can be applied directly on a point cloud or a triangle mesh, identifies a group of samples in a small area, computes their Gauss map using the samples' normals, and then performs clustering on the Gauss map to classify the neighborhood as belonging to a feature or not. 
Similar ideas can be applied to point set resampling with feature preservation~\cite{huang2013edge}.

A special case of such local estimators is Voronoi Covariance Measure estimator (VCM)~\cite{merigot2010voronoi}.
It is based on constructing Voronoi cells of the local neighborhoods of points and computing covariance matrices of these cells. 
From these matrices, normals, curvature, and feature curves can be estimated.
These methods require per-model tuning of parameters for both normal estimation and feature detection. 
In comparison, our method exploits the availability of datasets and automatically tunes its parameters to work on a collection of diverse shapes.

\emph{Surface Segmentation.} 
Instead of directly detecting features, methods based on surface segmentation identify surface patches first and then classify them as features the interface between them~\cite{Lin:2017}. 
Additional priors can be used to help the segmentation, for example, for patches that are known to be developable~\cite{Lee:2016}.
Several works~\cite{li2019supervised,sharma2020parsenet,le2021cpfn} have attempted to fit surface patches after segmentation, however these approaches do not use feature curves and produce a disconnected set of surface patches with rough boundaries.
These methods inherently require the entire model and cannot be applied to single views or incomplete models.
Differently, our approach is directly applicable to incomplete data.

\emph{Patch Fitting.}
Feature fitting methods use a predefined set of primitives~\cite{cao2016curve,torrente2018recognition} which are fitted to large regions of the mesh.
These approaches are inherently more resilient to noise but increase the computational cost and require the features to be contained within a set of predefined shapes.
Typical choices of features vary from a pair of planes sharing one edge~\cite{Lin:2015:line} to spline curves.

A related, but somewhat distinct method~\cite{daniels2007robust,daniels2008spline} relies on robust moving least squares (RMLS) \cite{fleishman2005robust}. 
This approach uses the quality of the local RMLS fit to determine the number of separate patches locally, and computes curve feature points as surface intersections, with several additional processing stages to obtain feature curves in the end.
As with other categories, many parameters need to be adjusted to obtain good results. 

\emph{Ground Truth and Representations.}
Only recently, multiple synthetic large-scale datasets with annotated features have been released~\cite{willis2020fusion,koch2019abc,sangpil2020large}.
In this work, we provide the first large-scale, objective comparison of algorithms working on triangle meshes and point clouds using the ABC dataset~\cite{koch2019abc} and a real scan dataset derived from it.

\emph{Data-Driven Approaches.}
The identification of points lying on a sharp feature is most commonly cast as a binary classification problem, using a surface neighborhood (and potentially the normals or curvature of the neighboring points) as (learning) features.
Different machine learning models were used, such as random forests~\cite{Hackel:2016,Hackel:2017}, pointwise MLPs~\cite{raina2019sharpness,yu2018ec,wang2020pie}, or capsule networks~\cite{Bazazian-EDCNet2021}. 
A recent work~\cite{himeur2021pcednet} presents a lightweight MLP-based architecture paired with differential geometry-inspired scale-space matrices that encode features discriminative for edge detection.
The methods that are closest to our work are~\cite{Liu:2021:PC2WF} and~\cite{wang2020pie}.
These approaches classify feature and corner points and fit analytic features connecting the corner points and approximating the detected features.
We compare against state-of-the-art learning-based methods, discussing results and details in Section~\ref{exper:comparative}.

\section{Overview}
\label{sec:framework}

The input to our algorithm is a set of depth images (possibly with missing data), for a given object. 
In the case of real scanned data, these images are obtained directly from the scanner; in the case of synthetic mesh data, we simulate the scanner to generate a collection of depth images from a mesh (Section~\ref{datasets:synthetic}).  
The algorithm outputs estimates of the truncated distance-to-feature scalar function for each input point. 
Figure~\ref{fig:teaser}\,(a)--(b) illustrates this process.

The four main components of our method are:
\begin{enumerate}
\setlength{\itemsep}{5pt}
\setlength{\parskip}{0pt}

\item \textit{Training Data Construction} (Section~\ref{sec:datasets}).
We generate synthetic training data using the ABC dataset~\cite{koch2019abc}, obtaining collections ranging from 16,384 to 262,144 training instances.
To fine-tune the model and evaluate its performance on real scans, we introduce a fabrication, scanning, and semi-automatic annotation pipeline to create a dataset of~84 real-world models.
Our data generation pipeline accepts a set of meshes and their associated feature annotations (edges marked as sharp) as input and produces a set of point-sampled local patches with point-wise distance-to-feature labels as output (Section~\ref{datasets:design}). 
We specify the details on the implementation of our two datasets, the synthetic \emph{DEF-Sim} and the real-world \emph{DEF-Scan}, in Sections~\ref{datasets:synthetic}--\ref{datasets:real-world}.

\item \textit{Patch-Based Deep Estimators} (Section~\ref{methods:learning}).
We train a family of deep feature estimators (DEF), which produce distance-to-feature estimates, on patches (depth images) of the synthetic dataset and fine-tune on a subset of the real-world dataset.  

\item \textit{Estimation on Complete 3D Models} (Section~\ref{methods:fusion}). 
The per-patch distance-to-feature predictions produced by DEFs are fused together by transferring estimates from each patch to overlapping patches and combining into a coherent global estimate.

\item \textit{Feature Fitting} (Section~\ref{sec:parametric}). 
The last (optional) component extracts explicit feature curves from the distance-to-feature function. 
We show that with our distance function estimate, simple corner detection, combined with kNN clustering and spline fitting, produces higher quality results than state-of-the-art methods.
\end{enumerate}

In the next sections, we describe each component in detail and provide a rationale for each algorithmic choice. 
  \section{Datasets with distance-to-feature annotation}
\label{sec:datasets}

\subsection{Dataset Design}
\label{datasets:design}

\emph{Feature Definition.} 
Each CAD model in the ABC dataset is defined by a boundary representation (B-Rep), providing a partitioning of its surface into a collection of CAD regions and associated parametric curves. 
Analytically, we identify sharp features as curves at the interface between any two regions for which the normal orientations defined in either region differ by more than a particular threshold $\alpha_{\text{norm}}$ (we use $\alpha_{\text{norm}} = 18\degree$) as was done during the construction of ABC dataset~\cite{koch2019abc}. 
The threshold is necessary as CAD models commonly have smooth areas partitioned in multiple regions, which would result in spurious features.

Directly using the original parametric representations, however, makes it difficult to construct a large training dataset, as B-Reps either need to be traversed using off-the-shelf geometric kernels~\cite{Opencascade,parasolid}, a software not designed for batch processing, or require re-implementing a set of elementary operations like closest point, which require nonlinear solvers on B-Reps.
To avoid these issues, we use the triangulated versions of the ABC models, where CAD region and sharp feature curve labels are available for vertices and edges in each mesh; we introduce a set of easily testable geometric conditions into our data generation procedure to prevent introducing significant geometric errors when sampling B-Rep data.  We use the curve annotation provided in the ABC dataset to identify the mesh edges which were marked as sharp to base our distance field on the proximity to the corresponding mesh edge.

\begin{figure}[t]
\centerline{\includegraphics[width=0.95\columnwidth]{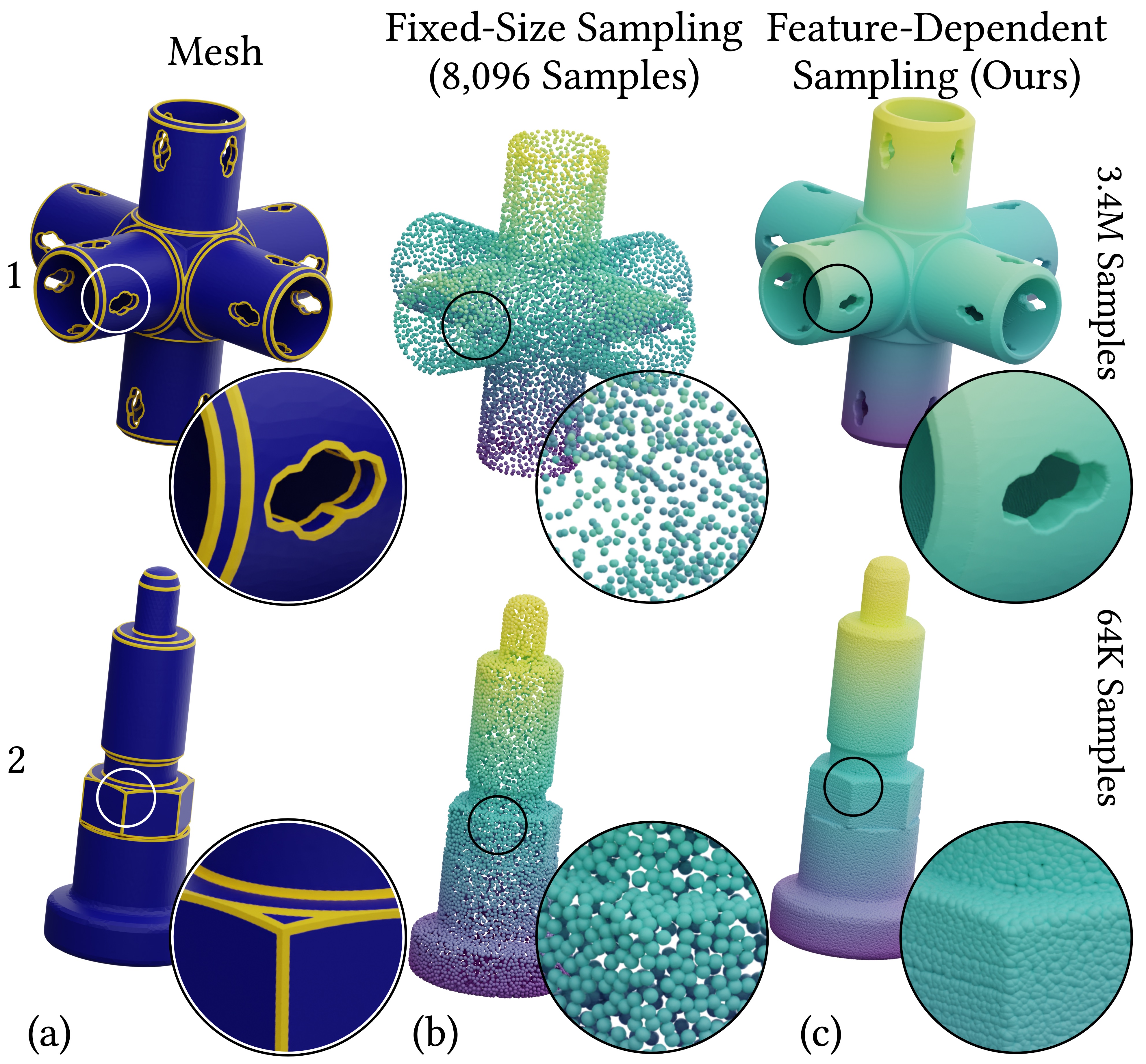}}
\caption{Differently from existing approaches, that represent all mesh models (a) by a fixed number of samples (b) despite dramatic differences in their geometric complexity (\cf rows 1 and 2), we decompose input 3D models into variable-length sets of local patches with a fixed number of samples; as a result, complete 3D shapes sampled using our method have variable number of samples (c).}
\label{fig:datasets_design_models_complexity}
\end{figure}

\emph{Patch and Feature Selection.}
Mesh models in ABC vary significantly in geometric complexity~\cite{koch2019abc}, requiring an adaptive number of samples to represent their 3D surface geometry (in the original dataset, meshes are sampled with~$10^2$--$10^7$ vertices), see Figure~\ref{fig:datasets_design_models_complexity}.
However, having variable size, high resolution 3D shapes as input is not a good fit for training most state-of-the-art learning algorithms, which require a fixed number of samples and require too much memory and training time to handle hundreds of thousands of samples~\cite{henderson2020towards}.
To address this problem, we decompose each shape into a \emph{collection of patches} with a small and fixed number of samples, see Figure~\ref{fig:datasets_synthetic_pipeline} (a)--(c); this is different from a number of existing trainable approaches~\cite{wang2020pie} that represent \emph{entire shapes} with the same (fixed) number of samples.

\begin{figure}[t]
\centerline{\includegraphics[width=0.95\columnwidth]{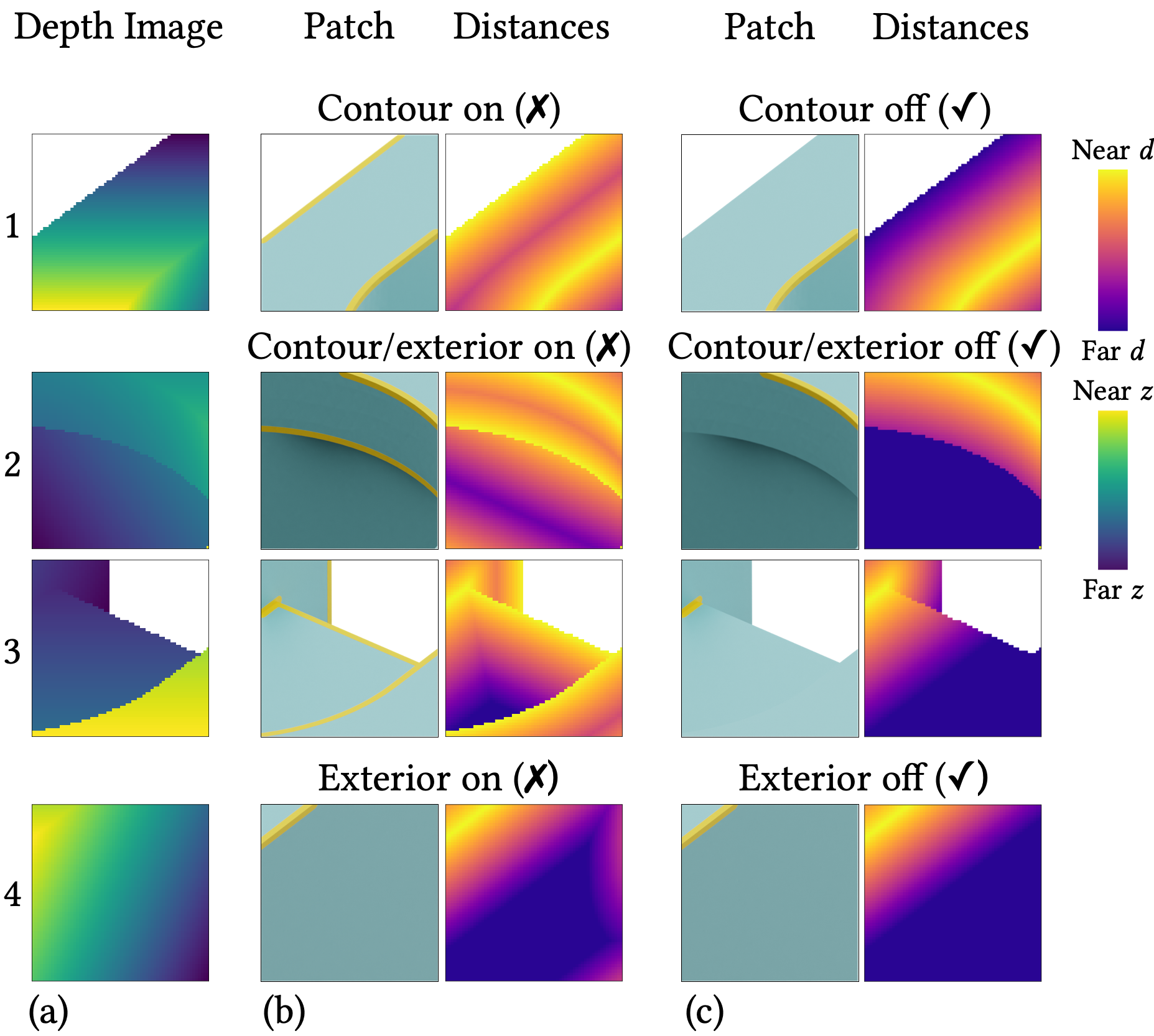}}
\caption{The same depth data in column (a) may be annotated differently, depending on which adjacent feature curves are included when computing distances. 
\emph{Contour} features (\ie, features adjacent to only a single visible surface patch; shown in column (b), rows 1--3) are difficult to distinguish from smooth contours; \emph{exterior} features in close proximity (\ie, features passing outside patch but within distance truncation radius $\varepsilon$; shown in column (b), rows 2--4) are impossible to detect due to absence of samples covering them.
We opt to generate the per-patch annotations locally in each patch, using only feature curves passing through the patch \emph{interior} (\ie, both adjacent surface patches are sampled, shown in column (c), rows 1--4).
}
\label{fig:datasets_design_patch_feature_selection}
\end{figure}

Selecting patches and feature curves for training has a direct impact on performance.
We distinguish between \textit{interior,} \textit{contour,} and proximal \textit{exterior} curves, depending on their visibility status; we keep interior curves for annotation and exclude the latter two types.
Features appearing as a contour of a sampled region are difficult to distinguish from smooth features; being adjacent to only a single visible surface patch provides insufficient spatial context for inferring these from point samples.
Exterior features pass within distance truncation radius~$\varepsilon$ but still outside the visible patch. 
Including exterior features would lead to distance-to-feature annotations indicating feature proximity; however, regressing such features from the local patch context would be impossible due to absence of samples covering them.
In contrast, we generate the per-patch annotations locally in each patch, using only feature curves passing through the patch interior.
Figure~\ref{fig:datasets_design_patch_feature_selection} demonstrates example annotations obtained by varying the set of included features.

Similarly, patches with depth discontinuities and gaps represent challenging cases with many contour feature curves, see Figure~\ref{fig:datasets_design_patch_feature_selection}, rows 2--3; however, these naturally occur due to shape self-occlusions or ray misses during both ray-casting and real scanning.
We have experimentally observed that including such instances in training improves performance, particularly at near-boundary pixels that are regressed more accurately; we discuss their effect and alternatives in our ablative experiment (Section~\ref{exper:ablative}).

\emph{Distance-to-Feature Computation.}
As our focus is on sharp feature detection, large values of the  distance-to-feature function have little impact on feature localization  but require more effort to predict correctly. For this reason, we define a \emph{truncated} distance-to-feature field $d^{\varepsilon}(p)$ in each location $p \in \mathbb{R}^3$ using the proximity to a subset of mesh edges corresponding to (sharp feature) curve segments $\Gamma = \{\gamma_k\}_{k = 1}^K$ in $\mathbb{R}^3$ as follows. 
We find for $p$ its closest (in Euclidean sense) neighbor located at one of the segments in~$\Gamma$, \ie a point $q(p)$ such that
\begin{equation}
\|q(p) - p\| = \min\limits_{\gamma_k \in \Gamma} \inf\limits_{q \in \gamma_k} \|q - p\|,
\label{eq:distance_to_features}
\end{equation}
and define the $d^{\varepsilon}(p)$ by 
\begin{equation}
d^{\varepsilon}(p) = \min(\|q(p) - p\|, \varepsilon),
\label{eq:truncated_distance_to_features}
\end{equation}
where we set our truncation radius~$\varepsilon$ to a multiple of the sampling distance~$r$ (we set $\varepsilon = 50$, $r_{\text{high}} = 1$ where $r_{\text{high}} = 0.02$ is a base sampling step), leaving a sufficiently wide envelope where our distance field may provide meaningful feature-related information.

\begin{figure}[t]
\centerline{\includegraphics[width=0.95\columnwidth]{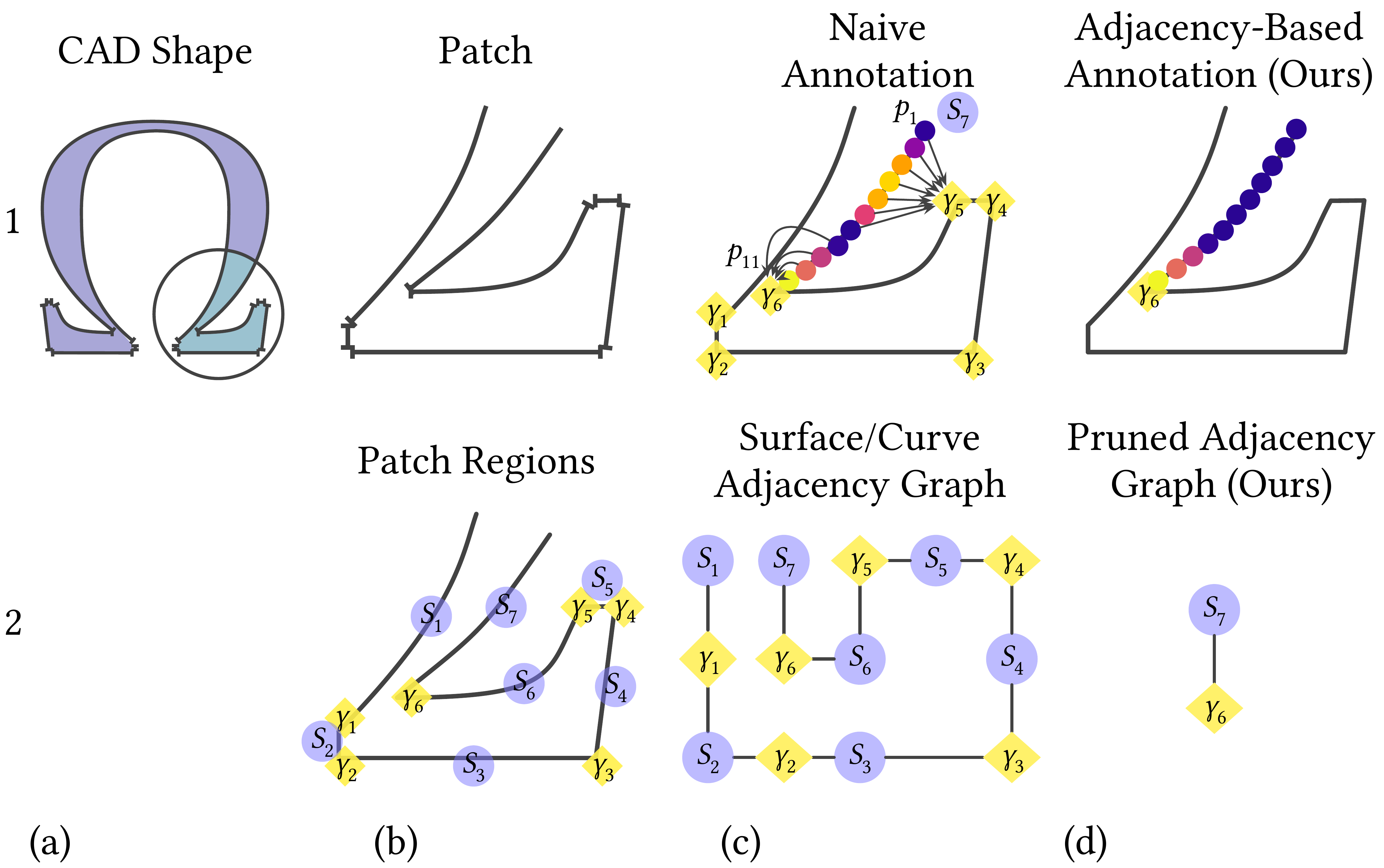}}
\caption{Extracting a patch from an example 2D CAD shape in (a) produces a mesh fragment consisting of seven surface regions $S_l$ along with six associated interior feature curves $\gamma_k$ (rows 1--2 (b)). 
For samples~$p_i \in S_7$, naive computation of distances $d^{\varepsilon}(p_i)$ maps $p_1, \ldots, p_7$ to the feature $\gamma_5$ (row 1 (c)) which is disconnected from the region $S_7$, despite proximity in the Euclidean sense (row 2 (c)). 
In contrast, we compute more natural distances, excluding non-contour curves for each surface region (for $S_7$, all but $\gamma_6$ are excluded as in row 1 (d)) by constructing and pruning the surface/curve adjacency graph (row 2 (d)).}
\label{fig:datasets_distance_computation}
\end{figure}

We use Euclidean distance as opposed to the geodesic distance along the surface.
We compute distance-to-feature annotation for a sampled point $p$ by associating it to the closest surface spline region within the patch (this association accounts for sampling noise) and only considering sharp feature curves belonging to the contour of that surface region in the ABC feature annotation, see Figure~\ref{fig:datasets_distance_computation}.
More generally, we construct a surface region/feature curve adjacency graph where each surface region and feature curve (two nodes) that share mesh vertices are connected by an edge, and perform depth-first search of depth~$k$ to determine which features should be included in the distance computation over a particular surface region.
We additionally record $q(p) - p$, directions to the closest points on the feature curves, for use in the ablation study.

\emph{Feature Size and Sampling Density.}
To accurately reconstruct the distance-to-feature function, it is not safe to rely on \emph{fixed-size} input point clouds for whole objects (as it is done in recent literature~\cite{wang2020pie,Liu:2021:PC2WF}), since many curves are left severely undersampled, see Figure~\ref{fig:datasets_design_models_complexity}.
Instead, we assume that most feature curves are sufficiently densely sampled, and that the presence of feature curves can be inferred from the positions of samples; that leads us to have an \emph{adaptive} number of point samples per object.
This assumption is motivated by a common practice in high-quality 3D data acquisition of adapting the number of points per object and sensor placement to the geometric complexity and size of the object.

\begin{figure}[t]
\centerline{\includegraphics[width=0.95\columnwidth,trim=0 10mm 0 0, clip]{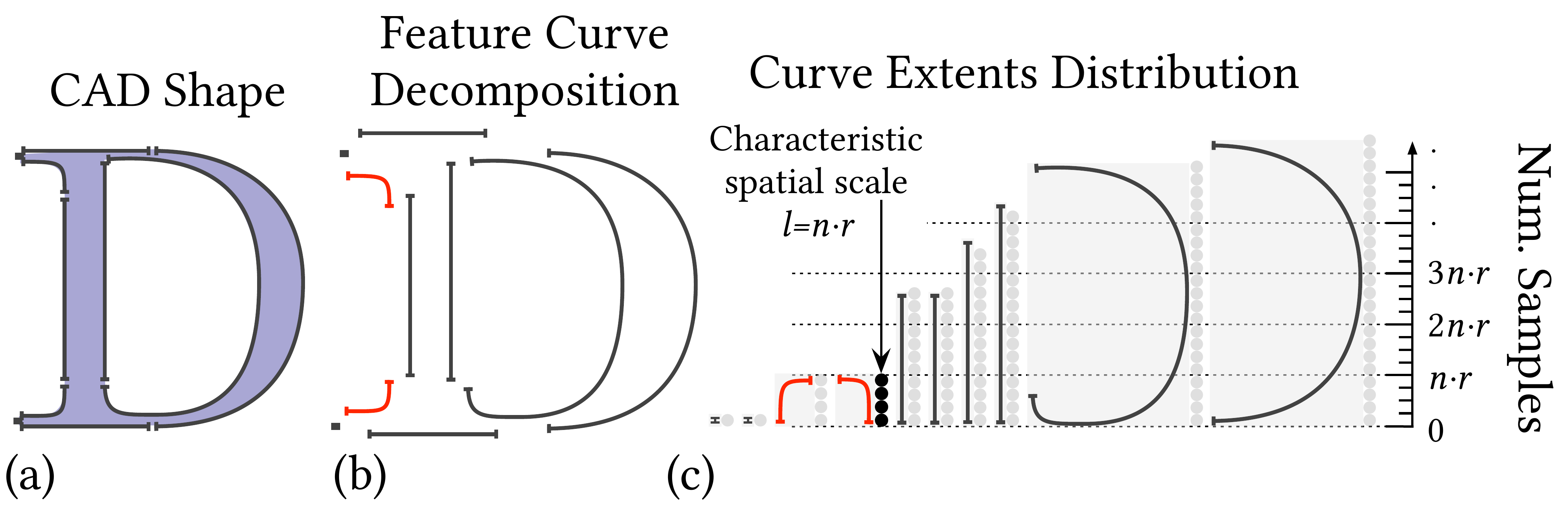}}
\caption{For an input CAD shape in (a), 
we analyze the distribution of sharp feature curve extents in (b) 
and relate a sampling radius~$r$ to features of characteristic spatial size~$l$, sampling these with at least~$n$ points in (c) (see Equation~\eqref{eq:sampling_density} and surrounding text).}
\label{fig:datasets_design_sampling}
\end{figure}

One way to reason about \enquote{sufficient} sampling is to choose a characteristic (object-dependent) spatial size $l$ for each shape and require that features of size close to~$l$ are represented by, on average, $n$~samples.  
Formally, we require the following relation to hold:
\begin{equation}
% r \cdot n = l \cdot s,
\underbrace{r}_{\substack{\text{sampling} \\ \text{distance}}} \times 
\underbrace{n}_{\substack{\text{num. samples} \\ \text{per feature}}} = 
\underbrace{l}_{\substack{\text{characteristic} \\ \text{spatial size}}} \times 
\underbrace{s}_{\substack{\text{scaling} \\ \text{factor}}},
\label{eq:sampling_density}
\end{equation}
where we are free to vary either the sampling step $r$ or the object scaling factor $s$ to achieve the equality (in practice, for each particular dataset, we fix~$r$ and vary~$s$).
Our characteristic spatial size~$l$ is a linear measure set to to 25\% lower quantile of the distribution of sharp feature curve extents, where \enquote{extent} denotes the maximum of three dimensions of an axis-aligned bounding-box enclosing a curve.
Figure~\ref{fig:datasets_design_sampling} provides an illustration of this scheme.

\emph{Patch-Based Datasets.}
We run our patch generating algorithm on the first five chunks of the ABC dataset (37,945 3D shapes) and obtain three major data varieties at low, medium, and  high resolution by choosing $n_{\text{low}} = 8$, $n_{\text{med}} = 2.5 \times 8 = 20$, and $n_{\text{high}} = 2.5^2 \times 8 = 50$ samples per curve. 
Each resolution corresponds to sampling distance $r_{\text{low}} = 0.125$,  $r_{\text{med}} = 0.05$, and $r_{\text{high}} = 0.02$, respectively.
Similarly to related methods~\cite{yu2018ec,wang2020pie}, we model acquisition uncertainty using additive Gaussian white noise; we use five scales in the viewing direction with a standard deviation $\sigma \in \{\frac {r} {8}, \frac{r}{4}, \frac{r}{2}, r, 2r\}$, for the high-resolution data only.
For each of the mentioned variations we obtain training sets of sizes ranging from 16,384 to 262,144 patches to assess the impact of dataset size on performance (see Supplementary material for details).

\emph{Complete 3D Model Datasets.}
Complementing our patch-based data, we constructed datasets of 3D shapes representing object-level data samples of 3D CAD models, both \textit{synthetic} and \textit{real}.

We emphasize that the design principles outlined in this section are used uniformly for both our synthetic and real-world datasets, enabling direct fine-tuning of our networks for the real scenario.

We have selected a diverse set of \ndefsimodels~distinct CAD models from the ABC dataset.
Our focus when choosing the models is to cover a variety of qualitative properties, including 
(1) presence of thin walls and 
(2) various types of surface regions (\eg, flat, cylindrical, splines, and spheres), 
(3) curved and straight features, 
(4) variety of angles incident on sharp features,
and (5) presence of fillets.
The statistics of the selected models are analyzed in the Supplemental.
The models are sampled and annotated as described in this Section to form the input complete 3D shapes.

\subsection{Synthetic Datasets: \emph{DEF-Sim}}
\label{datasets:synthetic}

Our synthetic datasets provide collections of local patches and \ndefsimodels~complete 3D models in varieties of low, medium, and high resolution, and several noise levels.

\paragraph{Shape Sampling.}
We set up $n_v$ virtual cameras with locations evenly distributed on a sphere around an object (we use Fibonacci sampling~\cite{hannay2004fibonacci}) and the $z$-axis pointing at its center of mass.
For each camera, we create a regular grid (image) with $64 \times 64$ pixels (we specify $r$ as the pixel size) and cast rays from each pixel's corner in a direction perpendicular to the grid, obtaining patches with up to 4,096 point samples each (some may not correspond to an object point and are set to a background value).

Knowing the camera parameters $(K, T)$ where $K \in \mathbb{R}^{2 \times 3}$ is an intrinsic matrix transforming point coordinates from the  camera coordinate frame to the image plane and $T \in \mathbb{R}^{4 \times 4}$ an extrinsic camera matrix transformation from the  camera coordinate frame  to a global coordinate frame~\cite{hartley_zisserman_2004}, sampled points $p_{ij} = (x_{ij}, y_{ij}, z_{ij})$ (in homogeneous coordinates) may be identified with a depth image $I = (z^{\text{cam}}_{ij})$, where $z^{\text{cam}}_{ij} = (K T^{-1} p_{ij})_3$ denotes $z$-coordinate of point $p_{ij}$ in the camera frame. 
We create the distance-to-feature annotations image by computing $d = (d^{\varepsilon}(p_{ij}))$ and record the pair $(I, d)$ as the training instance. 
We use $n_v = 18$ and augment the dataset by rotating and offsetting the image grid during data generation, but maintaining the same orientation of $z$-axis; we discuss the effect of having varying number of views~$n_v$ in the ablation study (Section~\ref{exper:ablative}).

\begin{figure*}[t]
\centerline{\includegraphics[width=0.95\textwidth]{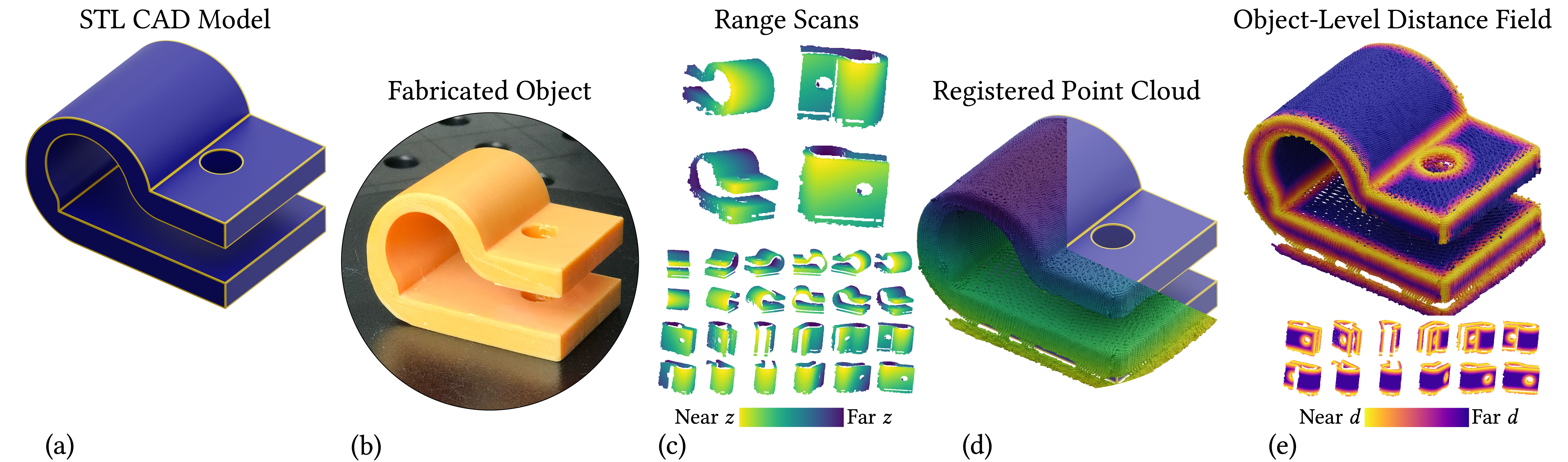}}
\caption{
(a) We have selected a diverse set of 84~3D CAD models from the ABC dataset and (b) fabricated them in thermoplastic using the 3D printing technology. 
(c) We further obtained 12 scans of each shape in two different orientations (totalling 24 scans per object) using a commercial structured-light 3D scanner. 
(d) We semi-automatically registered the 3D scans onto the original CAD model, computed distance-to-feature annotations in (e), and finally processed the scans to obtain our patch-based datasets.}
\label{fig:datasets_realworld_pipeline}
\end{figure*}

\subsection{Real-World Datasets: \emph{DEF-Scan}}
\label{datasets:real-world}

To support generalization to real-world scanning data, we constructed a dataset of \ndefscanmodels~real objects and semi-automatically annotated them.
Figure~\ref{fig:datasets_realworld_pipeline} presents an overview of the steps involved in the construction of our datasets; details on the selection of CAD models are mentioned in Section~\ref{datasets:design}.

\begin{figure}[b]   
\centerline{\includegraphics[width=\columnwidth]{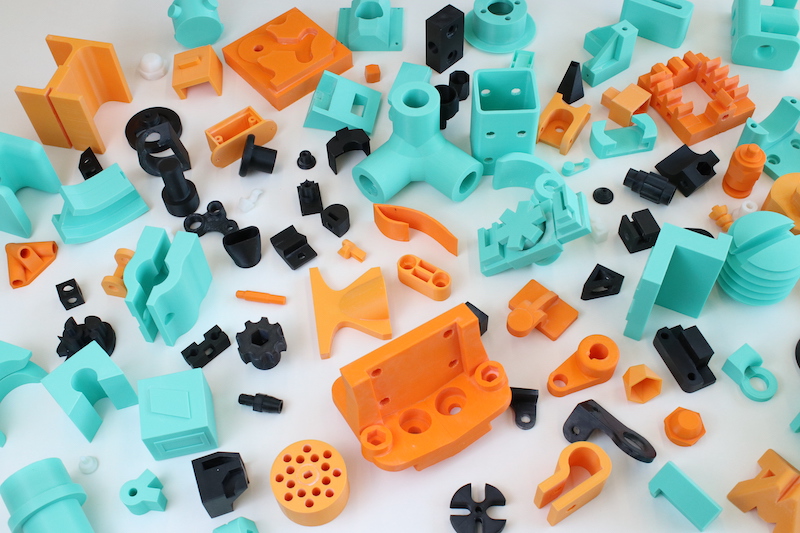}}
\caption{A photo of the thermoplastic 3D CAD models fabricated for the evaluation of our approach in a real-world setting.}
\label{fig:real_world_prints_photo}
\end{figure}

\paragraph{Fabrication.}
As we sought to fabricate a multitude of arbitrary 3D models with high geometric complexity, we opted for fabricating the models using 3D printing, as it can easily produce shapes directly from CAD models. 
We used two commodity polylactic acid (PLA) devices (Ultimaker 3 and Ultimaker S5) and considered implications of this choice (most importantly, its accuracy and layer thickness of 0.1\,mm). 
We choose the printed object size to allow acquisition with our 3D scanner at a specific sampling density of the features while simultaneously avoiding scanning any fabrication artifacts.  We pick a sampling density value $r > 0.1\,\text{mm}$ for our 3D scanner by selecting a scanning distance (see below), and compute a scaling factor $s_i$ a for each fabricated model $M_i$ individually using the relation~\eqref{eq:sampling_density}.
The fabricated CAD models are displayed in Figure~\ref{fig:real_world_prints_photo}.

\paragraph{Scanning.}
Our depth acquisition process seeks to obtain a homogeneous set of range scan data capturing most of the surface for the fabricated models and suitable for point-based and image-based training. 
We use RangeVision Spectrum~\cite{rangevision}, a commercial structured light 3D scanner, to acquire the geometry of the fabricated objects in the form of depth images.
The scanning sequence we use captures the object from two orientations w.r.t. the scanner, differing by $90\degree$; in each orientation, we take a scan every $30\degree$ using an automated turntable to minimize the operator time.
Our resulting scans are acquired from an average range of 2\,m and have the resulting sampling distance $r$ of approximately 0.5\,mm.
In total, we have acquired 1928~depth images corresponding to 166~scanning sequences of 84~unique objects.
We give more detailed statistics on our real-world dataset in the Supplemental.

\paragraph{Registration with the CAD Models.}
Our 3D scanner automatically provides an initial alignment between the obtained 3D scans; however, we found this alignment too coarse. 
Hence, we manually registered all scans to their respective CAD models using the Align Tool in MeshLab~\cite{cignoni2008meshlab} by first marking 3 points on each scan-mesh pair for rough manual alignment, followed by running the ICP algorithm for refinement.
We find that manual alignment results in significantly tighter fits.

  \section{Deep Estimation of Distance-to-Feature Fields}
\label{sec:methods}

\subsection{Learning Patch-Based Deep Estimators}
\label{methods:learning}

We train our deep regression models by solving the standard learning task: given a set of $N$ training instances, find
\begin{equation*}
\min\limits_{\theta} \frac{1}{N}\sum_i^N L(d_i, f(P_i; \theta)),
\end{equation*}
where~$d_i$ is the ground-truth distance-to-feature field for the patch~$P_i$, $f(\cdot; \theta)$ is the model with trainable parameters $\theta$, and $L$ is the loss function.
We have considered a few options for elements in this setup, to identify an optimal learning configuration.
We summarize these choices below and present the qualitative comparisons of different options in Section~\ref{exper:ablative} and their effect on method robustness in Section~\ref{exper:robustness}.

\paragraph{Network Architectures and Losses.}
Overall, we found that CNNs working with regularly resampled data outperform point-based networks for our task (Table~\ref{tab:ablation_learning_framework}). 
We require our deep models to generalize to many unseen targets with high geometric variability, thus we search for network architectures with sufficient capacity.
We use the U-Net CNN model~\cite{ronneberger2015u}, which has proven effective for image-based dense regression~\cite{xue2019learning}, and probe the ResNet family~\cite{he2016deep}, selecting the largest (ResNet-152) base network based on the quality of  predictions on the validation set.
For full details on the influence of model size on performance, we refer to Supplemental.

\begin{table}[t]
\centering
\caption{In our experiments, directly optimizing Histogram loss~\cite{imani2018improving} significantly improves performance across different quality measures.
We present results computed using the validation set of depth images (with background), with sampling distance $r_{\text{high}} = 0.02$, and noise variance $\sigma^2 = 0$.}
\resizebox{0.995\columnwidth}{!}{%
\begin{tabular}{@{}lcccc@{}}
\toprule
Loss        & \rmse$\downarrow$ &   \qrmse$\downarrow$      &   \recall$(1r)$, \%$\uparrow$     &     \fpr$(1r)$, \%$\downarrow$         \\ 
 & $\times 10^{-3}$ & $\times 10^{-3}$ & & \\  \midrule
$L_2$ (MSE) & 101.3 & 643 & 24.2 & 0.11 \\
$L_1$ (MAE) & 108.7 & 691.2 & 23.5 & 0.06 \\ \midrule
Histogram   & \textbf{61.5} & \textbf{361.1} & \textbf{57.4} & \textbf{0.06} \\ \bottomrule
\end{tabular}%
}

\label{tab:ablation_loss_type}
\end{table}

We compare three types of losses for our regression task: $L_1$ loss, $L_2$ (MSE) loss, and the Histogram loss~\cite{imani2018improving}. 
The latter one requires the model to produce a histogram of values over a predefined interval; we empirically found out that histograms with 244\,bins work best on the validation set.
Overall, we observed that learning with the Histogram loss considerably improves regression quality measured by all metrics as presented in Table~\ref{tab:ablation_loss_type}.
We attribute this to the restriction being imposed on the range of the possible target (ground-truth) distances, allowing the network to focus on a narrow range of targets. 
Our final setup with the Histogram loss predicts a confidence score for each bin in the histogram and computes the final output as a weighted sum of bin centers multiplied with their respective normalized predicted scores.

\paragraph{Additional Inputs, Supervision, and Data Volume.}
The second critical ingredient that we investigate is the dataset size and features available in training datasets. 

To assess the gains from \textit{additional inputs,} we concatenate the additional values to the point coordinates: we used the binary sharp feature point segmentation labels obtained by the non-learning algorithm VCM~\cite{merigot2010voronoi}, ground-truth normals, as well as both of these values, keeping distances as our only target variable. Neither of these additional annotations resulted in performance improvement.

To evaluate whether learning configurations for our task benefit from richer \textit{supervision} compared to distances only, we introduce additional network heads regressing either normals, normalized directions towards the nearest sharp feature line, or both simultaneously. 
During training with these targets, we optimize a multi-task loss consisting of our main loss and a weighted sum of MSE losses with weights chosen to balance the magnitude of losses: $10^{-3}$ for normals, and $10^{-2}$ for directions. 
None of these configurations led to improved regression performance either.
We also trained the network on \textit{datasets of increasing size;} we observed that performance stabilizes for datasets with more than 64,000 training instances.

In summary, the best-performing choice of architecture was a CNN U-Net with ResNet-152 backbone, trained using the Histogram loss using the supervision from ground-truth distances $d(p)$ only, on datasets of size at least 64,000. 
We present detailed results of mentioned experiments in the Supplementary material.

\begin{figure}[t]
\centerline{\includegraphics[width=0.995\columnwidth]{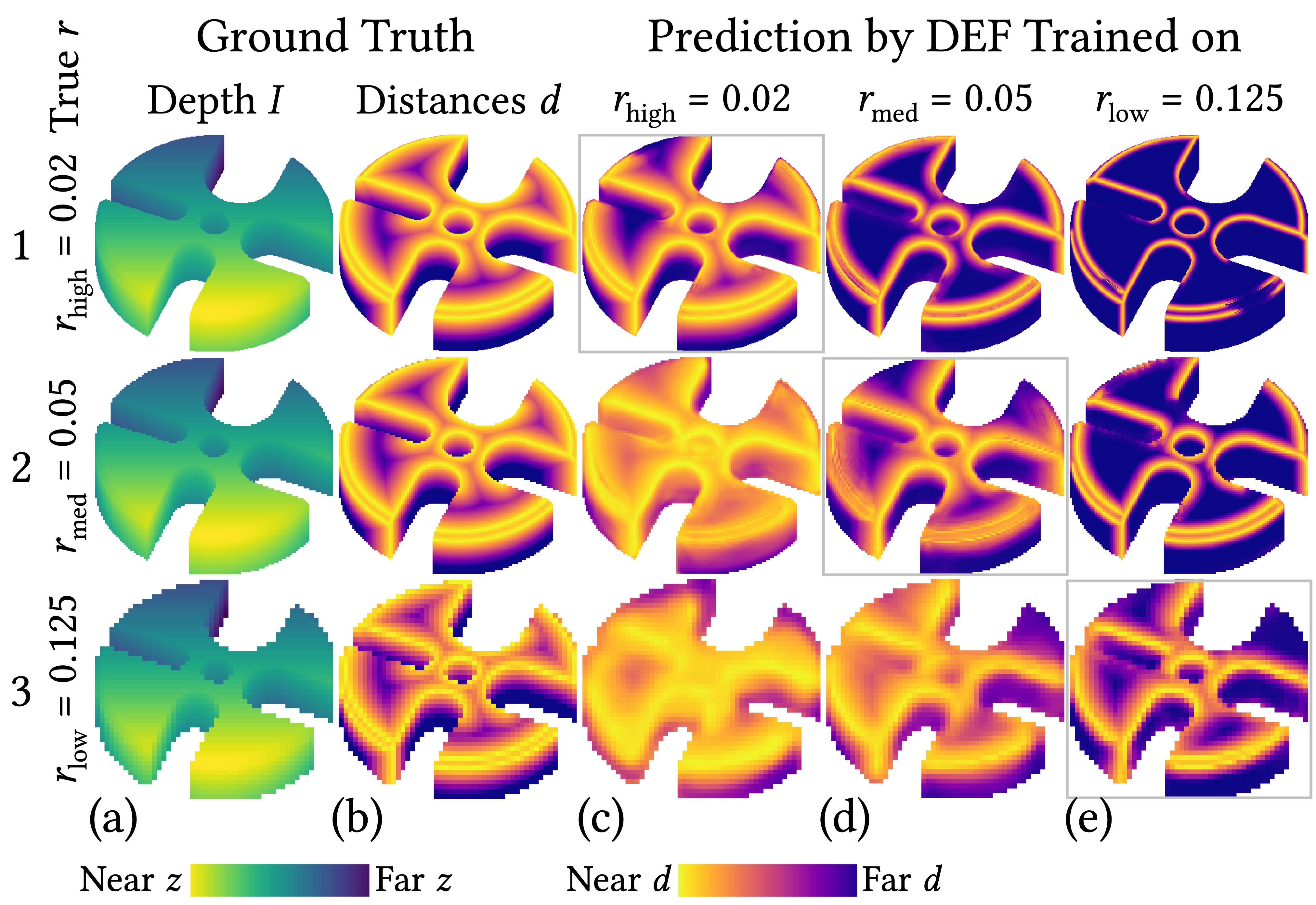}}
\caption{
Network responses to probe depth images sampled at different rates reveal high feature sensitivity and sampling robustness of our deep models; in instances with sufficient samples between feature curves, our method efficiently relates samples to respective closest feature lines.
We obtain ground-truth data (a)--(b) by raycasting a 3D model at sampling distances $r_{\text{high}} = 0.02$, $r_{\text{med}} = 0.05$, and $r_{\text{low}} = 0.125$ and produce predictions (c)--(e) using DEFs pre-trained for regressing features at $r_{\text{high}} = 0.02$, $r_{\text{med}} = 0.05$, and $r_{\text{low}} = 0.125$, respectively. 
}
\label{fig:methods_probe_images}
\end{figure}

\begin{figure*}[t]
\centerline{\includegraphics[width=\textwidth]{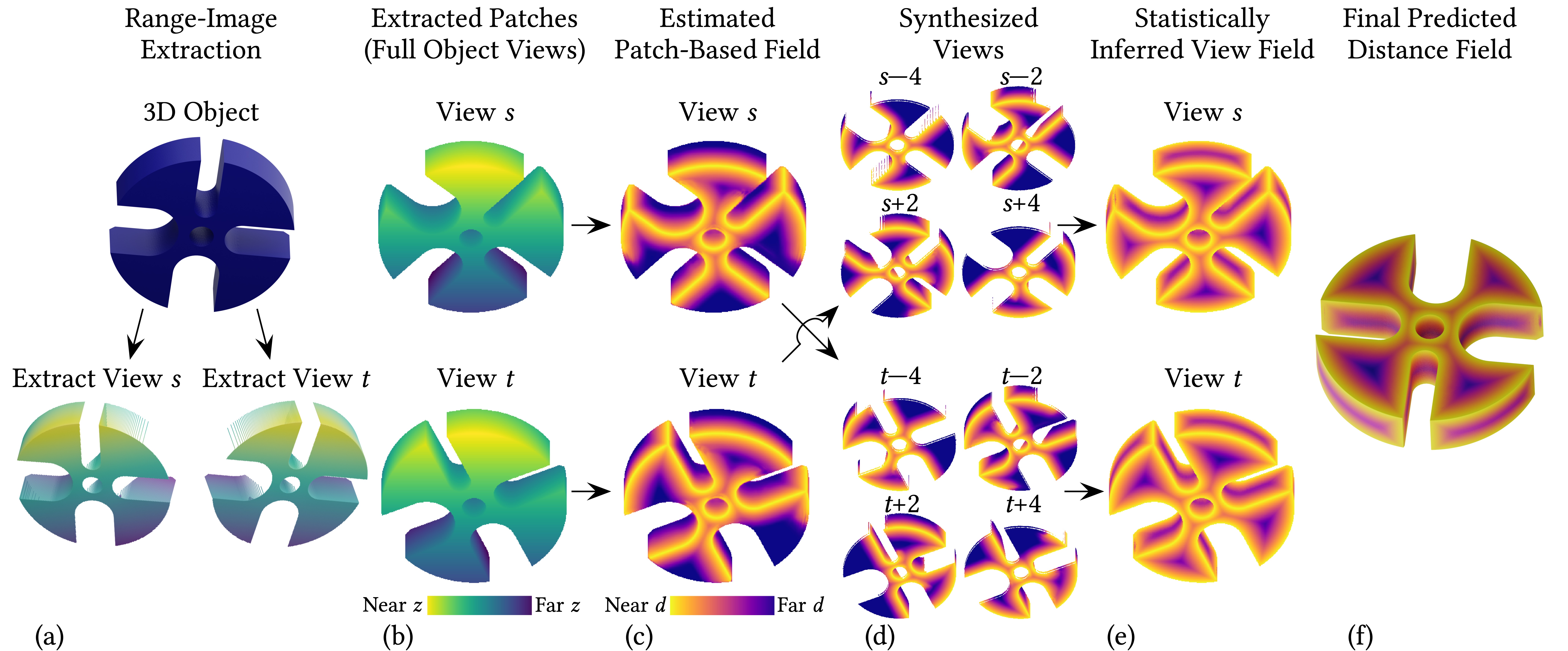}}
\caption{
Our method for reconstructing distance-to-feature fields on 3D shapes is built around postprocessing distance-to-feature predictions obtained in individual patches (or views).
(a)--(b) First, we extract a collection of overlapping patches by scanning an input shape from multiple viewing directions. 
(c) We process each patch using a DEF network to obtain patch-based predictions, sensitive to interior feature lines only.
(d) We leverage the multiple view stereopsis machinery to transfer distance-to-feature fields to adjacent views by reprojecting and linearly interpolating single-view predictions (warping-based view synthesis).
(e) The final estimate of our field on a complete 3D model is obtained by a robust statistical inference procedure.
}
\label{fig:methods_inference}
\end{figure*}

\paragraph{Feature Detection at Varying Sampling Distances.}
Each DEF network, though trained on data with a specific sampling rate~$r$ in~\eqref{eq:sampling_density}, can detect interior features sampled at significantly different rates; in
Figure~\ref{fig:methods_probe_images}, features sampled at $r_{\text{low}} = 0.125$ are robustly regressed by DEFs trained on $2.5 \times$ ($r_{\text{med}} = 0.05$) and $6.25 \times$ ($r_{\text{high}} = 0.02$) finer sampling, and vice versa.
Importantly, when sampling distance in inputs matches that of training datasets, DEF predicts a proper \emph{distance field}; otherwise, DEF produces a scale-transformed \emph{proximity field} whose iso-contours capture true features.

\subsection{Reconstructing Distance-to-Feature Fields on~Complete 3D Models}
\label{methods:fusion}

The trained deep estimators sense distance variations in the direct vicinity of the \textit{interior curves} visible in individual patches of an input shape; predictions in any two distinct patches may diverge substantially if feature curves are captured differently (\eg, a feature appears as an interior curve in one patch but shows up as a contour in another), see Figure~\ref{fig:methods_inference}~(c).
Given a set of these partial and inconsistent estimates (with known camera parameters), we reconstruct a distance-to-feature field defined globally on a complete 3D shape; we give an overview of this \textit{fusion} process in Figure~\ref{fig:methods_inference}.

\paragraph{Patch Extraction} (Figure~\ref{fig:methods_inference}~(a)--(b).)
We convert an input 3D model into a collection $\{I_i\}_{i=1}^{n_v}$ of $n_v$ range images suitable for our patch-based DEF.
We assume that the input 3D shape either already comes as range images (\eg, for range scanning) or can be resampled (\eg, represents volumetric data).
In the latter case, we obtain depth maps of the input shape from multiple distinct directions using raycasting.
As our deep models are fully convolutional, we employ \textit{full-object views} $I_i$ of input 3D models to compute predictions, which we found to perform similarly to predicting on patches of the size our network was trained on, while being more convenient.

Crucially for the completeness of the reconstruction, sufficient number of views of the input shape must be provided to capture most features; features not visible in at least one view are likely to be missed.
We observed that for all of the considered 3D shapes, using $n_v = 128$ directions is sufficient to sample more than 97\% of triangles of the corresponding meshes with at least 8\,samples; we study the influence of the number of input views in Section~\ref{exper:robustness}.
However, some shapes with many parts of their surfaces visible only from narrow cone of directions, different for each (\eg, with many deep indentations) may require many additional directions. 

\paragraph{Patch-Based Distance-to-Feature Estimation.}
Each patch~$I_i$ is processed independently using our neural network (Section~\ref{methods:learning}), yielding predictions~$\widehat{d}_i$ sensitive to interior feature curves, as shown in Figure~\ref{fig:methods_inference}~(c).

\paragraph{Transfer of Predictions across Patches.}
The aim of this stage is to gather predictions from multiple processed patches in each sampled point, integrating feature-sensitive information across the complete 3D shape.
The central idea is to employ a warping-based view synthesis mechanism (similar to~\cite{khot2019learning}): taking each pair of source and target views, we synthesize distance signal in the target view conditioned on the information inferred from the source view.
Computational complexity of our distance estimation method depends on the number of sampled points in each view and (quadratically) on the number of views~$n_v$.

Let a particular pair $(s, t)$ of source and target views be represented by depth images $I_s, I_t$, their associated intrinsic $K$ and extrinsic $T_s, T_t$ matrices, and distance-to-feature estimate $\widehat{d}_s$ available in the source view; we seek to construct a warped signal $\widehat{d}_t^{s \to t}$ from this information.
For each pixel $p = (u, v)$ in a target image~$I_t$, we compute the warped coordinates~$\widehat{p}$ in the source view by re-projecting $p$ to the image plane of $I_s$:
\begin{equation*}
\widehat{p}
    = K T_s^{-1} T_t 
    \big(
    I_t(p) \cdot K^{-1} p
    \big).
\end{equation*}
To compute the warped distance-to-feature estimate $\widehat{d}_t^{s \to t}(p)$ at the target pixel~$p$, we resample a local continuous distance field obtained by bilinearly interpolating~$\widehat{d}_s$ on the grid of samples of the source patch $I_s$ around the warped coordinates~$\widehat{p}$:
\begin{equation*}
\widehat{d}_t^{s \to t}(p) = 
  \widehat{d}_s(\widehat{p}).
\end{equation*}
We additionally compute a binary visibility mask $v_t^{s \to t}(p)$ indicating which pixels have been correctly interpolated as some pixels have insufficient number of neighbors to resample from (see Supplementary material for details).
The number of predictions for a pixel $p$ is equal to the number of depth images from which the pixel is visible. 
Example interpolation results are shown in Figure~\ref{fig:methods_inference}~(d).

As a result, each 3D sample~$p$ captured by each depth image~$I_i$ is described by a set~$D_p$ of valid predictions interpolated from all views $\{I_s\}_{s = 1}^{n_v}$:
\begin{equation}
\label{eq:valid_synthesized_preds}
D_p = 
\big\{ 
  d_s 
    | d_s = \widehat{d}_i^{s \to i}(p) 
    \text{ where } 
    v_i^{s \to i}(p) = 1 
\big\}_{s = 1}^{n_v}.
\end{equation}

\paragraph{Inference of the Final Distance Field.}
The assembled predictions are processed to form a final distance estimate by feeding the set $D_p$ into an inference set-function $g$.
We have considered a number of approaches to constructing~$g$ (we present an ablation study in Section~\ref{exper:ablative}); computing a minimum over all predictions of the distance $\widehat{d}(p) = \min \limits_{d_s \in D_p} d_s$ proved to be the most accurate among all approaches we tried, which includes computing simple, robust, or truncated averages, variants of weighting schemes, and fitting a robust locally linear regression. 
More details on computing the variants of the inference function are presented in the Supplemental.

  \section{Application: Extraction of Parametric Feature Curves}
\label{sec:parametric}

\begin{figure*}[t]
\centerline{\includegraphics[width=\textwidth,trim=20em 0 20em 0,clip=True]{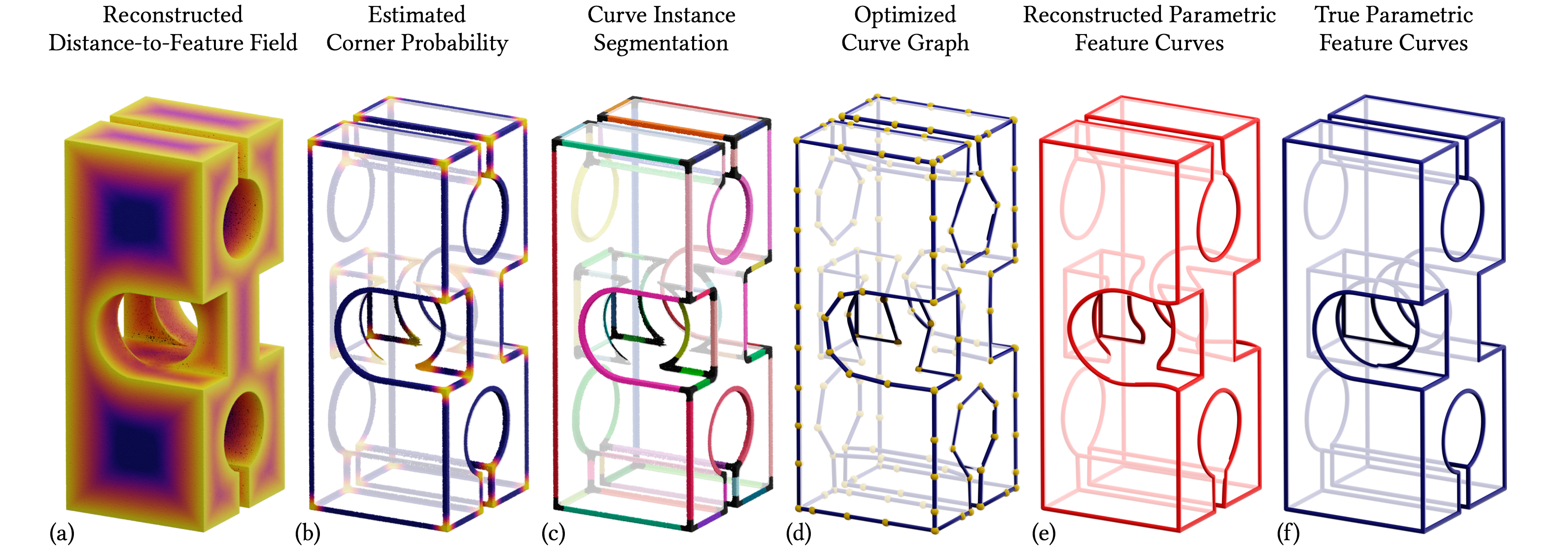}}
\caption{(a) We propose a parametric curve extraction method based on an input dense point cloud with a per-point estimated distance-to-feature field. 
We threshold distances to obtain a subset of samples $P_{\text{sharp}}$ that we use to (b) estimate corner probabilities and (c) construct curve instance segmentation (black clusters correspond to the detected corner neighborhoods).
(d) Detected corners and curves allow building and optimizing a curve graph that reflects the curve connectivity.
(e) We finally translate the curve graph into a set of accurate parametric curves that reflect feature geometry of the reference shape (f).
}
\label{fig:parametric_inference}
\end{figure*}

To evaluate the quality of distance-to-feature fields reconstructed using our method, we designed an algorithm for extracting parametric feature curve networks employing the estimated fields. 
Our algorithm is based on simple local classifiers for detecting corner vertices, heuristic graph structure analysis, and spline fitting.  
Making a number of careful choices, we are able to fit significantly more accurate feature curve networks compared to recent methods PIE-NET~\cite{wang2020pie} and PC2WF~\cite{Liu:2021:PC2WF}. 

A preliminary version of our method was presented in~\cite{matveev20213d}; similarly to the method described in this section, it uses DEF's distance-to-feature output to produce a set of feature curves.
We keep the overall structure of the approach, re-use its segmentation and spline fitting steps, and follow the same stages as in the earlier work.
However, we contribute an improved corner and curve endpoint detection criteria in~\eqref{eq:corner-grad},\,\eqref{eq:embedding}; a more robust $k$NN-based polyline construction stage and an optimization functional in~\eqref{eq:graph-opt}; a post-processing technique in~\eqref{eq:fscore}, all resulting in significant performance improvements of the method.
We refer the reader to Figure~\ref{fig:parametric_vs_wf} for qualitative demonstration of the difference between the two algorithms.

\paragraph{Initialization.}
At the initial stage, given a point cloud $P$, we select $P_{\text{sharp}}$ that consists only of points with estimated distance $\widehat d$ less than $d_{\text{sharp}}$. 
To further reduce the number of points, we apply Poisson disk sampling, leaving only 10\% of points to reduce the size of the set and make the point distribution more even.

\paragraph{Corner Detection.}
Corner detection is designed as an aggregation procedure of several corner estimates constructed from a grid of parameters.
We sample anchor points across~$P_{\text{sharp}}$ (we use 20\% of points in $P_{\text{sharp}}$ chosen by farthest point sampling) and build sets~$B_i$ of points contained in overlapping Euclidean balls of a radius $R_{\text{corner}}$ centered at the anchor points and covering $P_{\text{sharp}}$. 

We approximate each of these local sets by an ellipsoid by computing PCA on points in the set and obtain vector of variances $(\lambda_1, \lambda_2, \lambda_3)$ such that $\lambda_1 \leqslant \lambda_2 \leqslant \lambda_3$ and $\sum_{k=1}^3 \lambda_k = 1$, describing lengths of ellipsoid axes. 
For each specific set $B_i$, we use these vectors to compute a squared distance-normalized aggregate:
\begin{equation}
\label{eq:corner-grad}
    \Lambda_i = \sum_{k=1}^3\sum_{j\in \mathcal{N}_i} \left( \frac{\lambda_k^i - \lambda_k^j}{\delta_{ij}} \right) ^ 2,
\end{equation}
where $\mathcal{N}_i$ is a collection of indices of the sets $B_j$ nearest to the set $B_i$, and $\delta_{ij}$ is a Euclidean distance between anchor points of sets $B_i$ and $B_j$.  This quantity measures how much a specific ellipsoid deviates from the neighboring ones.

We decide whether a local set $B_i$ belongs to corner cluster by comparing $\Lambda_i$ against the characteristic threshold $T_{\text{variance}}$, and mark $B_i$ as either corner or curve type set:
\begin{equation}
\label{eq:corner-det}
    \begin{split}
        \mathcal{B}_{\text{corner}} &= \{B_i \, | \, \Lambda_i > T_{\text{variance}}\}, \\
        \mathcal{B}_{\text{curve}} &= \{B_i \, | \, \Lambda_i \leqslant T_{\text{variance}}\}.
    \end{split}
\end{equation}

We evaluate this classification for all combinations of 
$\mathcal{N}_i$, $T_{\text{variance}}$, and $R_{\text{corner}}$, each varying over a small range, for a total of 60 combinations, and compute a probability of $B_i$ to be a corner based on the fraction of corner classifications in this set. Refer to Section~\ref{exper:parametric} for more details.

This value is available only for the anchor points of $B_i$. 
To extend it to the whole point cloud, we apply $k$ nearest neighbors regressor with $k = 50$, thus obtaining per-point values $0 \leqslant w(p) \leqslant 1$.

The set of points near corners is obtained by thresholding weights:
\[
P_{\text{corner}} = \{p \in P_{\text{sharp}}: \, w(p) > T_{\text{corner}}\}.
\] 

\paragraph{Curve and Corner Segmentation.}
For curve segmentation, we consider the set of corner points $P_{\text{corner}}$ and the set $P_{\text{curve}} = P_{\text{sharp}} \setminus P_{\text{corner}}$ consisting of near-sharp points not detected as corners; we process both these sets to extract clusters defining individual corners and curves, respectively. 
To segment points belonging to individual curves, we construct a dense $k$NN graph by creating edges between all points in $P_{\text{curve}}$ located within sampling distance~$r$~\eqref{eq:sampling_density} from each other, and cut it into connected components. 
We treat each connected component as defining one of $n_{\text{curve}}$ curves, together they constitute the set of point clusters corresponding to each curve: 
\[
\mathcal{P}_{\text{curve}} = 
    \big\{ 
    P_c \subseteq P_{\text{curve}} \, | \, 
    \forall p \in P_c 
    \, 
    \exists q \in P_c, p \neq q:
    \|p - q\| \leqslant r
\big\}_{c=1}^{n_{\text{curve}}}.
\]
For corner points $P_{\text{corner}}$, the procedure is similar; we extract the final corner clusters $\mathcal{P}_{\text{corner}}$ by separating connected components of the detected corner sets.

\paragraph{Extraction of Curve Graph.}
From the segmentation, we construct a curve graph fitted to $P_{\text{sharp}}$,  separately processing each set of points corresponding to a curve. 
The next steps include (1) detecting endpoints for each curve, marking curves as either open or closed based on the detections, (2) approximating each curve with a short path polyline, (3) connecting fitted polylines, corners, and endpoints into a complete shape curve graph, and (4) refining endpoint and corner locations.

To detect endpoints for a segmented curve cluster~$P_c$, we construct a neighborhood-based endpoint detector similar to our corner detector. 
We construct Euclidean neighborhoods~$E_i$ with the radius~$R_{\text{endpoint}}$ centered at the anchor points~$p_{ai}$ sampled in~$P_c$, compute their straight-line approximations (we compute PCA on points in~$E_i$ and reduce its dimensionality to one), and parameterize each point $p \in E_i$ by a single coordinate $t(p)$ obtained from PCA.
To identify curve endpoints, we compute the share of points $p \in E_i$ whose parametric coordinates~$t(p)$ are greater or smaller than the parametric coordinate~$t_{ai}$ of the anchor~$p_{ai}$:
\begin{equation}
\label{eq:embedding}
    V_i = \Big\lvert\frac{1}{|E_i|}\sum_{p \in E_i} \sign (t(p) - t_{ai}) \Big\rvert,
\end{equation}
declaring $p_{ai}$ an endpoint if $V_i$ is greater than threshold $T_{\text{endpoint}}$. 
Intuitively, $V_i = 0$ corresponds to a fully symmetric case (equal shares of points parameterized by coordinates with either sign) while $V_i = 1$ indicates strong prevalence of points on either side of an anchor.
For a curve cluster~$P_c$, if only one such anchor exists, we select an anchor~$p_{ai}$ with the second largest value of $V_i$ as a second endpoint; for more than two detected endpoints, we select the two most distant ones; if no such points are detected, the curve is considered to be closed.

Next, we compute polyline approximations of curves. 
For an open curve, we construct $k$NN graph by connecting all the curve anchor points~$p_{ai}$ sampled in~$P_c$ within twice the average sampling distance from each other, and initialize the polyline with a shortest path in such graph connecting the detected endpoints.  

To create a polyline for a closed curve, we sample three points from the cluster by farthest point sampling, connect them to compose a triangle, and proceed with the subdivision strategy.
The candidate subdivision points are identified by computing
\begin{equation}
\label{eq:split}
p_{\text{split}} = \argmax_{p_i\in P_c} \big\lvert 
\widehat d_i - \|p_i - \min_l \pi^l(p_i)\| 
\big\rvert
\end{equation}
over points $p_i$ from the current curve cluster $P_c \in \mathcal{P}_{\text{curve}}$, where $\min_l \pi^l(p_i)$ is a projection of $p_i$ onto the nearest polyline segment $l$. 
To proceed with subdivision, we check an absolute difference between the estimates $\widehat d_i$ and the actual distances $\|p_i - \pi^l(p_i)\|$  against the threshold $T_{\text{split}}$; for candidate points $p_{\text{split}}$ exceeding this value, we subdivide the polyline by assigning $p_{\text{split}}$ a new polyline node and splitting the corresponding segment in two.
This choice of $p_{\text{split}}$ aims to keep the maximum polyline approximation error below $T_{\text{split}}$ for individual curves.

Finally, we substitute the detected open curve endpoints with the respective nearest corner cluster centers, yielding a final curve graph $G(q,e)$ defined by the node positions $q$ (corner cluster centers and nodes of polylines) and connections $e$ between them.
The last step is node position optimization:
\begin{equation}
\label{eq:graph-opt}
    \min_q \Big( \frac{1}{\lvert P_{\text{sharp}} \rvert} \sum_{p \in P_{\text{sharp}}} \lvert \widehat d(p) - \big\|p - \pi^{G(q,e)}(p) \big\| \rvert - \sum_{\overline q \in l\lbrack G(q,e) \rbrack} \cos{\overline q} \Big),
\end{equation}
where $\pi^G(p)$ is the projection of a point $p$ onto the nearest edge in the curve graph $G$, and $\sum_{\overline q \in l\lbrack G(q,e) \rbrack} \cos{\overline q}$ is sum of cosines of angles between the two consecutive edges incident to the node $\overline q$, computed only for the set of nodes $l\lbrack G(q,e) \rbrack$ such that they have exactly two incident edges (hence, it is locally linear).
Intuitively, the second term represents rigidity of polylines that prevents the acute angles between edges. 
Optimization helps to position graph nodes more accurately, especially at the intersections of multiple feature curves, and the rigidity term makes polyline segments more straight.
After this step is finished, we can identify the final corner positions as coordinates of graph nodes with more than two incident segments.

\begin{figure*}[t]
\centering{\includegraphics[width=0.95\textwidth]{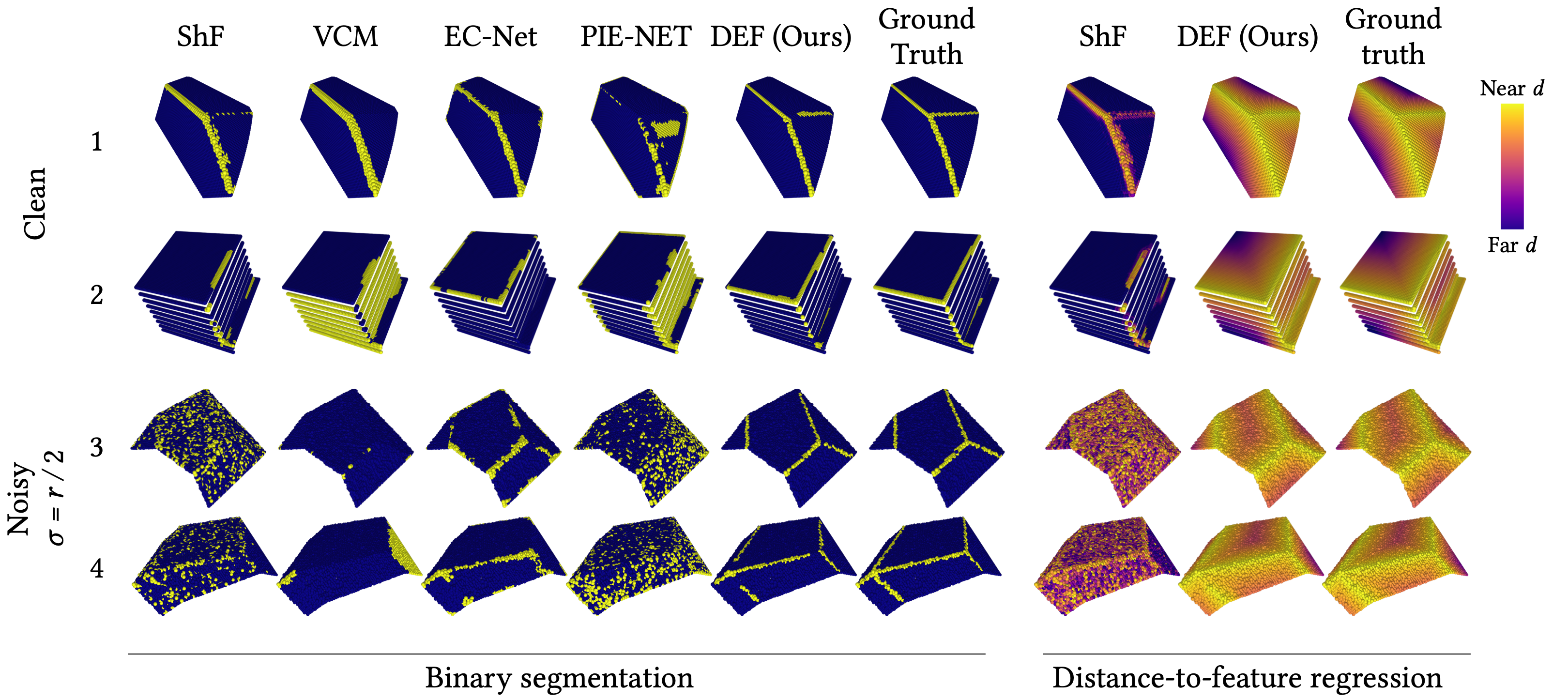}}
\caption{Visual comparison of DEF vs. competitor approaches on challenging image patch instances (synthetic image patches, $n = 50$, $r=0.02$).
Observe that, for segmentation (left part of gallery), VCM struggles to detect subtle features (rows 1, 3) and leads to substantial amounts of false positives when encountering large density variations or noisy inputs (rows 2, 4); 
EC-Net likewise tends to miss features (rows 1--2) and yield overall unstable predictions in presence of noise (rows 3--4).
Most evidently, ShF and PIE-NET deteriorate drastically in presence of noise (see rows 3--4) while producing imperfect predictions for clean data.
Additionally, PIE-NET, EC-Net, and VCM were not designed to estimate distances to nearest sharp edges (right gallery part); the only previous method for predicting distances, ShF, shows extreme sensitivity to sampling and noise (rows 1--4).
In contrast to most competitor methods, our deep models are able to accurately perform segmentation and robustly estimate distance-to-feature fields; DEF successfully survives non-uniform, irregular, or noisy sampling patterns, remaining sensitive to less pronounced features. 
}
\label{fig:synthetic_patches}
\end{figure*}

\paragraph{Spline Fitting and Optimization.}
For spline fitting one needs to obtain a consistent parameterization of each feature curve.
We do that by partitioning the curve graph into shortest paths between graph nodes with degree not equal to 2, each path serving as a proxy to a curve that defines parameter coordinates of points along feature curve.
For a path $g$ represented as a sequence of graph nodes $q_g = \{q_i\}_{i=1}^{\lvert g \rvert}$ we get a set of nearest points $P_g \in P_{\text{sharp}}$, and compute projections $\pi^{g}(p_i), \, p_i \in P_g$ and obtain values of parameters $u_g = \{u_i\}_{i=1}^{\lvert P_g \rvert}$ as a cumulative sum of norms of $\pi^{g}(p_i)$ along the path $g$.
Simultaneously, we compute knots $t_g$ as evenly spaced parameters; number of knots is defined as $\max \left(5, \frac{\lvert g \rvert}{2} \right)$.

Fitting a spline $s_g$ to the path $g$ results in a set of control points $c_s$ that define the exact shape of the spline curve. 
Once the spline is fitted, we can evaluate points $P_s(c_s) = \gamma(u_g, P_g, t_g, c_s)$ on the spline curve $s_g$. 
These points, ideally, should be precisely as far away from point cloud points $P_g$ as a distance field $\widehat d$ suggests. 
To enforce this property, we optimize over control points to shape the spline to the distance values:
\begin{equation}
\label{eq:spline-opt}
\begin{split}
    \min_c \sum_{i=1}^{\lvert P_g \rvert} \left( \widehat d_i - \|p_i - \gamma(u_i, p_i, t_i, c) \| \right) ^2,
\end{split}
\end{equation}
where $p_i \in P_g$, $\widehat d_i$ is a corresponding distance value, and $\gamma(u_i, p_i, t_i, c)$ is a point corresponding to $p_i$ evaluated on the spline $s_g$.
Additionally, we impose constraints on the spline endpoints to match the polyline endpoints.

The optimization problem and constraints are similar for the closed curves: endpoints of the spline should meet at the same point, and the tangents at the endpoint positions should be equal. 

\begin{figure*}
\centering{
\includegraphics[width=1.025\columnwidth]{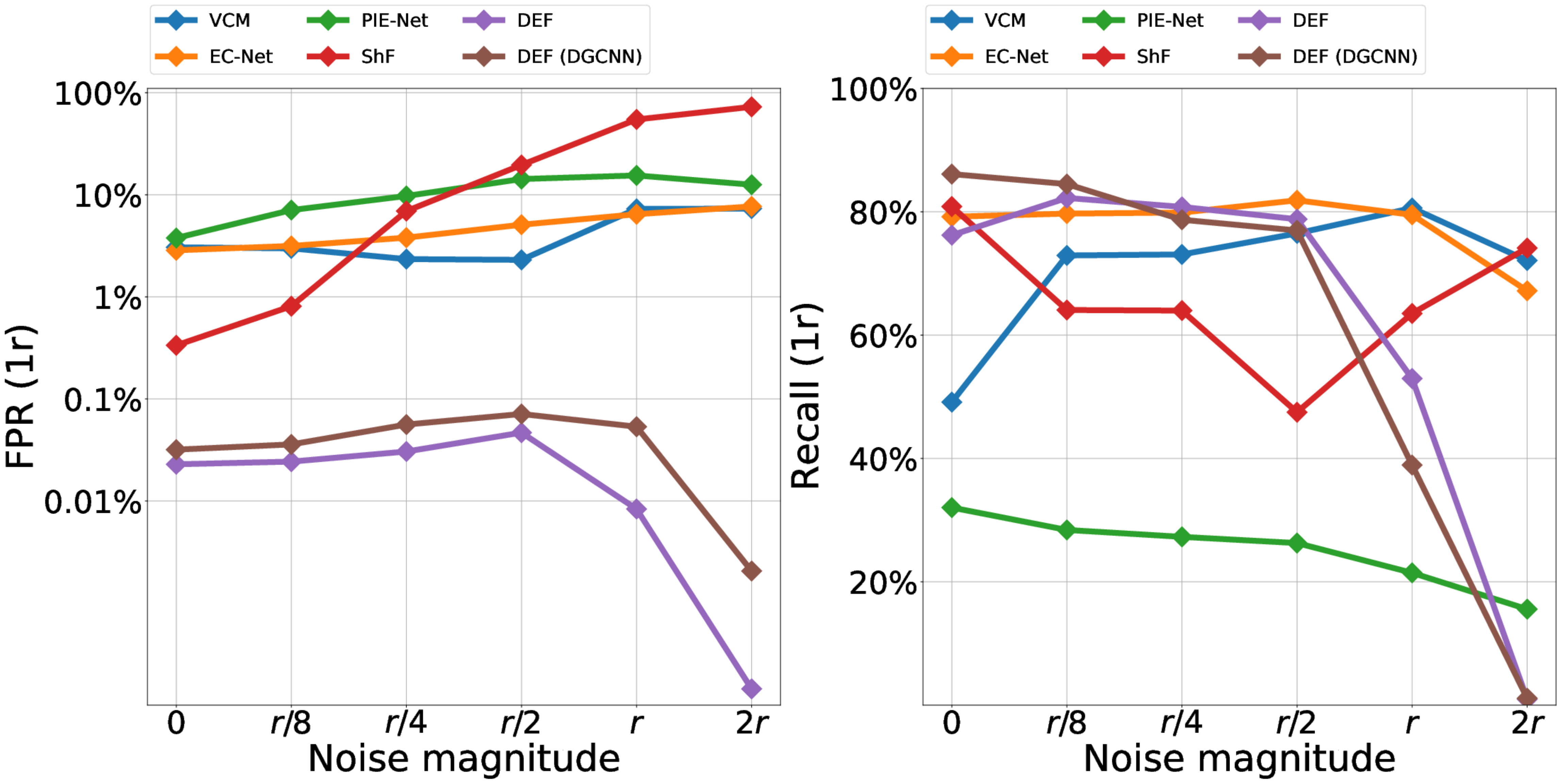}
\includegraphics[width=0.975\columnwidth]{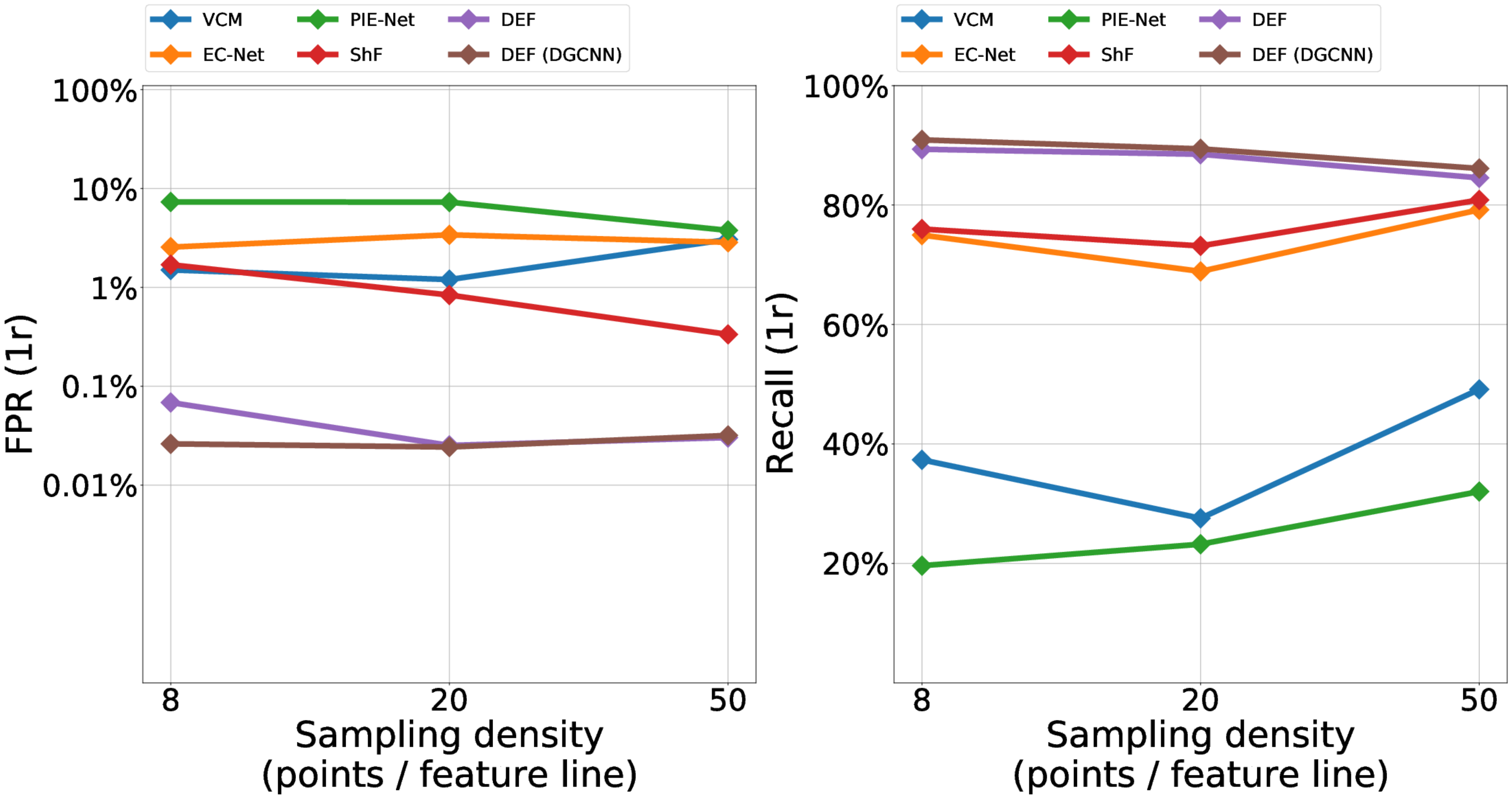}
}
\caption{DEF is significantly more robust to acquisition noise, compared to other approaches (the two left plots). Compared to the baseline approaches, DEF is robust to feature sampling density (the two right plots).}
\label{fig:noise_resolution_robustness}
\end{figure*}

\paragraph{Spline Post-Processing.}
To improve the final result, we apply post-processing procedure that helps to keep only the curves that have a good fit. 
First, we compute the quality metric as an $F_1$ score of the Chamfer distances between sampled curves and $P_{\text{sharp}}$ and vice versa, thus getting the fit quality. 
Second, we turn off each curve separately and compute the metric again. 
If the quality drops or stays the same, we keep the curve in the final set of curves. 
Otherwise, we eliminate that curve.  
The quality metric is given by:
\begin{equation}
\label{eq:fscore}
\begin{split}
    \chamfer_{X \rightarrow Y} &= \frac{1}{N_X} \sum_{x\in X} \inf_{y \in Y} \|x - y \| ^ 2,\\
    F_1(T_{\text{metric}}) &= \frac{2 \cdot \mathbbm{1}(\chamfer_{X \rightarrow Y} \leqslant T_{\text{metric}}) \cdot \mathbbm{1}(\chamfer_{Y \rightarrow X} \leqslant T_{\text{metric}})}{\mathbbm{1}(\chamfer_{X \rightarrow Y} \leqslant T_{\text{metric}}) + \mathbbm{1}(\chamfer_{Y \rightarrow X} \leqslant T_{\text{metric}})},
\end{split}
\end{equation}
where $\chamfer_{X \rightarrow Y}$ is a Chamfer distance from point set $X$ to point set $Y$, $\mathbbm{1}$ is an indicator function, and $T_{\text{metric}}$ is a threshold to convert the real-valued distances into 0-1 hard labels. 
When using this metric for post-processing, we assigned $P_{\text{sharp}}$ as one of the point sets, and a discretized set of curves as another.

Finally, we apply filtering of curves based on their length. 
This includes detecting the connected sets of curves, for each set we count the number of curves that form it and compute the total length of all curves in it. 
If the set contains less than four curves with total length smaller than $20r$, we discard such set altogether. 

Our method requires setting the following parameters: threshold on distances for selection of points near feature lines $d_{\text{sharp}}$, corner detector threshold $T_{\text{corner}}$, endpoint detector radius $R_{\text{endpoint}}$, endpoint detector threshold $T_{\text{endpoint}}$, polyline optimization threshold $T_{\text{split}}$. 
We express all of the parameters in the scale of sampling distance~$r$~\eqref{eq:sampling_density}.
We discuss the exact values of parameters in Supplementary material.

For the illustration of the vectorization pipeline and the results of our spline fitting procedure, refer to Figure~\ref{fig:parametric_inference} and Figure~\ref{fig:parametric_gallery}.

  \section{Experiments}
\label{sec:experiments}

We start our experimental study by introducing the measures of quality and providing training details in Section~\ref{exper:setup}. 
We further evaluate our models against prior art in a variety of synthetic and real-world settings in Section~\ref{exper:comparative}. 
Section~\ref{exper:parametric} demonstrates a parametric curve extraction application. 
We investigate alternative choices of model architecture and training configurations in Section~\ref{exper:ablative}. 
We conclude with testing the robustness of our approach w.r.t. sampling patterns and density variations in Section~\ref{exper:robustness}.

\subsection{Experimental Setup}
\label{exper:setup}

\paragraph{Measures of Quality.}
We evaluate our feature estimation method in terms of several quality measures (distance-to-curve regression and segmentation, as both are relevant in our case). 
We compute the following measures to assess feature estimation performance:
\begin{itemize}

    \item \textit{\rmse}: the root mean squared error between the predicted distances $\widehat{d}(p)$ and the ground-truth distance-to-feature field $d(p)$. 
    For a set of instances, we report the mean \rmse across the respective items.

    \item \textit{\qrmse}: the 95\% quantile value of \rmse across a set of instances captures the width of distance error distribution.

    \item \textit{\recall$(T)$}: we compute Recall using the predicted thresholded labels $\widehat{s}_i = \mathds{1}(\widehat{d}_i < T)$ and the ground-truth distances $s_i = \mathds{1}(d_i < T)$.
    We use $T_{\text{sim}} = r$ for synthetic instances but increase the threshold for real data to $T_{\text{scan}} = 4 r$ to account for scan misalignments.
    Recall estimates the quality of feature line estimation \textit{in the direct proximity} of the ground-truth feature line.
    As before, we report the mean value of \recall computed across test instances. 

    \item \textit{\fpr$(T)$}: we compute the False Positives Rate using the thresholded predictions and report mean \fpr across patches or full models.
    \fpr estimates the fraction of points predicted as belonging to a sharp feature line but located \textit{outside the direct proximity} of the ground-truth feature line.

    \item \textit{\chamfer}, \textit{\hausdorff} and \textit{\sinkhorn}: We use \textit{Chamfer Distance}, \textit{Hausdorff Distance} and \textit{Sinkhorn Distance}, respectively, for evaluating parametric curve extraction.
    These measures assess the discrepancy between the extracted and the ground-truth sets of curves.
    
\end{itemize}

We provide the exact formulae for our quality measures in Supplementary material. 
Unless specified otherwise, we present measure values averaged across test instances (patches or full models).

\paragraph{Data and Training.}
We train networks on 4 nVidia Tesla V100 16Gb GPUs in parallel; we use the synchronous version of batch normalization in all our architectures.
All experiments were performed using the PyTorch framework~\cite{pytorch}, its higher-level neural network API PyTorch Lightning~\cite{falcon2019pytorch}, and the Hydra framework~\cite{Yadan2019Hydra} for configuring experiments.
We use Adam optimizer~\cite{kingma2014adam} with an initial learning rate of 0.001, multiplying it by 0.9 every epoch, and train all our models with a total batch size of~32.
We validate network performance on a validation set of patches every epoch, stopping training when the \rmse metric has no improvement over the ten consecutive epochs, and select the model with the best performance on the validation set of patches.

All training patches consist of~4096 ($64 \times 64$) pixels. 
We divide depth values in each patch by the $95\%$\,quantile value computed among max depths for each patch across the training dataset; no augmentations were applied to depth images.
Unless specified otherwise, our training datasets consist of~65,536 patches.
The validation set and test set include approximately~32,000 patches. 
We observed that increasing the size of the training set further does not lead to significant improvement in performance, and report more details in the Supplementary material.

\subsection{Comparisons}
\label{exper:comparative}

\paragraph{Baseline Approaches.} 
We compare DEF against five state-of-the-art methods either directly designed or adapted for extracting feature lines from sampled 3D shapes. 
Four of these methods are deep learning-based, representing natural interest for comparisons~\cite{raina2019sharpness,yu2018ec,wang2020pie,Liu:2021:PC2WF}; the fifth method is the best-performing traditional approach based on local set-based feature detection~\cite{merigot2010voronoi} (see Section~\ref{sec:related} for more context).
We briefly review the main principles underlying these approaches below.
Most competitor methods have a number of tunable parameters, commonly adjusted to obtain the best results for a specific input shape; as we aim to compare on relatively large datasets, we determine fixed parameters that maximize method performance on the whole validation set, as explained in the Supplemental; to obtain predictions, we run each method with the selected set of its parameters on both local patches and complete point-sampled 3D shapes.

\emph{Voronoi Covariance Measure (VCM)}~\cite{merigot2010voronoi} is a non-learning method for hard segmentation of a point cloud into sharp and non-sharp points.
For this, \emph{VCM} computes the Voronoi covariance measure of a point as a covariance matrix of the intersection of an estimated Voronoi cell with a ball of radius $R$, where $R$ is a parameter of the method; a convolution radius $\rho$ is used for smoothing the measure.
The input points are labelled by thresholding the ratio of the smoothed covariance matrix's eigenvalues, with threshold $T$ being another parameter.
We have optimized the parameters $(\rho, R, T)$ to maximize $\recall(1r)$ on each dataset, by a direct search, for each data variety.
\emph{VCM} is expected to perform robustly across a range of noise and sampling variations.

\emph{Sharpness Fields (ShF)}~\cite{raina2019sharpness} is a CNN  for predicting the sharpness field --- a real-valued function with values close to~1 for points near the feature lines and 0 in smooth areas. 
To this end, \emph{ShF} constructs local neighborhoods with fixed-size ($30 \times 30$), uniformly spaced points sampled from the underlying Moving Least Squares proxy surface of the point cloud. 
The method requires normals as an additional input, that we estimate using a neighborhood-based method with the number of neighbors empirically set to 100. 
\emph{ShF} accepts a noise-free, uniformly sampled point cloud as input, thus, we expected its performance to deteriorate for noisy inputs.
We have observed that, in most cases, predicted values do not increase monotonically with distance to the feature line; however, the predicted field is suitable for producing segmentation by thresholding; we thus run a sweep to select the threshold value that would produce the highest \recall on the training set.
We also made an effort to compare our distance-to-feature field outputs to the sharpness fields produced by \emph{ShF} directly: to that end, we find the most suitable linear transformation of our field on the train subset.

\emph{Edge-Aware Consolidation Network (EC-Net)} \cite{yu2018ec} includes a PointNet++ \cite{qi2017pointnet++} derived method for detection of sharp feature lines as an auxiliary signal for point cloud upsampling.
The network predicts point locations exactly on the sharp feature curves; we map this output to our patches by selecting one nearest neighbor for each of the sharp points from \emph{EC-Net,} resulting in a hard segmentation-like output.
In our comparisons, we use the original pretrained model, that was trained on sampled patches with an additive noise, possibly making it robust to noise variations of the kind we use for evaluation.

\emph{PIE-NET}~\cite{wang2020pie} has a two-stage prediction pipeline which (1) segments sharp feature curves and corner points using a PointNet++ architecture~\cite{qi2017pointnet++} and (2) generates parametric curve proposals using a separate network, refining these using an optimization approach.
\emph{PIE-NET} expects a noise-free, uniform sample with 8,096 points representing a complete 3D shape, moreover, samples are expected to land exactly on the sharp feature lines; for these reasons, \emph{PIE-NET} is unlikely to perform robustly on most of our datasets.
We use their pre-trained models to both segment points lying in the proximity of the sharp feature curve and extract parametric curves in the form of their point samples.

\emph{PC2WF}~\cite{Liu:2021:PC2WF} is a learning-based approach to infer parametric sharp feature lines, assuming only straight lines segments are present. 
From an input point cloud, possibly noisy, \emph{PC2WF} detects corner points and infers edge segments connecting these corners; the method is able to process relatively large point sets of up to 200,000 points.
\emph{PC2WF} was not designed to detect sharp features in point clouds, so we compare the wireframe extraction quality only.
We use their pre-trained models.

\emph{Wireframes}~\cite{matveev20213d} is an earlier version of our parametric curve extraction pipeline. 
It accepts the same input as our current vectorization method, a point cloud of arbitrary size with per-point distance-to-feature estimates from DEF neural network.
Although \emph{Wireframes} share the overall structure with our current method, previous approach has major flaws in its design which we have resolved in the current method.

\begingroup
\tabcolsep=4pt
\def\arraystretch{1.075}

\begin{table}[b]
\centering
\caption{
Our \emph{local patch-based} networks for \emph{distance-to-feature estimation} and feature line \emph{segmentation} are more effective compared to competitor methods across a variety of segmentation and regression quality measures (evaluated on synthetic image patches, $n = 50, r = 0.02$).}
\resizebox{\columnwidth}{!}{%
\begin{tabular}{@{}lcccc@{}}
\toprule
Method & 
  \rmse$\downarrow$   & 
  \qrmse$\downarrow$   & 
  \recall$(1r)$, \%$\uparrow$   & 
  \fpr$(1r)$, \%$\downarrow$    \\
  & 
  $\times 10^{-3}$ & 
  $\times 10^{-3}$ & 
  & \\ 
\midrule
\multicolumn{5}{l}{\emph{Evaluation using DEF-Sim datasets}} \\
VCM~\cite{merigot2010voronoi}   & ---     &  ---  & 49.1    & 3.1   \\
EC-Net~\cite{yu2018ec}          & ---     &  ---  & 79.2    & 2.9   \\
DEF \emph{(Trained on EC data)} & 124.1   & 501.1 & 56.0    & 0.15  \\
PIE-NET~\cite{wang2020pie}      & ---     &  ---  & 32.0    & 3.8   \\
DEF \emph{(Trained on PIE data)}& 86.2    & 451.8 & 57.1    & 0.1   \\
ShF~\cite{raina2019sharpness}   & 18.0    & 95.7  &\tb{80.9}& 0.3   \\
\midrule
DEF (Ours)                      &\tb{11.1}&\tb{42.5} & 80.02 & \tb{0.02} \\ 
\midrule
\multicolumn{5}{l}{\emph{Evaluation using EC-Net datasets}} \\
DEF \emph{(Trained on EC data)} & 192.9   & 573.1   & 46.3  & 1.5 \\
DEF (Ours)                      &\tb{153.0}&\tb{526.1}& 46.4  &\tb{1.3}\\
\bottomrule
\end{tabular}
}

\label{tab:experiments_comparative_patches}
\end{table}

\endgroup

\paragraph{Patch-Based Comparison (DEF-Sim).}
We start with comparisons to prior art by evaluating DEF vs. the baselines using our synthetic \textit{patch datasets} \textit{(DEF-Sim)} to provide a direct network-to-network comparison and eliminate the influence of postprocessing.
We present a statistical evaluation in Table~\ref{tab:experiments_comparative_patches}, compare results visually in Figure~\ref{fig:synthetic_patches}, and plot dependencies of performance vs. noise and resolution parameters for all methods in Figure~\ref{fig:noise_resolution_robustness}.

\begin{figure*}
\centering{\includegraphics[width=0.975\linewidth,keepaspectratio]{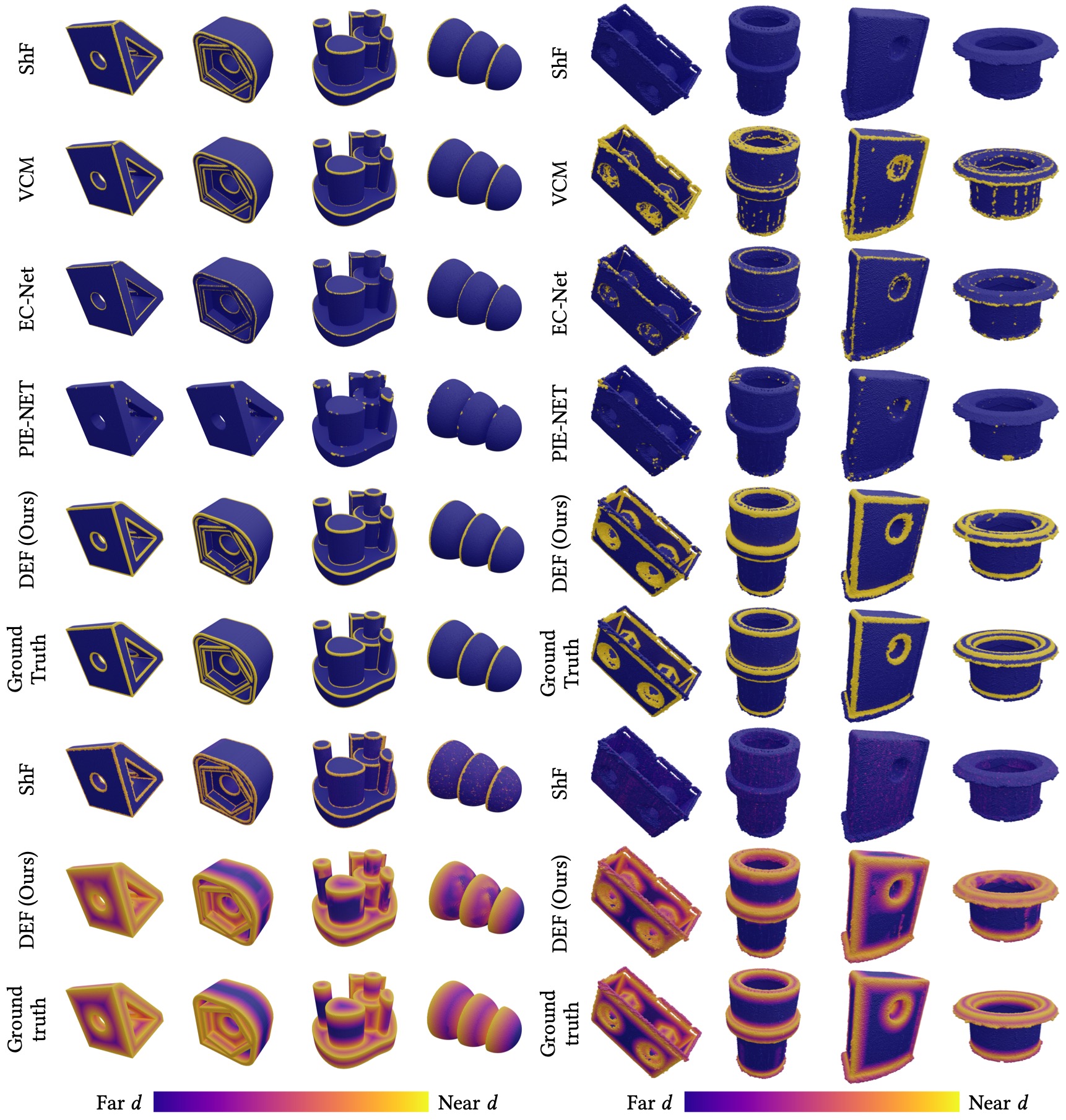}}
\caption{
Comparison to state-of-the-art sharp feature line estimation methods on high-resolution synthetic full shape datasets (a) and real scanned datasets representing full 3D shapes (b). 
Our method is able to robustly reconstruct a pointwise distance-to-feature field and scales to 3D shapes represented by millions of points.}
\label{fig:complete_3d_models}
\end{figure*}

Qualitatively, we observe that our method compares favorably to all competitors (most evidently, \emph{ShF,} \emph{VCM,} and \emph{PIE-NET}) on less pronounced features that have smaller normal jumps (Figure~\ref{fig:synthetic_patches}, rows 1,3); while these methods tend to be less sensitive to such subtle features, DEF demonstrates increased robustness when facing such geometry.
For instances with large sampling distance variations (Figure~\ref{fig:synthetic_patches}, row 2), \emph{ShF} and \emph{EC-Net} miss features while \emph{VCM} and \emph{PIE-NET} produce substantial numbers of false positive, particularly in under-sampled regions; for \emph{VCM,} this is due to the uniform surface sampling assumed in the model; DEF remains capable of accurately localizing feature locations.
In comparison with \emph{ShF} and \emph{PIE-NET,} DEF performs notably better on noisy data for noise magnitudes of up to $r / 2$, with a moderate decrease in \recall but almost no change in \fpr, compared to two orders of magnitude increase in \fpr from 0.33\% to 19\% for \emph{ShF} (Figure~\ref{fig:noise_resolution_robustness}, left two plots). 
This leads to the results of these methods being unusable for noisy point clouds, see Figure~\ref{fig:synthetic_patches}; however, such results are expected as \emph{ShF} and \emph{PIE-NET} models that we used were not optimized on noisy datasets.
For varying sampling distance values, DEF still compares favorably according to \recall and \fpr measures (Figure~\ref{fig:noise_resolution_robustness}, right two plots). 

We made an effort to train our algorithm using the datasets described in~\cite{yu2018ec,wang2020pie} to ensure conformity in terms of training sets and input-output requirements.
For the EC-Net dataset, we use the original 32\,mesh files and feature annotations; to create a PIE-NET-like dataset, we select meshes with up to 30,000 vertices containing only Line, Circle, or BSpline curves; in each case, we generate a dataset of 65,536 images for training our method using the pipeline from Sections~\ref{datasets:design}--\ref{datasets:synthetic}.
We present results in Table~\ref{tab:experiments_comparative_patches}. 
Evaluation using DEF-Sim datasets indicate that our method performs significantly better than \emph{PIE-NET}; compared to \emph{EC-Net}, our network keeps having $10\times$ lower \fpr but delivers less accurate distance predictions; this is likely due to a low geometric diversity of training data: the volume of the \emph{EC-Net} dataset is two orders of magnitude lower compared to our datasets.

\begingroup
\tabcolsep=4pt
\def\arraystretch{1.075}

\begin{table}[]
\centering
\caption{Our method is able to reconstruct a robust estimate of a distance-to-feature field defined for a complete 3D shape. 
While DEF achieves similar \recall to \emph{VCM}, it does so by truncating an accurate distance field and demonstrates nearly $10\times$ lower \fpr.\\
$^*$ \emph{PIE-NET} was invoked with 8,096 samples as input.}
\resizebox{\columnwidth}{!}{%
\begin{tabular}{@{}lcccc@{}}
\toprule
Method & 
  \rmse$\downarrow$   & 
  \qrmse$\downarrow$   & 
  \recall$(1r)$, \%$\uparrow$   & 
  \fpr$(1r)$, \%$\downarrow$    \\
  & 
  $\times 10^{-3}$ & 
  $\times 10^{-3}$ & 
  & \\ 
\midrule
VCM~\cite{merigot2010voronoi} & --- &  ---  & \textbf{79.2} & 4.8 \\
EC-Net~\cite{yu2018ec}        & --- &  ---  & 48.5 & \textbf{0.2} \\
PIE-NET$^*$~\cite{wang2020pie}    & --- &  ---  & 73.6 & 2.9 \\
ShF~\cite{raina2019sharpness} & 623 & 761.4 & 69.8 & 0.3 \\
\midrule
DEF (Ours)     & \textbf{115.1} & \textbf{200.1} & 79.0 & 0.5 \\
\bottomrule
\end{tabular}
}

\label{tab:synthetic_complete_models}
\end{table}

\endgroup

\paragraph{Complete 3D Models (DEF-Sim).}
To obtain results on complete models, we use DEF-Sim, the synthetic validation set of \ndefsimodels~sampled 3D shapes (see Section~\ref{datasets:design}), and apply our patch-based DEF to each view of each shape without any fine-tuning on these data.
We further reconstruct a complete, object-level distance-to-feature field using the algorithm described in Section~\ref{methods:fusion}; for our fusion, we use $n_v = 128$ views and perform view synthesis in orthographic projection using 4 neighbors for each sampled point.
To obtain the final statistical estimate, we extract minimum value from the set of valid interpolated predictions in~\eqref{eq:valid_synthesized_preds}.

We compare our approach with competitors statistically in Table~\ref{tab:synthetic_complete_models} and visually in Figure~\ref{fig:complete_3d_models} (a).
Most our complete 3D shapes include from~$10^6$ to~$10^7$ point samples.
Qualitatively, our method is able to more robustly regress features with smaller difference in normal orientations, undersampled features, or feature curves with large density variations across the feature line, such as features in internal cavities of a 3D shape.

\begin{figure}[b]
\centerline{\includegraphics[width=0.9\columnwidth]{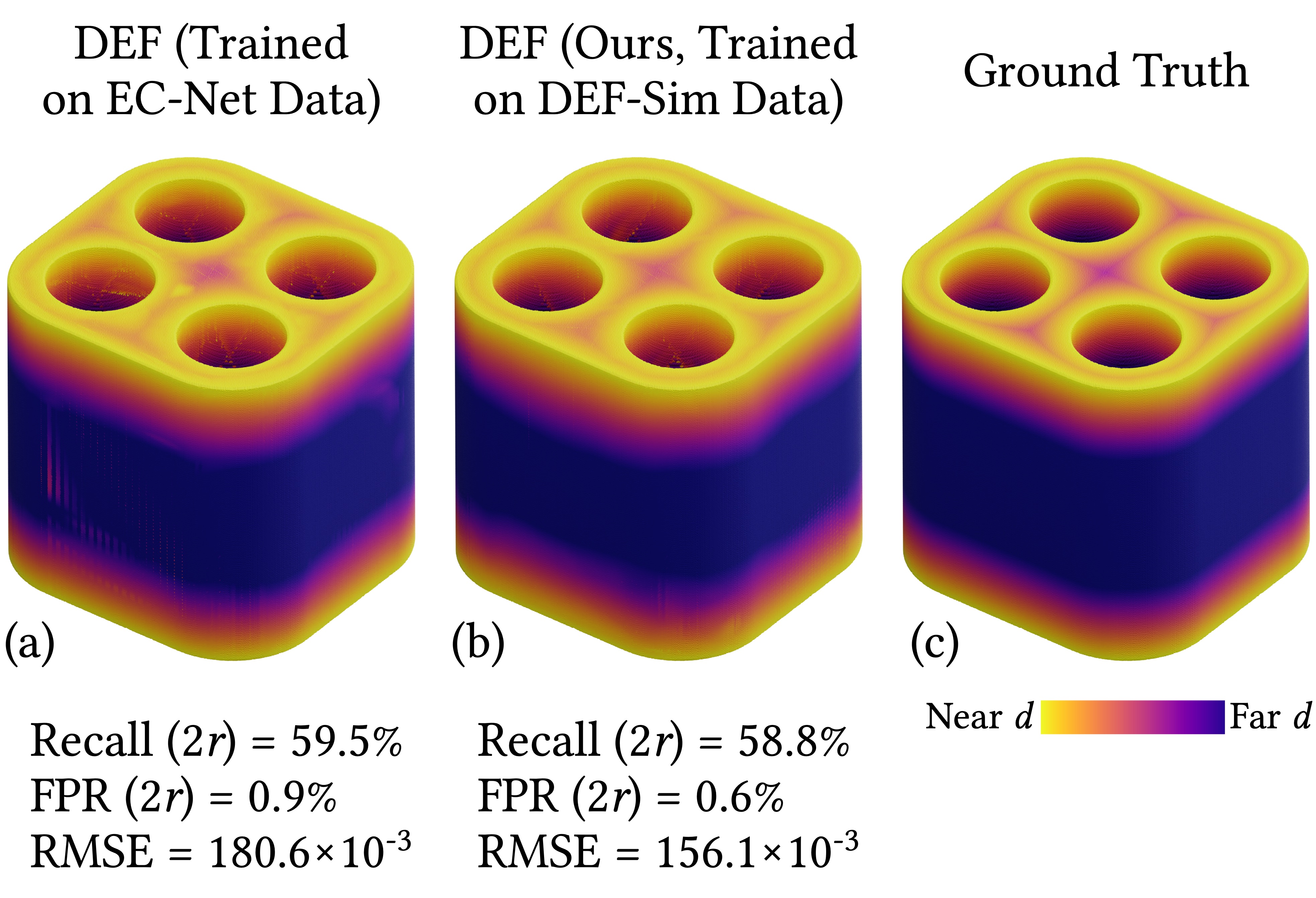}}
\caption{Our method is able to leverage various feature-annotated training collections. 
A complete object-level field then can be reconstructed from predictions by a model pre-trained on (a) the \emph{EC-Net} dataset~\cite{yu2018ec} and (b) our DEF-Sim dataset (see Section~\ref{exper:comparative}).
As our data is two orders of magnitude larger in size, predictions obtained using our model are generally more accurate.}
\label{fig:experiments_comparative_ecnet_data}
\end{figure}

In Figure~\ref{fig:experiments_comparative_ecnet_data}, we additionally demonstrate an example reconstruction of a complete object-level distance field using DEF trained on patches in the EC-Net dataset described above.

\begingroup
\tabcolsep=4pt
\def\arraystretch{1.075}

\begin{table}[t]
\centering
\caption{Compared to the closest state-of-the-art competitor approach, \emph{VCM}, our method achieves $3 \times$ higher \recall$(4r)$ on noisy and incomplete scanned data, while maintaining a moderate \fpr$(4r)$. 
Quantitatively, our method reconstructs the full distance-to-feature field with $\rmse = 1.5\,\text{mm}$ and \qrmse = 2.9\,\text{mm} at a sampling distance of $r = 0.5\,\text{mm}$.}
\begin{tabular}{@{}lcc@{}}
\toprule
Method & 
  \recall(2\,mm), \%$\uparrow$   & 
  \fpr(2\,mm), \%$\downarrow$    \\
\midrule
VCM~\cite{merigot2010voronoi} &         29.5  &          10.2 \\
EC-Net~\cite{yu2018ec}        &         10.1  &  \textbf{0.8} \\
\midrule
DEF (Ours)                    & \textbf{91.7} &          20.1 \\
\bottomrule
\end{tabular}

\label{tab:real_complete_models}
\end{table}

\endgroup

\begin{figure*}[t]
\centering{\includegraphics[width=0.975\textwidth,trim=0 0 30em 0,clip=True]{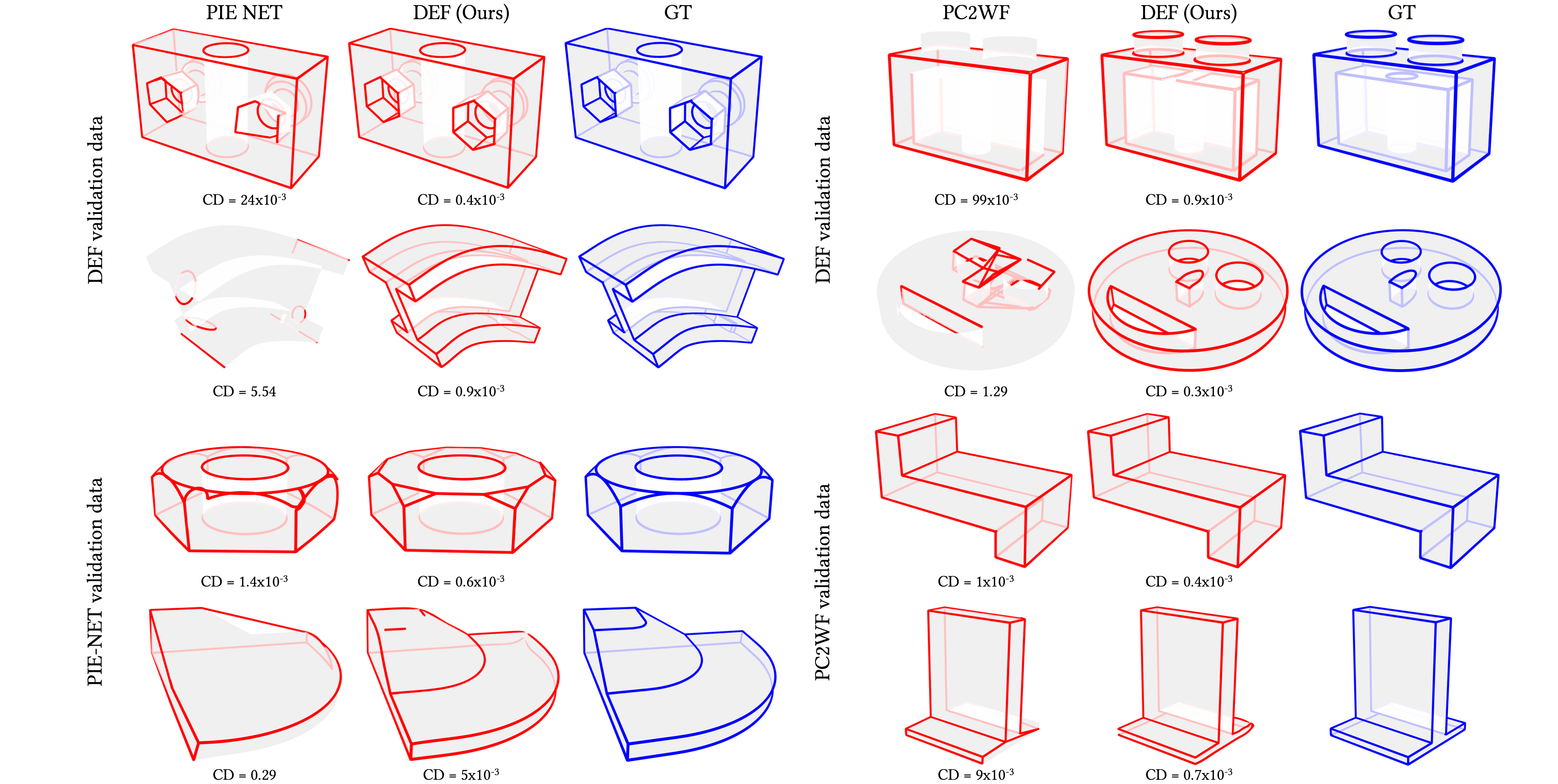}}
\caption{We use distance field estimates obtained by our method for complete, large sampled shapes (up to $10^7$ points) to reconstruct full parametrizations of their feature curves.
We compare our inference results to PIE-NET (a) and PC2WF (b) using our validation set (rows 1--2) and on validation shapes from the corresponding papers (rows 3--4).}
\label{fig:parametric_comparison}
\end{figure*}

\paragraph{Real 3D Shapes (DEF-Scan).}
To perform an experimental evaluation of distance-to-feature prediction quality for real-world noisy 3D scans, we use our real-world dataset of complete 3D scanned shapes with sharp feature annotations. 
We first select a DEF CNN model pre-trained on a synthetic dataset (with sampling distance $r_{\text{med}} = 0.05$) and fine-tune it using the real annotated depth images. 
To this end, we split the 84~scanned objects into training (42~objects, 981~scans), validation (21~shapes, 479~scans), and final testing (21~objects, 468~scans) subsets, and optimize our model until convergence on the validation set. 
Next, we apply the optimized network to each view of the testing dataset and reconstruct a complete distance-to-feature field using our fusion algorithm (Section~\ref{methods:fusion}) using $n_v = 12$ views available for each 3D shape; here we perform view synthesis in perspective projection using 4 neighbors for each sampled point.

Overall, our method reconstructs the complete distance field with $\rmse = 1.5\,\text{mm}$ and \qrmse = 2.9\,\text{mm}.
We report performance against competitor approaches in Table~\ref{tab:real_complete_models} using \recall$(4r)$ and \fpr$(4r)$ measures where the real-world sampling distance $r = 0.4\,\text{mm}$. 
Compared to \emph{VCM} and \emph{EC-Net,} our results suggest that DEF systematically outperforms the competitor methods by a significant margin (\eg, DEF achieves $3\times$ higher \recall compared to the best-performing competitor method, \emph{VCM}); the methods \emph{ShF} and \emph{PIE-NET} produced little to no sharpness detections for all shapes that we have used.
These observations are also reflected in qualitative results in Figure~\ref{fig:complete_3d_models} (b).

\subsection{Extracting Parametric Curves}
\label{exper:parametric}

We run our vectorization method on the complete 3D shapes sampled using $n_v = 128$ views, where predictions have been computed by the DEF network and a complete object-level distance field has been obtained in the previous steps (Section~\ref{exper:comparative}). 
After setting parameters, we run our method without manual intervention. 
The output consists of (1) spline curve parameters and (2) endpoint coordinates for straight lines, readily available for further processing.

\begingroup
\tabcolsep=4pt
\def\arraystretch{1.075}

\begin{table}[b]
\centering
\caption{Compared to \emph{PIE-NET} parametric feature curve extraction stage, DEF achieves an order of magnitude more accurate reconstruction.}
\begin{tabular}{@{}lccc@{}}
\toprule
Method & 
  \chamfer$\downarrow$   & 
  \hausdorff$\downarrow$   & 
  \sinkhorn$\downarrow$    \\
\midrule
PIE-NET~\cite{wang2020pie} &  0.97  &  2.19  & 0.84 \\
\midrule
DEF (Ours)     & 
  \textbf{0.04} & 
  \textbf{0.55} & 
  \textbf{0.05} \\ 
\bottomrule
\end{tabular}

\label{tab:parametric_quality_measures}
\end{table}

\endgroup

\emph{PIE-NET}~\cite{wang2020pie} requires subsampling our point clouds to 8,096 points. 
We applied the farthest point sampling technique to reduce the size of the point clouds. 
\emph{PIE-NET} parametric curves extraction stage produces a set of points sampled along the curves. 

\emph{PC2WF}~\cite{Liu:2021:PC2WF} is essentially free of point cloud size; however, to reduce the computation time and make sampling density closer to the point clouds of the original paper, we subsampled our shapes to 200,000 points each. 
\emph{PC2WF} outputs pairs of endpoint coordinates that represent a straight line wireframe.

\emph{Wireframes}~\cite{matveev20213d} has the same input and output as our method.

\begin{figure*}[t]
\centering{\includegraphics[width=0.95\textwidth,trim=0 0 30em 0,clip=True]{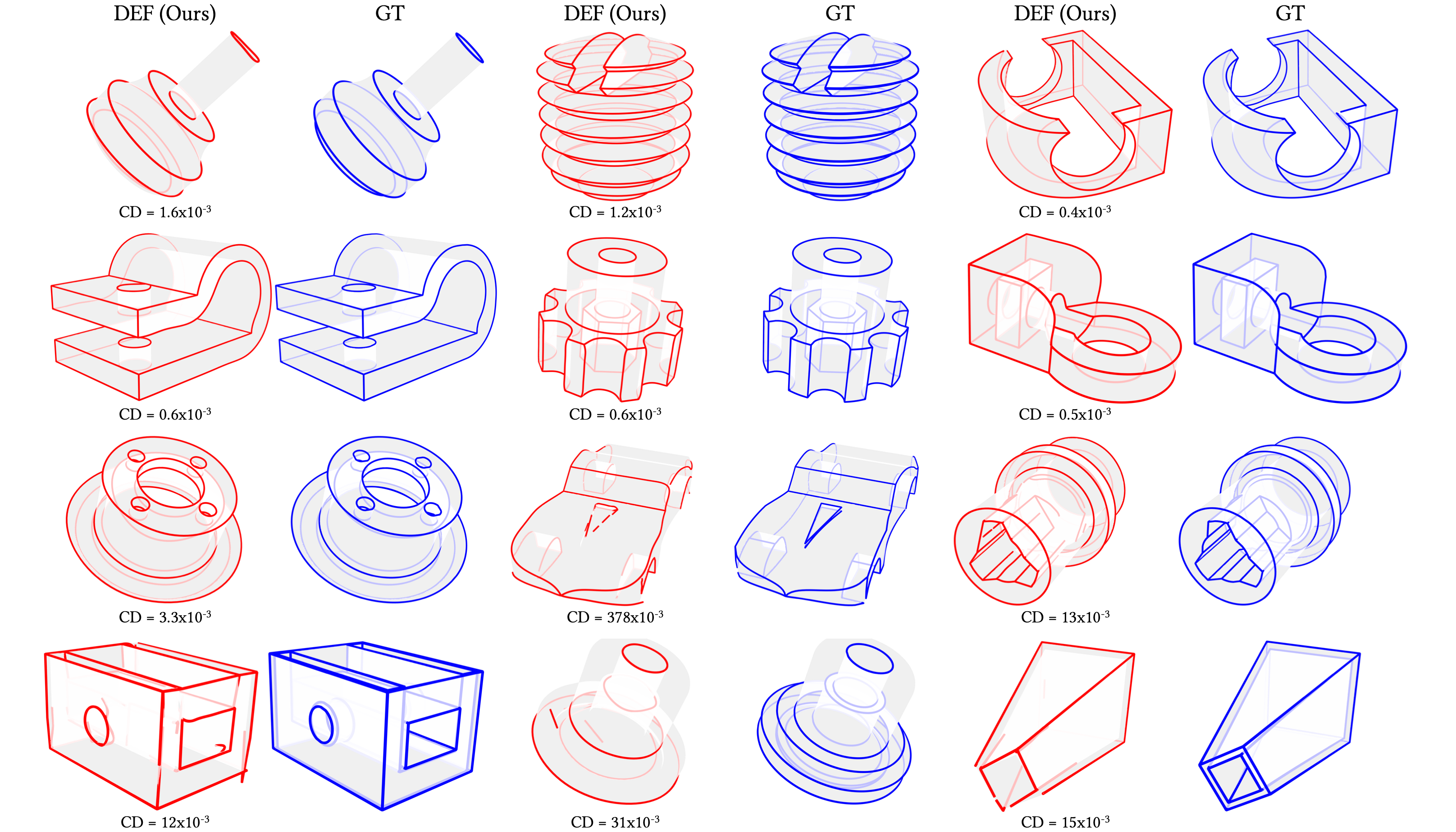}}
\caption{We showcase twelve additional examples of extracted parametric representations next to the ground truth sets of curves.
Row~4 includes visually inferior examples where our method struggles to output clean and complete parametric representation.}
\label{fig:parametric_gallery}
\end{figure*}

To assess the wireframe quality, we ran our pipeline on the validation set of \ndefsimodels{} complete 3D models (DEF-Sim) along with \emph{PIE-NET} and compared the obtained results to the ground truth parametric curves. 
To compute the metrics, we sampled all the predicted curves and lines along with the ground truth set of curves into point sets and derived distances between the closest points to calculate \chamfer, \hausdorff, and \sinkhorn. 
The aggregated statistical estimation of metrics for our method and \emph{PIE-NET} are reported in Table~\ref{tab:parametric_quality_measures}. 
We observed a significant difference between one-sided \chamfer's for \emph{PIE-NET} predictions. 
Specifically, the average distance from ground truth to prediction is 0.9, the average distance from prediction to ground truth is 0.064. 
That implies that \emph{PIE-NET} misses many curve instances, but it outputs relatively accurate reconstructions for the detected ones. 
In turn, the one-sided \chamfer of our method is 0.024 from ground truth to prediction and 0.02 for distance in the opposite direction. 
We refer the reader to Figure~\ref{fig:parametric_comparison} for the qualitative results.

Since \emph{PC2WF} outputs straight lines only, we did not run it on the whole set of validation shapes and report no statistical performance; instead, we provide qualitative results for their method only on the small subset of shapes presented in Figure~\ref{fig:parametric_comparison}.

For both \emph{PIE-NET} and \emph{PC2WF}, qualitative results depict the shapes from our validation set and figures from the respective papers that were used to evaluate the quality of the corresponding methods.

\begin{figure}[b]
\centering{
\includegraphics[width=\columnwidth,keepaspectratio,trim=10em 0 30em 0,clip=True]{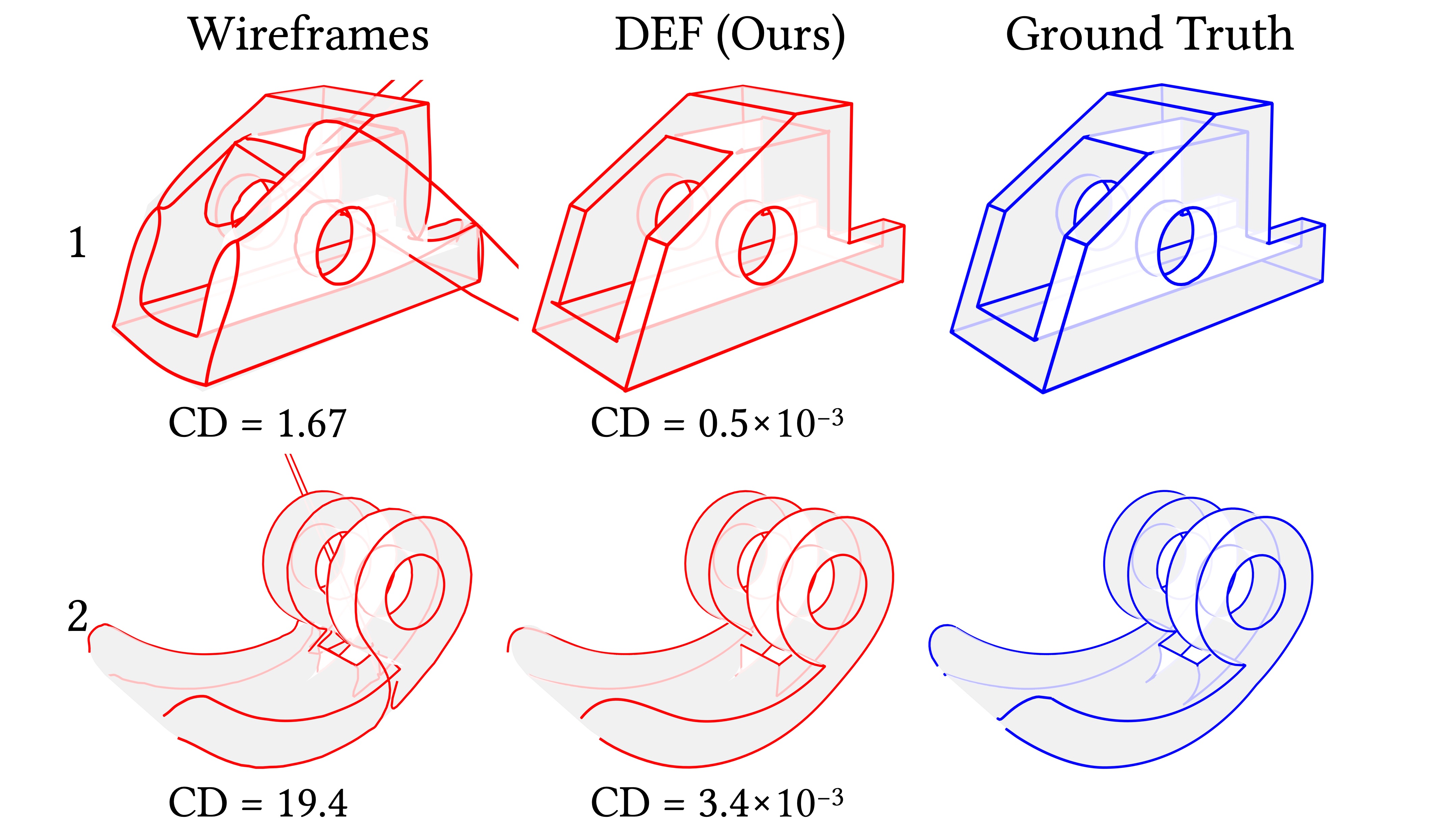}
}
\caption{Our current pipeline improves corner detection (row 1) and is able to resolve complex curves (row 2), whereas \emph{Wireframes} outputs imprecise curve graphs that lead to outlier curves with extreme variation.}
\label{fig:parametric_vs_wf}
\end{figure}

Results indicate that our method is more flexible and robust with respect to the shape sampling variation and geometric complexity. 
Compared to \emph{PIE-NET}, DEF detects more curve instances, and due to the predicted distance field, the fitting procedure does not rely solely on the point positions and is free of sampling issues. 
Our pipeline can fit curves of different types when \emph{PC2WF} has been designed for straight lines.
On the other hand, the performance of our method is strongly conditioned by the choice of parameters when both \emph{PIE-NET} and \emph{PC2WF}, as learning-based methods, are almost free of parameter tuning. 
We described a simple tuning procedure that only exploits the distance field estimation to mitigate that. 

Additionally, we demonstrate how our current vectorization pipeline compares to the previous version (\emph{Wireframes}). 
We compare the two methods in Figure~\ref{fig:parametric_vs_wf}.
The improved corner detection and $k$NN-based polyline construction enable our method to resolve cases of close corners and complex curves.
Curve graph topology guides the curve fitting stage and, if imprecise, may lead to outlier curves as it is seen in the \emph{Wireframes} output.

\subsection{Ablation Studies}
\label{exper:ablative}

\begingroup
\tabcolsep=4pt
\def\arraystretch{1.075}

\begin{table*}[]
\centering
\caption{Compared to point-based DGCNN~\cite{wang2019dynamic}, our CNN-based learning method more efficiently regresses distance-to-feature values. 
For image-sampled patches that tend to be non-uniform, adding prior sharpness estimates from \emph{VCM} yields no advantage to either method.}
\resizebox{0.9\textwidth}{!}{%
\begin{tabular}{@{}llccccc@{}}
\toprule
Dataset &
  Method & 
  \rmse$\downarrow$   & 
  \qrmse$\downarrow$   & 
  \recall$(1r)$, \%$\uparrow$   & 
  \fpr$(1r)$, \%$\downarrow$ \\
   & &
  $\times 10^{-3}$ & 
  $\times 10^{-3}$ & 
  & \\ 
\midrule
Regular images (no bg, reprojected to points) &
  DGCNN + Histogram loss & 
  11.3  & 
  55.5 & 
  80.9 & 
  $3.7 \times 10^{-2}$ \\
Regular images (no bg, reprojected to points) &
  DGCNN + Histogram loss + VCM & 
  13.6  & 
  70.0  & 
  68.8  & 
  $4.8 \times 10^{-2}$ \\
Regular images (no bg) & 
  CNN + Histogram loss (DEF) &
  \textbf{9.7} &
  \textbf{32.5} &
  \textbf{84.6} &
  $\mathbf{3 \times 10^{-2}}$ \\
Regular images (no bg) & 
  CNN + Histogram loss + VCM &
  10.9 &
  36.8 &
  80.4 &
  $3.7 \times 10^{-2}$ \\
\midrule
Regular images (with bg, DEF-Sim)  & 
  CNN + Histogram loss (DEF)  & 
  11.1 & 
  42.5 & 
  80.0 & 
  $2.2 \times 10^{-2}$ \\
\bottomrule
\end{tabular}
}

\label{tab:ablation_learning_framework}
\end{table*}

\endgroup

We conducted a large number of computational experiments to determine the optimal parameters of our method; our main conclusions were outlined in Section~\ref{sec:methods}; here, we summarize the results of the studies supporting these conclusions. 
We present a separate stress-test to explore the robustness of our approach in Section~\ref{exper:robustness}.

\noindent \paragraph{Learning Architectures.}
In this paper, our focus is on 3D data represented as a collection of depth images, one of the most common types of scanned 3D data. 
this allows us to use standard convolutional networks that take advantage of the regular sampling pattern in the data.  
To quantify the advantage obtained from using this additional regularity of sampling, we consider an alternative approach, ignoring depth image structure, and viewing the collection of images  as an unstructured point set.
As standard CNNs can no longer be applied to this type of data, we use the DGCNN network~\cite{wang2019dynamic}; we set depth $D = 6$ and width $W = 64 \times 1.35^{D - 3} \approx 150$.
Similarly to the CNN version, we trained the network using the Histogram loss, studying various modifications, most importantly, training the DGCNN using the ground-truth distances $d(p)$ \emph{and} VCM sharpness labels as an additional input.

\begin{figure}[b]
\centering{\includegraphics[width=\columnwidth]{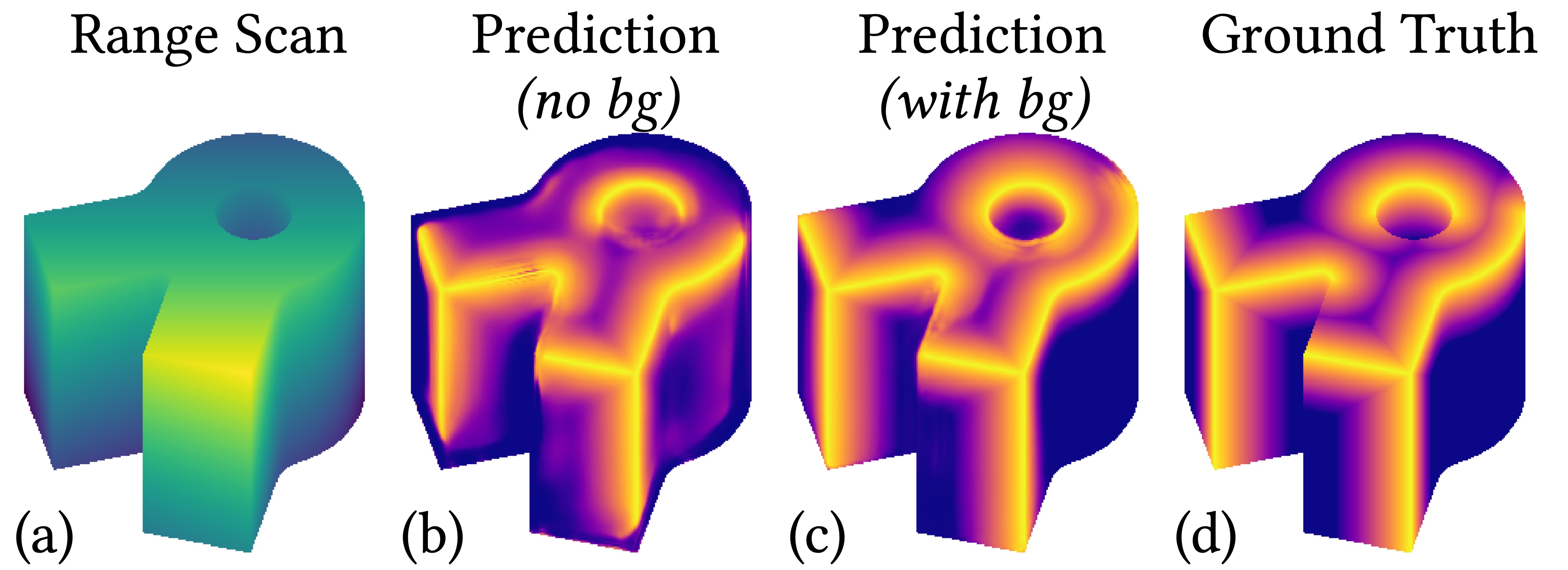}}
\caption{We opt for training on instances with background and depth discontinuities (\textit{with bg,}~(c)); excluding these  (\textit{no bg,}~(b)) yields suboptimal predictions, particularly near patch boundaries.}
\label{fig:experiments_ablative_effect_background}
\end{figure}

For highly non-uniform image-sampled patches (\eg, rays passing nearly in parallel to parts of the surface), VCM struggles to extract feature-related information. Thus, adding VCM labels yields no advantage for range-scan data for both the DGCNN and the CNN DEF models.
Generally, we observe DEF networks to outperform point-based models (DGCNN trained with Histogram loss supervised by $d(p)$ \emph{and} VCM) on regularly sampled range-scan data, see Table~\ref{tab:ablation_learning_framework}, middle rows.
CNN DEF models additionally demonstrate better noise-resistance compared to the point-based alternative, as can be seen in Figure~\ref{fig:noise_resolution_robustness}.
In this experiment, we train CNN DEF and DGCNN models on noisy sampled data, and find that the latter yields lower \recall and higher \fpr values across noise magnitudes. 

\noindent \paragraph{Data Generation.}
We mention an additional configuration of interest, obtained by considering two versions of the range-scan data: a filtered version that excludes patches with depth discontinuities or background pixels (we refer to it as \emph{no bg}), and a dataset including all types of patches (referred to as \emph{with bg}); we train models separately on either data variety.
DEF models trained on patch datasets without background pixels perform quantitatively better for similar testing data, see Table~\ref{tab:ablation_learning_framework}, bottom rows; however, as shown in Figure~\ref{fig:experiments_ablative_effect_background}, networks trained on data with background pixels yield more stable predictions, particularly on near-boundary pixels.

\begin{figure}[b]
\centering{\includegraphics[width=\columnwidth,keepaspectratio]{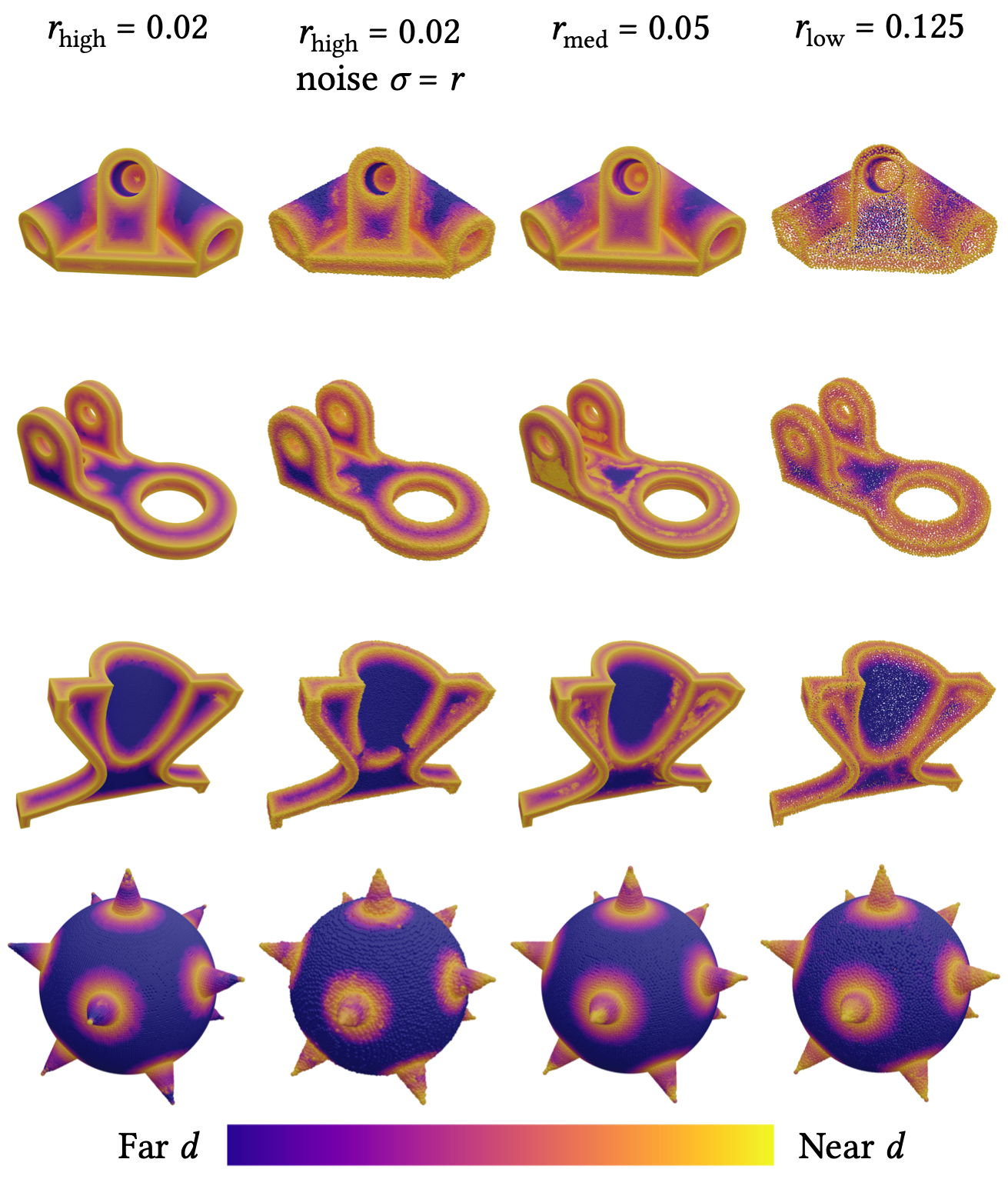}}
\caption{Our approach is able to withstand (b) high noise magnitudes and (a), (c), (d) large variations in sampling density.}
\label{fig:noise_robustness_examples}
\end{figure}

\noindent \paragraph{Loss Type} (Section~\ref{methods:learning}).
The results of our study of loss functions lead us to find the Histogram loss~\cite{imani2018improving} to perform favorably compared to $L_1/L_2$ losses (see Table~\ref{tab:ablation_loss_type}).

\noindent \paragraph{Reconstruction on Complete 3D Models.}
We investigate the two crucial factors in the reconstruction of distance-to-feature fields om complete sampled 3D shapes: the number of views $n_v$ and the inference function applied over the set $D_p$ of interpolated predictions in~\eqref{eq:valid_synthesized_preds}.
To this end, we consider an order of magnitude fewer set of $n_v = 18$ views and two additional inference functions: \emph{truncated min} and \emph{linear fit,} as well as compare against an aggregation method applied on top of DGCNN predictions.
\emph{Truncated min} is computed by removing 20\% of smallest values in $D_p$ and taking min; \emph{linear fit} fits a robust version of local linear regression~\cite{huber1973robust} to $d(p)$ in each sampled point $p$ by extracting local patches of Euclidean neighbors of size 50, and computes the final estimate as a fitted value in~$p$.

Statistical results for our sets of \ndefsimodels~synthetic and \ndefscanmodels~real scanned models are presented in Table \ref{tab:ablation_fusion}.
We focus our attention on the \recall and \rmse measures and conclude that having a sufficient number of views is crucial to the successful reconstruction of our distance function.
Comparisons of inference functions generally lead to \emph{truncated min} improving over \rmse but not \recall measure compared to \emph{min}, with \emph{linear fit} being inferior to both these approaches.

\begingroup
\tabcolsep=4pt
\def\arraystretch{1.075}

\begin{table*}[]
\centering
\caption{We demonstrate quantitative results of reconstructing distance-to-feature fields on complete 3D models using variations of our approach. 
For both DEF-Sim and DEF-Scan collections, we find a significantly better \recall being achieved by \emph{min} fusion, while \rmse favors \emph{truncated min}.}
\resizebox{0.95\textwidth}{!}{%
\begin{tabular}{@{}llcccccc@{}}
\toprule
Dataset   &
Method    & 
  \rmse, $\downarrow$    & 
  \qrmse, $\downarrow$   & 
  \multicolumn{2}{c}{\recall$(T)$,\,\%$\uparrow$} &
  \multicolumn{2}{c}{\fpr$(T)$,\,\%$\downarrow$} \\
  & 
  &
  $\times 10^{-3}$ & 
  $\times 10^{-3}$ & 
  $T=1r$ & 
  $T=4r$ & 
  $T=1r$ & 
  $T=4r$ \\ 
\midrule
DEF-Sim (crops)  &
DGCNN + Histogram loss ($n_v = 18$, min)  &
  247.6 & 
  287.9 & 
  52.4  & 
  92.3  &
  0.2   &
  2     \\
DEF-Sim  &
  DEF ($n_v = 18$, linear fit)  & 
  255.1 & 
  351.6 & 
  0     & 
  3.1   &
  0     &
  0     \\
DEF-Sim  &
  DEF ($n_v = 18$, truncated min)  & 
  120.8 & 
  227.4 & 
  12.5  & 
  74.9  &
  0     &
  0.7   \\
DEF-Sim  &
  DEF ($n_v = 18$, min)  & 
  100.2 & 
  214.1 & 
  47.9  & 
  92.3  &
  0.2   &
  2     \\
DEF-Sim  &
  DEF ($n_v = 128$, truncated min)  & 
  \textbf{62.4}  & 
  \textbf{157.1} & 
  31.8  & 
  90.9  &
  0     &
  1     \\
DEF-Sim  &
  DEF ($n_v = 128$, min)  & 
  115.1 & 
  200.1 & 
  \textbf{79}    & 
  \textbf{98}    &
  0.5   &
  5.3   \\
\midrule
Dataset   &
Method    & 
  \rmse,\,mm\,$\downarrow$    & 
  \qrmse,\,mm\,$\downarrow$   & 
  $T = 0.5\,\text{mm}$        &
  $T = 2\,\text{mm}$          &
  $T = 0.5\,\text{mm}$        &
  $T = 2\,\text{mm}$          \\
\midrule
DEF-Scan  &
  DEF ($n_v = 12$, linear fit)  & 
  1.27  & 
  2.36  & 
  ---   &
  70.1  & 
  ---   &
  \textbf{7.9} \\
DEF-Scan  &
  DEF ($n_v = 12$, truncated min)  & 
  \textbf{1.25}  & 
  \textbf{2.3}   & 
  ---   &
  80.9  & 
  ---   &
  9.5 \\
DEF-Scan  &
  DEF ($n_v = 12$, min)  & 
  1.54  & 
  2.85  & 
  ---   &
  \textbf{91.7}  & 
  ---   &
  20.1 \\
\bottomrule
\end{tabular}
}

\label{tab:ablation_fusion}
\end{table*}

\endgroup

\subsection{Robustness Study}
\label{exper:robustness}

\begin{figure}[t]
\centering
\includegraphics[width=\columnwidth]{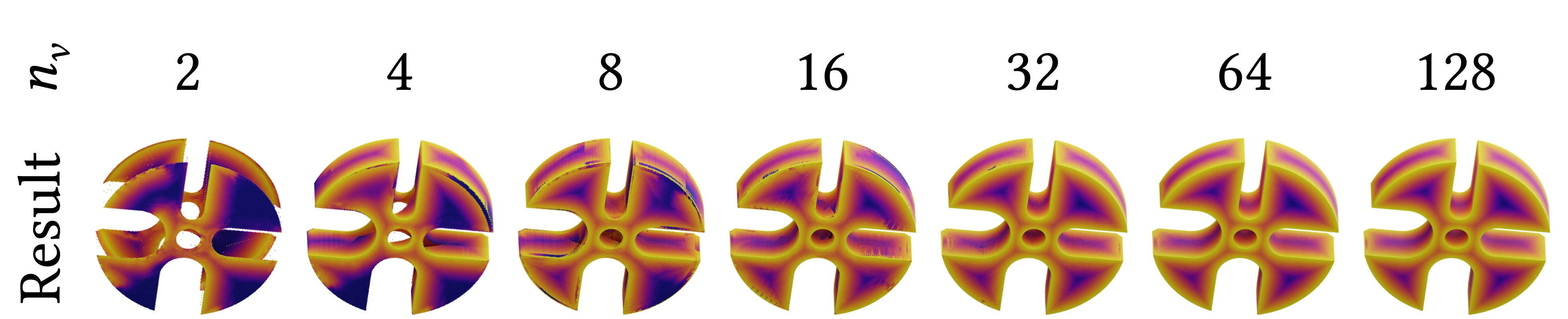}
\caption{We experimentally observe our method to benefit from increasing the number of views used during fusion. 
For this synthetic shape, $n_v = 18$ projections give an approximate \recall of 90\%.}
\label{fig:experiments_robustness_instance_n_views}
\end{figure}

\paragraph{Noise and Sampling Sensitivity.}
We examine the noise sensitivity of our method by training DEF CNNs on datasets with increasing noise levels and coarse sampling, and using these in reconstructing distance fields on complete 3D models. 
We vary the noise magnitude from $0$ up to $2r$, where $r$ is sampling distance. 
Performance of the networks in isolation drops moderately as noise magnitude rises, as seen in Figure~\ref{fig:noise_resolution_robustness}; the models show particular robustness to sampling distance variations, indicating weak influence of sampling on performance.
Figure~\ref{fig:noise_robustness_examples} demonstrates qualitative reconstruction results for a number of 3D shapes sampled in a variety of ways; note that overall prediction stays stable across various setups.

\begin{figure}[t]
\centerline{
\includegraphics[width=\columnwidth]
  {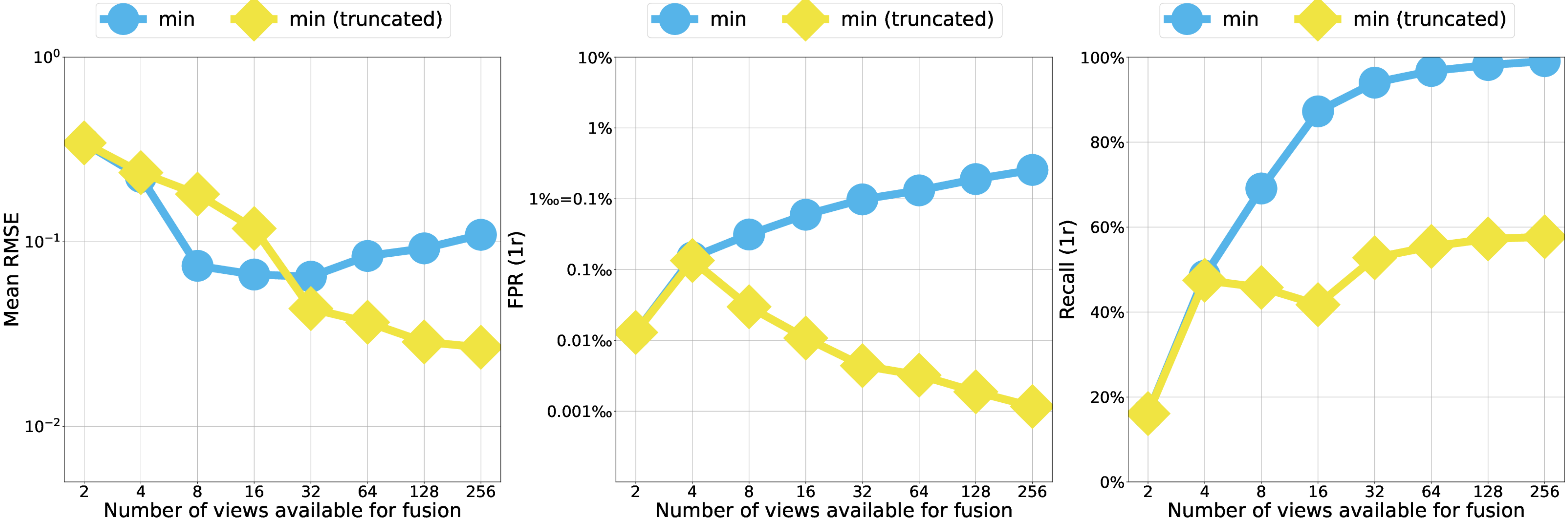}
}
\caption{Qualitatively, reconstructing distance-to-feature field on a complete 3D shape is able to detect the vast majority of features with around $n_v = 16$ views; increasing the number of views to $n_v = 32, 64$, or $128$ refines and stabilizes these detections.}
\label{fig:experiments_robustness_metrics_n_views}
\end{figure}

\paragraph{Sensitivity to Number of Views.}
We investigate how the performance depends on the number of available views; for this experiment, we take 1024~views following a geodesic spiral around the object, and perform fusion using $n_v = 2, 4, 8, 16, 64, 128, 256$ views.
We present qualitative reconstruction results in Figure~\ref{fig:experiments_robustness_instance_n_views} and demonstrate performance dynamics in Figure~\ref{fig:experiments_robustness_metrics_n_views}.
We observe a clear benefit from increasing the number of views, and achieve \recall of approximately~90\% with 16~views. 
The dynamics of \rmse and \recall/\fpr measures indicate different statistical effects for \emph{min} vs. \emph{truncated min} inference function in~\eqref{eq:valid_synthesized_preds}.
More specifically, while \emph{min} provides superior \recall, it stagnates on \rmse as more data are added, not representing correctly the true distance-to-feature field.
In contrast, \emph{truncated min} is able to continue improving both \rmse and \fpr measures, but shows saturation of \recall as smallest values are being cutoff from the set $D_p$ in~\eqref{eq:valid_synthesized_preds}.

  \section{Conclusions}
\label{sec:conclusions}

We presented a new learning-based pipeline for automatic sharp feature detection from sampled 3D data. 
Our approach is based on training and comparing different methods on a dataset annotated with distance-to-feature information derived from the ABC dataset of 3D CAD models. 
Our method works on patches sampled from the input shape, with predictions combined in a postprocessing step.

We demonstrate that the CNN-based model operating on regularly sampled range images, when such images are available as an input or via resampling the input, is an efficient predictor for distance-to-feature fields.
The image-based CNN model is also the most robust to input noise in our experiments.
A somewhat surprising observation is that training a regression model benefits from using a histogram loss.
At the same time, providing additional inputs, or including additional outputs in training, did not lead to significant improvements in accuracy either for image- or for point-based networks, except adding VCM as input to DGCNN. 

We compared our results to recent learning-based methods and a representative high-quality traditional method, demonstrating quantitative and qualitative improvements over these approaches.
For instance, the proposed DEF outperforms the best-performing approach by 4\% in terms of Recall measure while offering an order of magnitude improvement in false positives rate (from 0.3\% to 0.03\%).
Our method generalizes to real data after fine-tuning; we are not aware of any other feature estimation approach tested on a large collection of real data with manually annotated ground truth.
Our approach also scales to orders of magnitude larger point clouds, which has not been successfully shown before.

We make publicly available the two collections of datasets, the benchmarks, the implementation of all baselines, the reference implementation of our method, and our trained models to foster additional work in this direction.

  \section{Limitations and Future Work}

Limitations of our approach to feature estimation include 

\begin{enumerate}

\item \emph{Feature Definition.} 
Our definition of sharp geometric features depends on a relatively large $18^{\circ}$ normals angle threshold (normals inner product $\approx 0.95$).
However, for arbitrarily-oriented normals (\eg, the original ABC data~\cite{koch2019abc}), we use the absolute of the inner product, and our annotations do not reflect very sharp edges (\ie, those having normals whose inner product is larger than 0.95); this special case remains an open issue.

\item \emph{Data Annotation Procedure.} 
For complex geometry (\eg, folded shapes, shapes with rich geometric detail in internal cavities), our distance-to-feature annotations may produce spurious signal on flat surfaces due to feature curves that are close in Euclidean (but not geodesic) sense; we exclude such data from training.
In such instances, using \textit{geodesic} instead of \textit{local Euclidean} distances is more appropriate.

\item \textit{Visibility and Cross-View Consistency.}
Dependence on feature visibility can be viewed as a limitation of our approach; however, for common real data acquired by scanners, only visible features are present.
We eliminate inconsistency in per-view predictions in each 3D surface point by obtaining multiple likely distance-to-feature values, then statistically inferring a final value (\eg, by taking min).

\item \textit{Feature Ambiguity.}
Sufficiently dense sampling of nearby features is a crucial requirement for our algorithm to accurately distinguish individual features. 
In instances where having enough (\eg, 8 or more) samples between feature curves is possible, our method efficiently relates samples to respective closest feature lines; otherwise, close feature curves may cause incorrect clustering of points. 

\item \textit{Parametric Curve Extraction.}
Limitations of our vectorization method mainly stem from the quality of the extracted distance-to-feature field. 
For instances with varying sampling density or unstable distance values, our method may struggle with distinguishing close curves or concentric circles (see, \eg, Figure~\ref{fig:parametric_gallery}, row~4).
A partly related effect is gluing together two close corners (see, \eg, Figure~\ref{fig:parametric_comparison}, row~4).

\end{enumerate}

Future Work in the direction of our research may include

\begin{enumerate}

\item \textit{Extending to Features of Multiple Types.} 
We have used interior curves in all training examples \textit{on patches,} however we hypothesize that training with boundary (contour) curves \textit{on whole shapes} or \textit{patches with boundary,} \ie, distinguishing different feature types, might be beneficial.

\item \textit{Reconstruction of a Complete Distance Field.} 
Our procedure for inferring distance-to-feature fields on complete 3D shapes is agnostic to the type of function that it reconstructs; at the same time, our distance-to-feature is a non-negative, piecewise-linear, bounded function; incorporating such forms of explicit prior knowledge about this function can considerably improve prediction accuracy.

\item \textit{Real-World Prediction.}
We believe that extending our preliminary study of feature estimation in scanned 3D shapes to a full, robust algorithm capable of vectorizing real-world scans represents a promising research direction.

\end{enumerate}
  
\begin{acks}
We are grateful to Prof. Dzmitry Tsetserukou (Skoltech) and his laboratory staff for providing the 3D printing device and technical support. We thank Sebastian Koch (Technical University of Berlin), Timofey Glukhikh (Skoltech) and Teseo Schneider (New York University) for providing assistance in data generation. We also thank Maria Taktasheva (Skoltech) for assistance in computational experiments. We acknowledge the use of computational resources of the Skoltech CDISE supercomputer Zhores for obtaining the results presented in this paper~\cite{zacharov2019zhores}. The work was supported by the Analytical center under the \grantsponsor{1}{RF Government}{} (subsidy agreement 000000D730321P5Q0002, Grant No. \grantnum{1}{70-2021-00145 02.11.2021}).

\end{acks}
  
  \clearpage
\appendix
\noindent{\Huge DEF:~Deep~Estimation~of~Sharp}\\ {\Huge Geometric~Features~in~3D~Shapes}\\ {\Huge Supplementary~Material}
\section{Details on Training and Evaluation Datasets}
\label{supp:detail-datasets}

\subsection{Details on Datasets Construction}
\label{supp:detail-datasets-construction}

\begin{figure}[b!]
\centerline{
\includegraphics[
width=0.9\columnwidth,
trim=10em 2em 10em 10em,
clip=True]
{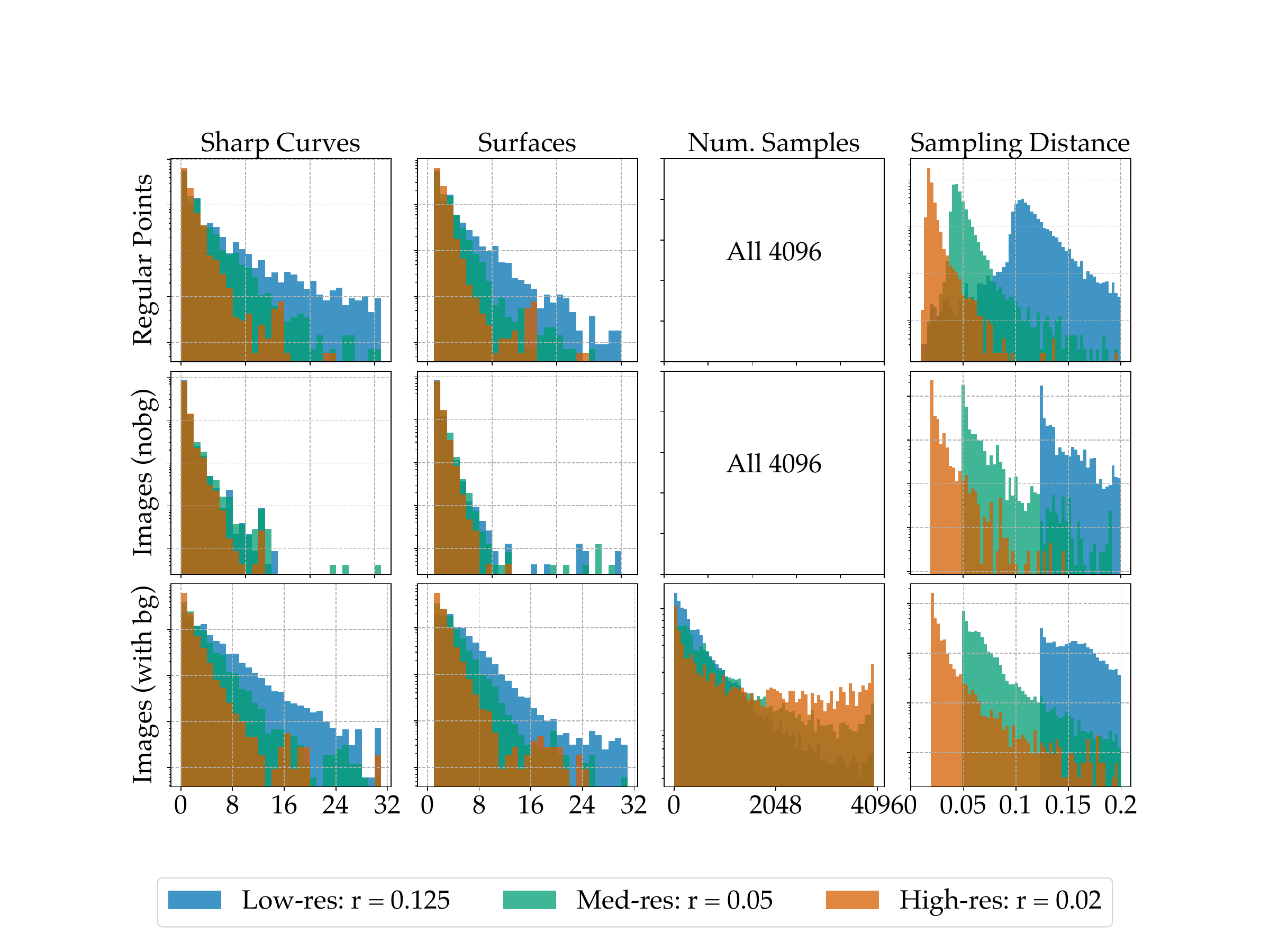}
}
\caption{Statistically, our local patch-based datasets differ substantially with respect to both sampling distance $r$, sampling pattern (regular and irregular), and data flags. 
We opt for range-images (with bg, lower row), as it statistically is able to include a wider variety of sampled geometry.}
\label{fig:fig_def_sim_patch_stats}
\end{figure}

\paragraph{Choosing Projection Planes.}
As outlined in the main text, our image-based datasets consist of range-image data obtained using a set of orthogonal projections.  
Each projection corresponds to a choice of a plane and placement of the image $64 \times 64$ grid (a virtual camera sensor) in the plane. 
The plane orientation is computed by composing three coordinate frame transformations, that help achieve larger degree of diversity in out datasets:
\begin{enumerate}
\item[1)] We pick a point on a sphere around the object and start with the tangent plane to the sphere; 
\item[2)] We translate the image in the picked plane, to capture different parts of the object from this view direction, by offsetting camera frame origin by $(s_x \frac {i_x}  {n_x}, s_y \frac {i_y} {n_y})$, where a $(s_x, s_y)$ is the object's bounding-box extent, as seen from picked view direction, $n_x, n_y$ are number of translations performed along camera x- and y-axes, respectively, and $i_{x,y} = -n_{x,y}/2, \ldots, n_{x,y} / 2$;
\item[3)] We rotate the sample grid orientation in the plane by choosing an uniformly distributed angle of rotation around the $z$-axis of the camera.
\end{enumerate}

\noindent \paragraph{Forming Mesh Patches.}
We form mesh patches and select feature curves for each patch by extracting entire surface spline regions that are found by association to any of the sampled points, along with their adjacent curves, removing boundary curves (see Section 4.1 in the main text).
This helps to ensure that the mesh patch does not have holes consisting of separate triangles not being encountered by raycasting.

\noindent \paragraph{Computing Annotations.}
We compute distance-to-feature annotations between points and sharp edges in extracted mesh patches using a fast implementation of KD-tree over axis-aligned bounding-boxes enclosing sharp edges, enabling us to compute annotations for millions of point samples quickly.

\begin{figure}[b!]
\centerline{
\includegraphics[
width=0.995\columnwidth,
trim=19em 0 8em 0,
clip=True]
{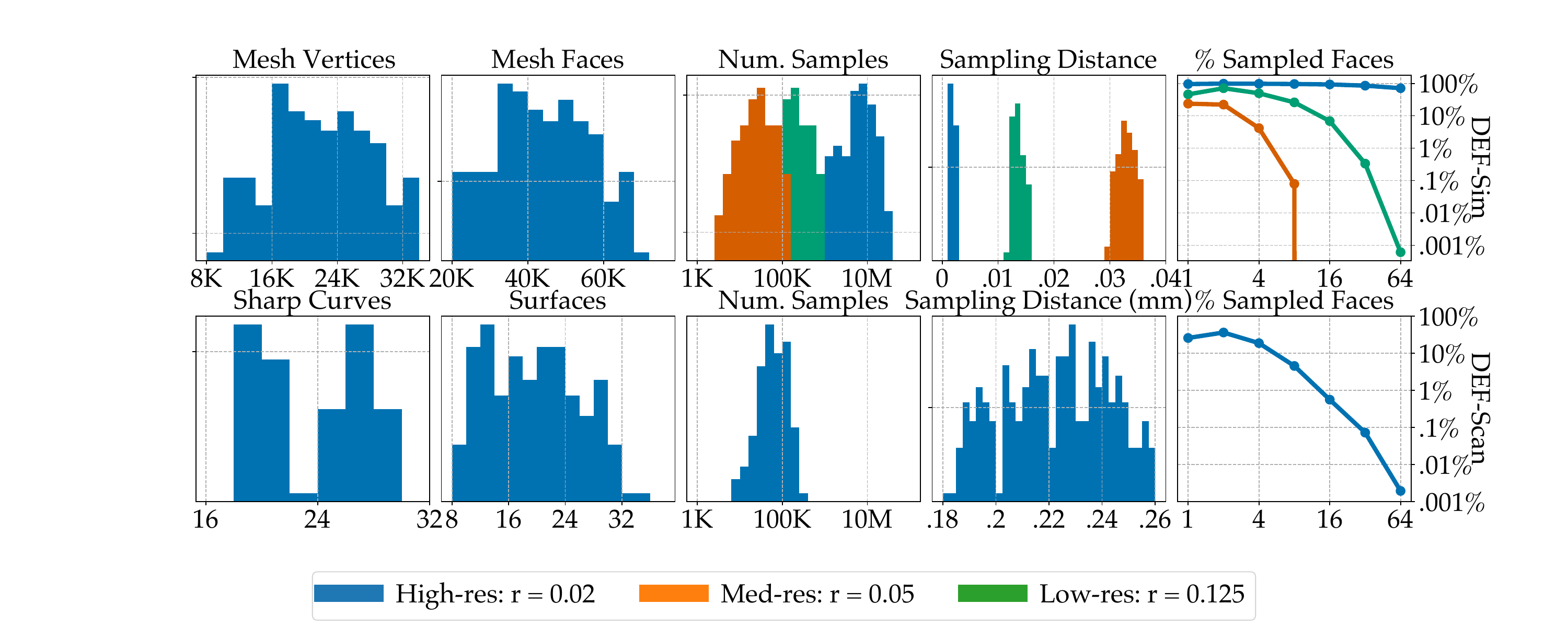}
}
\caption{Statistically, our simulated (top row) and scanned (bottom row) complete 3D shape datasets vary with respect to sampling distance $r$, and DEF-Scan is similar to a medium-resolution version of DEF-Sim in terms of sampling density per feature.}
\label{fig:fig_def_complete_stats}
\end{figure}

\noindent \paragraph{Data Flagging.}
The extremely high variability of geometry in our datasets suggests additional data labeling using a number of \textit{data flags,} providing indicators of specific traits encountered in the data. We used the following Boolean data flags:
\begin{itemize}
    \item \textit{Coarse surfaces (by the number of edges):}  spline patches for which triangulated versions have less than 8 edges along any side.
    \item \textit{Coarse surfaces (by mesh angles):}  spline patches for which triangulated versions have a median difference in angles of adjacent faces exceeding 10 degrees.
    \item \textit{Deviating resolution:}  point patches where the average distance between samples deviates by more than $r / 2$ from the specified sampling distance.
    \item \textit{Sharpness discontinuities:}  point patches for which difference in distance annotation in any two neighboring points exceeds the Euclidean distance between the two.
    \item \textit{Bad face sampling:}  point patches for which the average number of point sampled on each face is not in the range $[r, 100r]$.
    \item \textit{Raycasting background:} set to \textit{true} for images where at least one pixel contains background values.
    \item \textit{Depth discontinuities:} set to \textit{true} for images where depth changes by more than $T = 0.5$ units in neighboring pixels.
\end{itemize}
Our final datasets (DEF-Sim) are formed so that all flags are required to be false, except for \textit{Depth discontinuities}, and \textit{Raycasting background}, that we allow to take arbitrary values when forming \emph{with bg} versions of our data.

\subsection{Summary and Statistics of Our Datasets}
\label{supp:detail-datasets-summary}

\begin{table*}[t]
\centering
% \resizebox{0.995\textwidth}{!}{%
\caption{Overview of all data collections used within this work. 
For complete models and DEF-Scan, we provide estimates of the percentage of sampled faces, the sampling distance, and the number of samples.
We use the following shorthands for Patch Selection and Annotation:
SR: patch selection and annotation based on local surface regions;
FM: patch selection and annotation based on full 3D mesh model.
We use the following shorthands for Sampling:
RC: range-images sampling obtained using raycasting; 
RC$^*$: for full models, we concatenate range scans sampled using raycasting;
S: sampling pattern emerging for real-world scanning.
We use the following shorthands for Noise:
$\Sigma_3 = \{0.005, 0.02, 0.08\}$: the set of three noise magnitudes used for complete 3D shapes.
$\Sigma_6 = \{0.0025, 0.005, 0.01, 0.02, 0.04, 0.08\}$: the set of six noise magnitudes used for complete 3D shapes.
$^*$ designates an estimate computed over the concatenated scans;
$z$: adding noise in the direction of $z$-axis of the virtual camera;
$z^*$: for full models, we concatenate noisy range scans sampled using raycasting;
S: noise pattern emerging for real-world scanning.
}
\begin{tabular}{@{}lcccccccccccc@{}}
\toprule
Dataset     &   
    \rotatebox[origin=l]{90}{Num. Samples}      & 
    \rotatebox[origin=l]{90}{\% Sampled Faces}  & 
    \rotatebox[origin=l]{90}{Sampling Dist. $r$}  & 
    \rotatebox[origin=l]{90}{Noise std $\sigma$}  & 
    Train   &
    Val     & 
    Test    &
    \rotatebox[origin=l]{90}{Patch Selection}     &
    \rotatebox[origin=l]{90}{Sampling}        & 
    \rotatebox[origin=l]{90}{Annotation}      & 
    \rotatebox[origin=l]{90}{Noise}           &
    bg              \\
\midrule
% Dataset                               & n_pts  & % fac &   r         &  noise     & train   & val  & tst & PS & Smpl    & Ann &  N  &  bg \\
DEF-Sim (patch-high-0-nobg-$N$)         & 4096   &  ---  &  0.02       &   ---      & 2K-256K &  32K & 32K & SR &  RC     & SR  & --- & \xmark \\
DEF-Sim (patch-med-0-nobg-64k)          & 4096   &  ---  &  0.05       &   ---      &  64K    &  32K & 32K & SR &  RC     & SR  & --- & \xmark \\
DEF-Sim (patch-low-0-nobg-64k)          & 4096   &  ---  &  0.125      &   ---      &  64K    &  32K & 32K & SR &  RC     & SR  & --- & \xmark \\
DEF-Sim (patch-high-$\sigma$-nobg-64K)  & 4096   &  ---  &  0.02       & $\Sigma_6$ &  64K    &  32K & 32K & SR &  RC     & SR  & $z$ & \xmark \\
DEF-Sim (patch-high-0-wbg-$N$)          & 2913   &  ---  &  0.02       &   ---      & 2K-256K &  32K & 32K & SR &  RC     & SR  & --- & \cmark \\
DEF-Sim (patch-med-0-wbg-64k)           & 1880   &  ---  &  0.05       &   ---      & 64K     &  32K & 32K & SR &  RC     & SR  & --- & \cmark \\
DEF-Sim (patch-low-0-wbg-64k)           & 1201   &  ---  &  0.125      &   ---      & 64K     &  32K & 32K & SR &  RC     & SR  & --- & \cmark \\
DEF-Sim (patch-high-$\sigma$-wbg-64K)   & 2913   &  ---  &  0.02       & $\Sigma_3$ & 64K     &  32K & 32K & SR &  RC     & SR  & $z$ & \cmark \\
DEF-Sim (complete-high-0-68)$^*$        & 8456K  &  98\% &  0.002      &   ---      &   ---   &  --- & 68  & FM &  RC$^*$ & SR  & --- & \cmark \\
DEF-Sim (complete-med-0-68)$^*$         &  225K  &  71\% &  0.013      &   ---      &   ---   &  --- & 68  & FM &  RC$^*$ & SR  & --- & \cmark \\
DEF-Sim (complete-low-0-68)$^*$         &   36k  &  22\% &  0.033      &   ---      &   ---   &  --- & 68  & FM &  RC$^*$ & SR  & --- & \cmark \\
DEF-Sim (complete-high-$\sigma$-68)$^*$ & 8456K  &  98\% &  0.002      & $\Sigma_3$ &   ---   &  --- & 68  & FM &  RC$^*$ & SR  & $z^*$ & \cmark \\
\midrule
DEF-Scan (patches-med)$^*$              & 6878   &  ---  &  0.5\,mm    &     ---    &   981   & 479  & 468 & FM &  S      & FM  &  S  & \cmark \\
DEF-Scan (complete-med-scan)$^*$        &  83K   &  36\% & 0.22\,mm    & 0.328\,mm  & 86   &  41  &  39 & FM &  S      & FM  &  S  & \cmark \\
\bottomrule
\end{tabular}%
% }

\label{tab:tab_datasets_summary}
\end{table*}

We have computed a number of statistical quantities to better understand and characterize our data collections. 
Table~\ref{tab:tab_datasets_summary} presents an overview of core statistics for datasets used in this work, and Figures~\ref{fig:fig_def_sim_patch_stats}--\ref{fig:fig_def_complete_stats} represent patch and complete model statistics for DEF-Sim and DEF-Scan, respectively. 
We confirm that we have developed a variety of diverse synthetic and real-world datasets suitable for training and testing methods of detection sharp geometric feature curves.

\section{Details on Reconstruction for Complete 3D Models}
\label{methods:full-model-detail}

\noindent \paragraph{Inference Functions.}
We infer the final distance-to-feature estimate by computing the value of a inference set-function $g(\cdot)$ given a set $D_p = \{ \widehat{d}_1(p), \ldots, \widehat{d}_n(p) \}$ of predictions obtained (either directly or by interpolation) for each sampled point $p$.
To process these predictions, we have experimented with the following variants of pointwise aggregation.
Basic aggregation methods:
\begin{itemize}
    \item averaging $g(D_p) = \frac{1}{|D_p|} \sum_{\widehat{d} \in D_p} \widehat{d}$, 
    computing median $g \equiv \text{median}$, 
    and extracting minimum: $g \equiv \text{min}$.

    \item computing truncated average and minimum, computed by removing the largest and smallest 20\% of values, then computing the corresponding quantity;

    \item to perform inference based on predictions obtained using segmentation methods (e.g., \cite{merigot2010voronoi,raina2019sharpness,yu2018ec}), one can use the following simple scheme.
    Individual predictions $\widehat{d}_1(p), \ldots, \widehat{d}_n(p)$, with $\widehat{d}_i(p) \in \{0, 1\}$, can be combined using $g(D_p) = \mathbb{I}_{[T, 1]} (\frac{1}{|D_p|} \sum_{\widehat{d} \in D_p} \widehat{d})$, i.e. setting the fused prediction to 1 (sharp) when an average predicted value exceeds a threshold $T$.
\end{itemize}
Predictions obtained using one of the basic methods can be post-processed to improve smoothness by:
\begin{itemize}
    \item minimizing $L_2$ or total-variation (TV) based functionals of the form:
    \begin{equation*}
    \min_{\{\widehat{d}(p)\}} ||\widehat{d}(p) - \widehat{d}^0(p)|| + 
    \alpha \sum_{k=1}^K||\widehat{d}(p) - \widehat{d}(\text{NN}_k(p)||^{\gamma},
    \end{equation*}
    ($\text{NN}_k(p)$ denotes the $k$th nearest neighbor of the point $p$, we used $K=50$ and $\gamma \in \{1, 2\}$);

    \item fitting a robust version of local linear regression~\cite{huber1973robust} (we extract local point patches of $K=50$ neighbors of each point, reduce their feature dimensionality to 2, fit a outlier-robust linear regression model~\cite{owen2006robust} using the \texttt{scikit-learn} implementation (\texttt{HuberRegressor}), and extract predictions in the seed point).
\end{itemize}
Overall, we have found that setting $g \equiv \text{min}$ produces the best results for our test samples set.

\noindent \paragraph{Details on Transferring Predictions across Image Views.}
For a 3D point~$p$, we perform interpolation and estimation of visibility $\widehat{v}^{s\to t}(p)$ as indicated below.
To interpolate predicted distance values at the warped point $\widehat{p}$ in the reprojected image $I_s$, we construct a $K$-neighborhood $\{\text{NN}_k(\widehat{p})\}_{k=1}^K$ (we set $K=4$) and compute the linear bivariate B-spline representation of a surface~\cite{dierckx1995curve} using this neighborhood and respective distance values in $\widehat{d}_s(\widehat{p})$. 
We have chosen an implementation available in SciPy~\cite{virtanen2020scipy} and invoke the low-level \texttt{scipy.interpolate.bisplrep} over the wrapper \texttt{scipy.interpolate.interp2d} as the former offers direct control over the smoothness of the result. 
We evaluate the fitted B-spline at point $\widehat{p}$ to obtain an interpolated distance value (equivalently to a bilinear interpolation) and set the binary visibility mask $\widehat{v}^{s\to t}(p)$ to 1, or mark an interpolated value as not available when less than $K$ nearest neighbors exist within a Euclidean distance of $6 r$, where $r$ is the sampling distance. 
We do not perform interpolation for points on the patch boundary as we have discovered the corresponding estimates to be unstable.
In these instances, we set the binary visibility mask $\widehat{v}^{s\to t}(p)$ to zero.
We repeat the described process for all available pairs of images.

\section{Details on Experimental Evaluation}
\label{supp:experiments-detail}

\subsection{Experimental Setup}
\label{supp-experiments:setup-detail}

\paragraph{Measures of Quality.}
For each patch $P_i$, our computed quality measures are defined by:
\begin{align*}
\label{eq:rmse_metric}
    \rmse_i &= \frac{1}{\sqrt{N_i}}
    \sqrt{\sum\limits_{p \in P_i}
        \big(d_i(p) - \widehat{d}_i (p) \big)^2
    },\\
    \recall_i (T) &= \frac{
        \sum\limits_{p \in P_i} \widehat{s}_i(p) s_i(p)
    }{
        \sum\limits_{p \in P_i} s_i(p)
    },\\
    \fpr_i (T) &= \frac{
        \sum\limits_{p \in P_i} \widehat{s}_i(p)(1 - s_i(p))
    }{
        \sum\limits_{p \in P_i} (1 - s_i(p))
    },
\end{align*}
where $d_i(p)$ and $s_i(p)$ are the ground-truth distances and thresholded labels, respectively, $\widehat{d}_i(p)$, $\widehat{s}_i(p)$ their respective estimates, and $N_i = |P_i|$ the number of non-background samples in the patch $P_i$.
For methods producing hard segmentation labels, we directly use their predictions; for methods producing segmentation probability labels, we compute $\widehat{s}_i = \mathbbm{1} \big( \widehat{r}_i > 0.5 \big)$ where $\widehat{r}_i (p)$ is the estimated probability for $p$ to be a sharp point. 
We provide \qrmse for a collection of patches $\{P_i\}$ by computing the 95\% quantile of respective $\rmse_i$ values.
We calculate the metrics for a set of patches by averaging metrics obtained for individual patches.

To measure the curve extraction quality, we used metrics defined by:
\begin{align*}
% \label{eq:curve_metric}
    \chamfer_{P \rightarrow Q} &= \frac{1}{|P|} \sum_{p\in P} \inf_{q \in Q} \|p - q \| ^ 2,\\
    \chamfer &= \chamfer_{P \rightarrow Q} + \chamfer_{Q \rightarrow P},\\
    \hausdorff &= \max\{ \sup_{p \in P} \inf_{q \in Q} \| p - q \|, \,  \sup_{q \in Q} \inf_{p \in P} \| p - q \|\},
\end{align*}
where $P$ and $Q$ are point clouds that are compared.

Here Chamfer distance \chamfer reflects the average discrepancy in two sets of curves, and Hausdorff distance \hausdorff measures the worst-case deviation between the curves.

Our third metric, Sinkhorn distance \sinkhorn, is an approximation of the Wasserstein optimal transportation. It uses blurring the transport plan through the addition of an entropic penalty to reduce the computational cost. \sinkhorn is computed as a series of iterative updates, for more details refer to~\cite{feydy2019interpolating}.

\subsection{Parameter Choices}
\label{supp-experiments:param-choice-detail}

\emph{Voronoi Covariance Measure (VCM)}~\cite{merigot2010voronoi} 
We ran a direct grid search to obtain the set of parameters with the maximal \recall for each sampling distance and noise level. 
Each of the parameters was varied over a grid of 11 values: $ \{0.01, 0.05\} \bigcup \{0.1i\}_{i=1}^9$. 
For each combination we ran \emph{VCM} inference on the validation set, computed \recall value and determined the set of parameters maximizing the metric. 
The selected parameters are presented in Table~\ref{tab:vcm_params}. 

\begin{table}[b!]
\centering
\caption{Parameters of VCM for different types of data.}
\begin{tabular}{@{}ccccc@{}}
\toprule
Sampling & Noise & $R$ & $\rho$ & $T$ \\ 
distance $r$ & magnitude $\sigma$ & & & \\
\midrule
$r_{\text{high}}$ &  0, $r/8$, $r/4$, $r/2$, $r$    & 0.05 & 0.1 & 0.3 \\
$r_{\text{high}}$ & $2r$, $4r$    & 0.1 & 0.3 & 0.3 \\
$r_{\text{med}}$ & 0     & 0.1 & 0.1 & 0.4 \\
$r_{\text{low}}$ & 0     & 0.2 & 0.1 & 0.4 \\
\bottomrule
\end{tabular}

\label{tab:vcm_params}
\end{table}

\emph{Sharpness Fields (ShF)}~\cite{raina2019sharpness} 

\emph{ShF} outputs a real-valued field similar to ours, which has value 0 far from feature line and reaching 1 at the feature. 
In practice we observed that this field is more narrow than ours, meaning that for a fair comparison we needed to find a linear transformation to equalize them.
To do that, we implemented the following transformation selection procedure:
\[
\min_{\alpha} \sqrt{\frac{1}{N} \sum_{i=1}^N \left(\lbrack 1 - ShF_i \rbrack - \max \{ d_i / \alpha, 1 \} \right)^2},
\]
where $d_i$ is our ground truth distance-to-feature field of $i$-th patch, $ShF_i$ is a prediction by their network on $i$-th patch, $\alpha$ is an equalizing coefficient, $N$ is the size of validation set.
Intuitively, this functional measures \rmse between the predictions by \emph{ShF} and our transformed field. 
We computed these values for a range of coefficients $\alpha = \{0.01i\}_{i=1}^{10}$ on validation set and selected $\alpha = 0.06$ as the one minimizing this functional.

Other competitors \emph{Edge-Aware Consolidation Network (EC-Net)}~\cite{yu2018ec}, \emph{PIE-NET}~\cite{wang2020pie}, \emph{PC2WF}~\cite{Liu:2021:PC2WF} have no parameters to tune.

\subsection{Parameter Choices for Vectorization Pipeline}
\label{supp-experiments:parametric-param-choice-detail}

The sampling technique in DEF-Sim ensures that the pairwise point distance $r_{\text{high}}$ is 0.02 for individual images on average; we choose to relate all parameters to this value. 
We observed that the parameters with the strongest effect on the final result were the proximal points selection threshold $d_{\text{sharp}}$ and the corner detection threshold $T_{\text{corner}}$.
To set these parameters, we implemented a parameter sweep over a grid: we varied $d_{\text{sharp}}$ in the range $[2r,4r]$, and $T_{\text{corner}}$ in the range $[0.6,0.85]$.
For each set of parameters, we ran the whole vectorization procedure and computed a symmetric \chamfer between the sampled spline curves and a point set $P_{\text{sharp}}$ that consists only of points with estimated distance $\widehat d$ less than $d_{\text{sharp}}$, thus measuring the goodness of fit. 
The resulting set of parameters is chosen by the lowest value of \chamfer.
We found reasonable default settings for the rest of the tunable parameters that do not affect the result as much.

For the endpoint detection, we choose $R_{\text{endpoint}} = 10r$. 
The threshold $T_{\text{endpoint}} = 0.6$, which means there should be 60\% more points on one side from the query point compared to the other side to consider a ball center to be the curve endpoint.
Finally, the choice of splitting threshold is $T_{\text{split}} = 4r$. 
We want the polyline controlled by this value to accurately reflect the corresponding curve geometry.

Finally, we discuss parameters $\mathcal{N}_i$, $T_{\text{variance}}$, and $R_{\text{corner}}$ used in corner detection procedure. It is designed as aggregation of several corner estimates, hence it doesn't require setting the exact parameter values. We vary $R_{\text{corner}}$ in the range of $5r, \ldots, 8r$, the number of neighbor sets $\mathcal{N}_i$ in the range 10, 20, 40, and the threshold $T_{\text{variance}}$ in the range 5, 10, 15, 20, 25.
With this grid of parameters we obtain 60  different corner estimates, for each set of estimates we compute the fraction of cases where a specific set $B_i$ was labeled as a corner and normalize it by 60, eventually obtaining a probability for each set to be a corner.

\subsection{More Ablative Experiments}
\label{supp:experimental-ablative}

\noindent \paragraph{Data volume.} 
As a part of the ablative studies, we conduct training on datasets of increasing size. 
We performed training for each dataset size (we used noise-free patch datasets with sampling distance $r = r_{\text{high}}$) until convergence.
We present results in Table~\ref{tab:perfomance_vs_dataset_size}, where we observe that metric values stabilize for datasets with around 64k training patches.
Not surprisingly, larger training datasets improve performance.
The subsequent experiments were performed with 64k training patches.

\noindent \paragraph{Model capacity.} 
We performed an additional experiment to identify the optimal configuration of our backbone CNN. 
We instantiated a series of ResNet~\cite{he2016deep} backbones with significantly varying number of parameters and trained each until convergence on the validation set. 
Table~\ref{tab:perfomance_vs_model_size} presents results, that generally indicate some increase in performance for larger models. 
We select the ResNet-152 backbone network for all subsequent studies.

\begingroup
\tabcolsep=4pt
\def\arraystretch{1.075}

\begin{table}[]
\centering
\caption{For DEF networks trained on datasets of increasing size, performance generally stabilizes for 16K--64K patches (DEF-Sim, no bg, $r = r_{\text{high}}, \sigma = 0$).
We opt for 64K patches as this dataset size provide the most diversity for training.}
\resizebox{\columnwidth}{!}{%
\begin{tabular}{@{}lcccc@{}}
\toprule
Train Size & \rmse$\downarrow$ & \qrmse$\downarrow$ & \recall$(1r)$, \%$\uparrow$ & \fpr$(1r)$, \%$\downarrow$    \\
 & $\times 10^{-3}$ & $\times 10^{-3}$ & & \\ 
\midrule
2k          & 118.7 & 545.7 & 0     & 0 \\
4k          & 138.6 & 609.4 & 0     & 0 \\
8k          & 105.5 & 581.4 & 37.65 & 0.1 \\
16k         & 57.5  & 341.8 & 63.4  & 0.18 \\
32k         & 61.4  & 403.2 & 70.5  & 0.22 \\
64k (Ours)  & 61.5  & 361.1 & 57.36 & 0.06 \\
256k        & 85    & 424.9 & 45.01 & 0.07 \\
\bottomrule
\end{tabular}
}

\label{tab:perfomance_vs_dataset_size}
\end{table}

\endgroup

\begingroup
\tabcolsep=4pt
\def\arraystretch{1.075}

\begin{table}[]
\centering
\caption{As image-based backbone grows in capacity, DEF results generally improve on validation set (DEF-Sim, no bg, $r = r_{\text{high}}, \sigma = 0$). 
We end up selecting the largest resnet152 backbone for the remaining experiments.}
\resizebox{\columnwidth}{!}{%
\begin{tabular}{@{}lccccc@{}}
\toprule
Backbone & \rmse$\downarrow$ & \qrmse$\downarrow$ & \recall$(1r)$, \%$\uparrow$ & \fpr$(1r)$, \%$\downarrow$    \\
(\# Params) & $\times 10^{-3}$ & $\times 10^{-3}$ & & \\ 
\midrule
resnet26  (34.4\,M)   & 9.3 & 37 & 72.47 & 0.02 \\
resnet34  (30\,M)     & 9.8 & 34.7 & \textbf{83.81} & 0.02 \\
resnet50  (44\,M)     & 7.3 & 24 & 82.12 & 0.02 \\
resnet101 (63\,M)     & 8.2 & 26.5 & 79.85 & 0.02 \\
resnet152 (78.6\,M)   & \textbf{7.2} & \textbf{23.1} & 83.39 & 0.02 \\
\bottomrule
\end{tabular}
}

\label{tab:perfomance_vs_model_size}
\end{table}

\endgroup

\noindent \paragraph{Additional Inputs.}
We evaluated the effect of adding auxiliary inputs by concatenating the \emph{VCM} prior sharpness estimates, normal vectors, and both simultaneously to the raw range images (sampling distance $r = r_{\text{high}}$, no noise). 
We trained ResNet-152 on depth images with additional input channels; we present statistical results in Table~\ref{tab:patch_perfomance_ablation_study}, upper rows, where we compare these configurations againt the baseline where an input range-image $P$ is regressed onto a distance labels $d(p)$.
Metric values demonstrate that no conclusive gain in performance is observed for regression metrics, compared to such a baseline.
Hence, we further train on range-images without additional inputs in all instances.

\noindent \paragraph{Additional Outputs.} 
Similarly to the previous experiments, we performed an ablative study to understand how the auxiliary tasks affect feature line estimation performance. 
We experimented with concatenating direction-to-feature, ground-truth normals, and both simultaneously to the distance labels $d(p)$, and adding additional heads to our network to predict these quantities. 
We present statistical results of this experiment in Table~\ref{tab:patch_perfomance_ablation_study}, middle rows.
In all cases regressing the normals, directions towards the feature line, or both of them at the same time did not lead to increasing the quality of feature line extraction. 
Hence, we proceed further without using any additional outputs.

\begingroup
\tabcolsep=4pt
\def\arraystretch{1.075}

\NewDocumentCommand\dist{}{%
    \ifmmode d(p)
    \else $d(p)$
    \fi %
}
\NewDocumentCommand\norm{}{%
    \ifmmode n(p)
    \else $n(p)$
    \fi %
}
\NewDocumentCommand\svcm{}{%
    \ifmmode \widehat{s}_{\text{VCM}}(p)
    \else $\widehat{s}_{\text{VCM}}(p)$
    \fi %
}
\NewDocumentCommand\distpred{}{%
    \ifmmode \widehat{d}(p)
    \else $\widehat{d}(p)$
    \fi %
}
\NewDocumentCommand\normpred{}{%
    \ifmmode \widehat{n}(p)
    \else $\widehat{n}(p)$
    \fi %
}
\NewDocumentCommand\dirpred{}{%
    \ifmmode \widehat{r}(p)
    \else $\widehat{r}(p)$
    \fi %
}

\begin{table}[]
\centering
\caption{We perform experiments to study the effect of introducing additional signals at the input, and additional supervision at the output of our networks (results obtained on DEF-Sim, no bg, $r = r_{\text{high}}, \sigma = 0$). 
As input in addition to depth image $P$, we supply ground-truth normals $n(p)$, prior sharpness estimates $\widehat{s}_{\text{VCM}}(p)$ obtained by \emph{VCM}, and their combinations.
As output in addition to distance estimates $\widehat{d}(p)$, we require our model to predict normals $\widehat{n}(p)$, $3\times 1$ directions $\widehat{r}(p)$ to the closest point on the sharp feature curve, and their combinations. 
We end up selecting the most basic scheme where we predict distance estimates $\widehat{d}(p)$ from the input depth image $P$.}
\resizebox{\columnwidth}{!}{%
\begin{tabular}{@{}llcccc@{}}
\toprule
Input & Output & \rmse$\downarrow$ & \qrmse$\downarrow$ & \recall$(1r)$, \%$\uparrow$ & \fpr$(1r)$, \%$\downarrow$    \\
 &  & $\times 10^{-3}$ & $\times 10^{-3}$ & & \\ 
\midrule
$P, \norm$          & $\distpred$               & 7.2 & 34.1 & 69.31 & 0.02 \\
$P, \svcm$          & $\distpred$               & 8.6 & 26.8 & 78.09 & 0.03 \\
$P, \norm, \svcm$   & $\distpred$               & 6.2 & 25.9 & 76.53 & 0.02 \\
\midrule
$P$                 & $\distpred, \normpred$    & 8.1 & 31.8 & 74.69 & 0.01 \\
$P$                 & $\distpred, \normpred, \dirpred$ & 8.5 & 33.2 & 74.82 & 0.02 \\
$P$                 & $\distpred, \dirpred$     & 8.3 & 33.9 & 74.09 & 0.02 \\
\midrule
$P$                 & $\distpred$               & \textbf{7.2} & \textbf{23.1} & \textbf{83.39} & \textbf{0.02} \\
\bottomrule
\end{tabular}
}

\label{tab:patch_perfomance_ablation_study}
\end{table}

\endgroup

\subsection{More Experiments on Complete 3D Models}
\label{supp:experimental-full-models}

We have performed more experiments to investigate the limits of robustness of our method to reduction in sampling density and increase in noise strength. 
To this end, we employed $n_v = 18$ views of the same models in DEF-Sim dataset, but have augmented respective range-images with noise acting in the camera direction, and performed sampling to model decrease in point density as $r$ grows.
Table~\ref{tab:full_models_robustness} presents quantitative evaluation of our method with such input data. 
We conclude that sparse data, at least the ones we studied, did not result in a significant degradation of our approach, apart from the large increase in the \fpr measure, indicating that more false positives shall be identified.
Adding noise, in contrast, significantly impacts results, as the method tends to no longer detect features, instead focusing on averaging predictions across the shape in an attempt to reduce noise.
Even so, our method remains generally stable for noise magnitudes of up to $r$, all with using only 18 views for reconstruction, that we have identified is a modest number of views.

\begingroup
\tabcolsep=4pt
\def\arraystretch{1.075}

\begin{table*}[]
\centering
% \resizebox{\columnwidth}{!}{%
\caption{Results of reconstructing object-level distance-to-feature field (DEF-Sim, 68 shapes) indicate that DEF is able to perform robustly w.r.t. sampling distance, with only \fpr indicating performance degradation for lower resolution datasets. 
DEF is additionally resilient to noise with signal-to-noise ratios of up to 1:1, as indicated by $\recall(4r)$; for larger noise magnitudes, performance inevitably degrades.}
\begin{tabular}{@{}cccccccc@{}}
\toprule
Sampling & 
    Noise &
    \rmse$\downarrow$ & 
    \qrmse$\downarrow$ & 
    \recall$(1r)$, \%$\uparrow$ & 
    \fpr$(1r)$, \%$\downarrow$  & 
    \recall$(4r)$, \%$\uparrow$ & 
    \fpr$(4r)$, \%$\downarrow$ \\
distance $r$ & 
magnitude $\sigma$ &
 $\times 10^{-3}$ & 
 $\times 10^{-3}$ & & & & \\ 
\midrule
$r_{\text{high}}$   & 0             & 100.2 & 214.1 & 47.9  & 0.2  & 92.3 &  2   \\
$r_{\text{med}}$    & 0             &  88.4 & 197   & 38.4  & 0.2  & 94.1 &  4.5 \\
$r_{\text{low}}$    & 0             & 134.7 & 272.1 & 47.3  & 2.1  & 95.5 & 22.7 \\
\midrule
$r_{\text{high}}$   & $r / 4$       & 658.5 & 817   & 56.7  & 2.6  & 96.1 & 16.7 \\
$r_{\text{high}}$   & $r$           & 651.4 & 786.2 &  6    & 0.3  & 71.8 & 10.5 \\
$r_{\text{high}}$   & $4 r$         & 541.5 & 730.8 &  0    & 0    & 37.9 & 4.9 \\
\bottomrule
\end{tabular}
% }

\label{tab:full_models_robustness}
\end{table*}

\endgroup

\begin{figure}[b!]
\centerline{
\includegraphics[
width=0.995\columnwidth]
{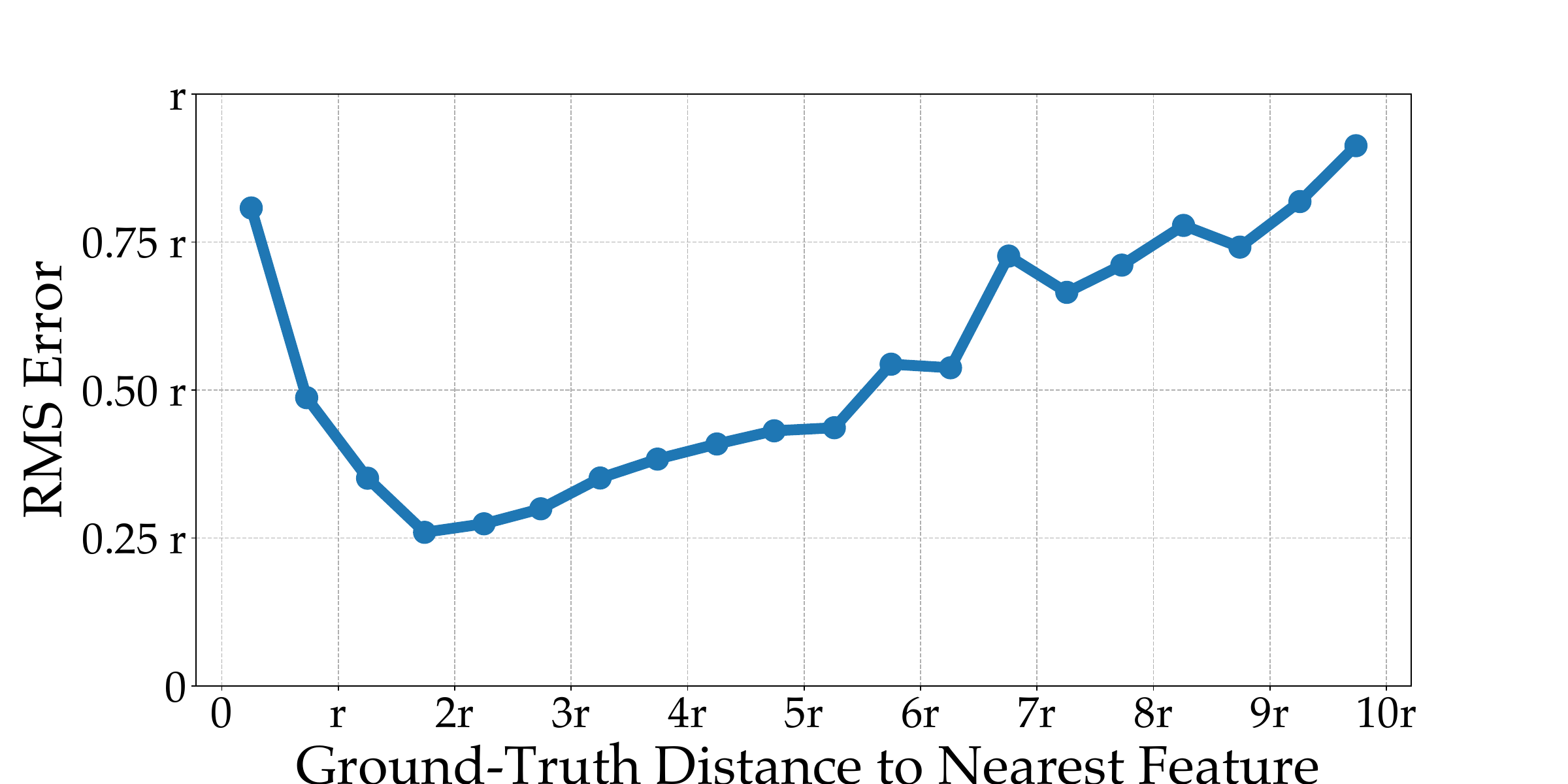}
}
\caption{Statistically, our method has the lowest \rmse in locations spaced around $r$--$2r$ from the sharp feature curves. 
This observation explains why in particular instances our method demonstrates performance drops in \recall$(1r)$ while remaining robust according to \recall$(4r)$.}
\label{fig:fig_rmse_vs_distance_fusion}
\end{figure}

We additionally investigated how our method's performance depends on the location of the predictions, relative to a sharp feature curve.
As can be seen in Figure~\ref{fig:fig_rmse_vs_distance_fusion}, our method has a performance peak at around $r$ to $2r$, which indicates that predicting distances in locations exactly on the feature curve or far away from the curve might be more difficult than doing so in some proximity from the curve.

\section{Alternative Point-Based Pipeline}
\label{supp:point-based-alternative}

\subsection{Dataset Construction}
\label{supp:point-based-datasets}

We describe an alternative procedure to obtain point-sampled patches $P$ with $N = |P| = 4096$ points with distance-to-feature annotations $d(p), p \in P$.

\noindent \paragraph{Dataset Design.}
We follow exactly the same procedure for feature definition, feature selection, distance-to-feature computation, deciding on feature size, and computing sampling density, as described in the main text. 

\noindent \paragraph{Patch and Feature Selection.}
We extract local patches from triangulated 3D shapes by selecting all mesh faces inside or intersecting with a sphere of radius $\sqrt{N} r/2$ ($N = 64^2$), centered at 128~uniformly distributed (using Poisson Disk Sampling~\cite{bowers2010parallel}) points on the model surface. 
Among all connected parts of the mesh inside the sphere, if any, the largest one is selected. 

\noindent \paragraph{Shape Sampling.}
We obtain point clouds using Poisson Disk Sampling~\cite{bowers2010parallel}, similar to~\cite{wang2020pie} and unlike~\cite{yu2018ec} that use raycasting similarly to our image-based datasets.
If the number of samples generated on this patch with Poisson disk sampling is larger than $N$, $N$ points closest to the center are retained; if the number of sampled points is less than $N$, this particular patch is discarded. 

\noindent \paragraph{Patch-Based Datasets.}
We have constructed a dataset of 65,536 patches for training, 32,000 patches for validation, and 32,000 for testing our model.

\noindent \paragraph{Complete 3D Model Datasets.}
To construct a sampled and annotated version for a complete 3D model, we first compute a Poisson Disk Sampling of the complete 3D mesh.
Next, to compute distance-to-feature annotations over the complete 3D shape, we extract overlapping local regions in the mesh as mentioned above, associate the sampled points to each local mesh region, and annotate these points using our normal procedure; this results in multiple annotations available for each point as local regions overlap. 
We compute a minimum over the available annotations in each point to produce the final complete annotations.

\subsection{Methods}
\label{supp:point-based-methods}

\noindent \paragraph{Learning Architecture.}
We use the DGCNN architecture~\cite{wang2019dynamic} and systematically vary the size of the base network, by simultaneously increasing both width $W$ and depth $D$ according to the relations $W=64 \times 1.35^k$, $D = 3 + k$, varying $k$ from $-2$ to $3$.
The quantitative results suggest that for $k \geqslant 1$ the gains in performance stabilize; we end up choosing $k = 3$, DGCNN with depth $D = 6$, width $W = 158$. 
While the DGCNN model was trained using the Histogram loss using the supervision from ground-truth distances $d(p)$ only, we discovered that adding prior sharpness estimates from \emph{VCM} has the potential improve performance considerably; this is in contrast with the effect \emph{VCM} has on image-based data.
However, adding \emph{VCM} labels requires an additional effort to compute these scores before running the model on the new shapes.

\begingroup
\tabcolsep=4pt
\def\arraystretch{1.075}

\begin{table*}[]
\centering
\caption{Experiments using point-based DGCNN~\cite{wang2019dynamic} demonstrate promising results for unstructured sampling patterns with uniform sampling; however, image-sampled patches tend to be significantly non-uniform, impairing DGCNN performance; adding prior sharpness estimates from \emph{VCM} yields no advantage for this method.}
\resizebox{0.9\textwidth}{!}{%
\begin{tabular}{@{}llccccc@{}}
\toprule
Dataset &
  Method & 
  \rmse$\downarrow$   & 
  \qrmse$\downarrow$   & 
  \recall$(1r)$, \%$\uparrow$   & 
  \fpr$(1r)$, \%$\downarrow$ \\
   & &
  $\times 10^{-3}$ & 
  $\times 10^{-3}$ & 
  & \\ 
\midrule
Unstructured points & 
  DGCNN + Histogram loss &
  10.0 &
  38.1 &
  89.5 &
  $7 \times 10^{-2}$ \\
Unstructured points &
  DGCNN + Histogram loss + VCM & 
  7.8 & 
  25.6 & 
  90.0 & 
  $8 \times 10^{-2}$ \\
\midrule
Regular images (no bg, reprojected to points) &
  DGCNN + Histogram loss & 
  11.3  & 
  55.5 & 
  80.9 & 
  $3.7 \times 10^{-2}$ \\
Regular images (no bg, reprojected to points) &
  DGCNN + Histogram loss + VCM & 
  13.6  & 
  70.0  & 
  68.8  & 
  $4.8 \times 10^{-2}$ \\
\bottomrule
\end{tabular}
}

\label{tab:point_based_learning_framework}
\end{table*}

\endgroup

\begin{figure*}[t]
\centering{
\includegraphics[width=0.95\textwidth]
{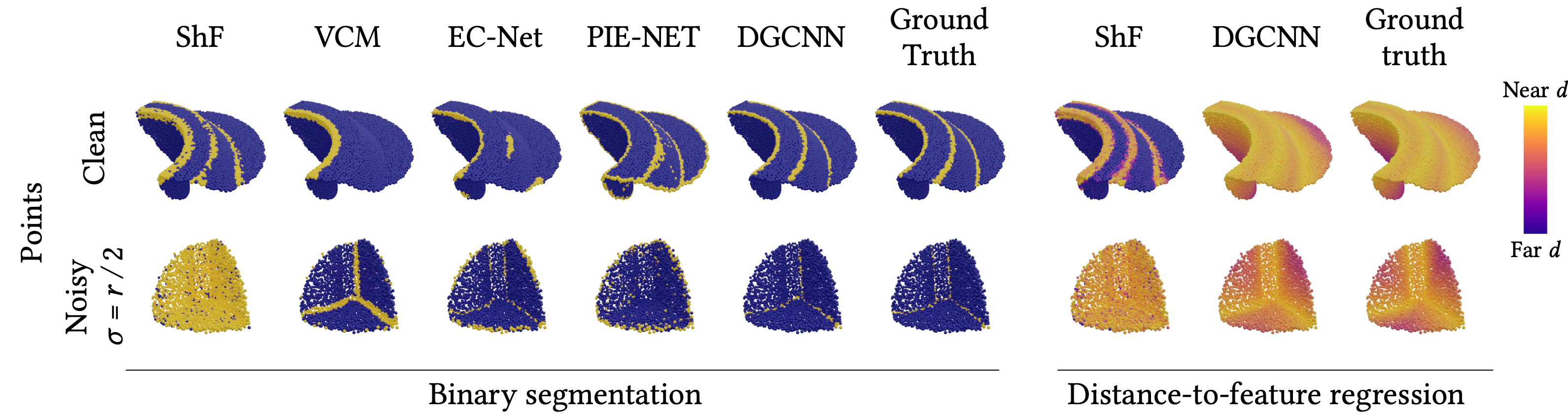}
}
\caption{Comparative prediction results for a DGCNN model pre-trained on a point-based collection vs. the competitor approaches \emph{ShF}, \emph{VCM}, \emph{EC-Net}, and \emph{PIE-NET}.
The DGCNN model trained on the datasets we use is able to perform competitively on sampled data. }
\label{fig:pointbased_synthetic_patches}
\end{figure*}

\noindent \paragraph{Reconstruction for Complete 3D Models.}
To compute a distance-to-feature field for an input complete 3D shape $P$, we first extract point patches $P_i$ with $4096$ points. 
We use an adjacency graph of the points based on their $k$ nearest neighbors (we use $k = 5$), extracting the largest connected component of this graph.
Each patch is obtained by a breadth-first search from a vertex, and we add patches until each point is covered by at least 10 patches.
For each of these local patches $P_i$, we predict a distance-to-feature field $\widehat{d}_i(p)$ using a DGCNN model, resulting in a set of predictions 
\begin{equation}
D_p = 
\big\{ 
  d_p \,
    | \, d_p = \widehat{d}_i(p) = \text{DGCNN}(p | P_i), p \in P_i
\big\}_{p = 1}^{|P|}
\end{equation}
for each point $p$ in the input point cloud (here $\text{DGCNN}(p | P_i)$ denotes DGCNN prediction in the same point $p$ given the context point patch $P_i$).
The set~$D_p$ of predictions for all patches containing~$p$ is filtered by excluding predictions from 20\% of points in the patch furthest away from its center of mass.
Finally, we compute a minimum over all predictions of the distance $\widehat{d}(p) = \min \limits_{d \in D_p} d$.
Other possibilities described in Section~\ref{methods:full-model-detail} of this document can also be applied.

\subsection{Experimental Results}
\label{supp:point-based-results}

\noindent \paragraph{Training Details.}
All training patches consist of 4,096 points; we applied a random 3D rotation to each patch as an augmentation.

\noindent \paragraph{Patch-Based Results.}
Table~\ref{tab:point_based_learning_framework} contains statistical results of the influence of sampling pattern and prior sharpness estimates from \emph{VCM} on performance. 
We note that the two datasets are not directly comparable, even though they represent point-sampled geometry with the same feature size distribution (sampling distance $r = r_{\text{high}}$).
Specifically, while the point-sampled geometry does contain similar geometric patterns, the sampling pattern is more regular, which is ensured by the Poisson Disk Sampling; in contrast, range-scans produced by ray-casting have significantly non-uniform sampling, where density may vary significantly for surfaces on either side of a sharp feature curve.
We conclude that training a model from point-based data benefits from adding prior hard sharpness estimates from the \emph{VCM} method, likely due to benefits offered by sampling; this is not the case for image-sampled data.

We demonstrate a qualitative comparison of patch-based feature estimation performance in the same fashion as for image-based datasets in the main text. 
For this experiment, competitor approaches were optimized according to the same procedure as for the image-based datasets (see Section~\ref{supp-experiments:setup-detail} of this document).
Figure~\ref{fig:pointbased_synthetic_patches} displays a comparison of point patches where competitor approaches are compared against a DGCNN model trained on a patch-based dataset (see Section~\ref{supp:point-based-datasets} above).
We conclude that a point-based network pre-trained on the kind of datasets we use can generalize well to unseen instances and present a viable alternative to competitor approaches.

\noindent \paragraph{Results on Complete 3D Models.}
We make an effort to compare the DGCNN-based method for reconstructing distance-to-feature fields for complete 3D models (see Section~\ref{supp:point-based-methods}) on the same data collection of 68 shapes as our method, DEF, was tested on. 
The results in Table~7 of the main text indicate that DGCNN-based method is capable of producing nearly the same \recall and \fpr values, however it is outperformed by a large margin ($2\times$) according to \rmse measure. 
As we require our distance field to be as accurate as possible (e.g., for the reconstruction of the set of parametric representations of sharp feature curves), we made an eventual choice in favor of the image-based method.

  \bibliographystyle{ACM-Reference-Format}
  \bibliography{sharp_features}

\end{document}